\documentclass[acmsmall, nonacm]{acmart}

\renewcommand\footnotetextcopyrightpermission[1]{} 
\usepackage{fancyhdr}

\usepackage{xspace}

\usepackage{bbm}

\makeatletter
\DeclareRobustCommand\onedot{\futurelet\@let@token\@onedot}
\def\@onedot{\ifx\@let@token.\else.\null\fi\xspace}

\def\eg{\emph{e.g}\onedot} 
\def\ie{\emph{i.e}\onedot} 
\def\cf{\emph{c.f}\onedot}

\def\etal{\emph{et al}\onedot}
\makeatother

\usepackage{amsmath}

\usepackage{amsfonts}
\usepackage{mathtools}
\usepackage{booktabs}
\usepackage{multirow}
\usepackage{color}
\usepackage{adjustbox}

 \usepackage[caption=false,font=footnotesize]{subfig}

\AtBeginDocument{%
  \providecommand\BibTeX{{%
    \normalfont B\kern-0.5em{\scshape i\kern-0.25em b}\kern-0.8em\TeX}}}

\setcopyright{acmcopyright}
\copyrightyear{2021}
\acmYear{2021}
\acmDOI{10.1145/1122445.1122456}

\acmJournal{tomm}
\acmVolume{1}
\acmNumber{1}
\acmArticle{1}
\acmMonth{9}

\begin{document}

\title{A Survey on Temporal Sentence Grounding in Videos}

\author{Xiaohan Lan}
\email{lanxh20@mails.tsinghua.edu.cn}
\affiliation{%
  \institution{Tsinghua Shenzhen International Graduate School}
  \streetaddress{University Town of Shenzhen}
  \city{Shenzhen}
  \country{China}}

\author{Yitian Yuan}
\authornote{This work started when Yitian Yuan was at Tsinghua university.}
\affiliation{%
  \institution{Meituan}
  \city{Beijing}
  \country{China}}
\email{yuanyitian@foxmail.com}

\author{Xin Wang}
\authornote{Corresponding Authors}
\affiliation{%
  \institution{Tsinghua University}
  \streetaddress{30 Shuangqing Rd}
  \city{Beijing}
  \country{China}}
 \email{xin_wang@tsinghua.edu.cn}

\author{Zhi Wang}
\affiliation{%
  \institution{Tsinghua Shenzhen International Graduate School}
  \streetaddress{University Town of Shenzhen}
  \city{Shenzhen}
  \country{China}}
 \email{wangzhi@sz.tsinghua.edu.cn}

\author{Wenwu Zhu}
\authornotemark[2]
\affiliation{%
  \institution{Tsinghua University}
  \streetaddress{30 Shuangqing Rd}
  \city{Beijing}
  \country{China}}
 \email{wwzhu@tsinghua.edu.cn}
  
\renewcommand{\shortauthors}{Lan, et al.}

\begin{abstract}
Temporal sentence grounding in videos~(TSGV), which aims to localize one target segment from an untrimmed video with respect to a given sentence query, has drawn increasing attentions in the research community over the past few years. 
Different from the task of temporal action localization, TSGV is more flexible since it can locate complicated activities via natural languages, without restrictions from predefined action categories. 
Meanwhile, TSGV is more challenging since it requires both textual and visual understanding for semantic alignment between two modalities~(\ie, text and video). 
In this survey, we give a comprehensive overview for TSGV, which i) summarizes the taxonomy of existing methods, ii) provides a detailed description of the evaluation protocols~(\ie, datasets and metrics) to be used in TSGV, and iii) in-depth discusses potential problems of current benchmarking designs and research directions for further investigations.
To the best of our knowledge, this is the first systematic survey on temporal sentence grounding.
More specifically, we first discuss existing TSGV approaches by grouping them into four categories, \ie, two-stage methods, end-to-end methods, reinforcement learning-based methods, and weakly supervised methods. 
Then we present the benchmark datasets and evaluation metrics to assess current research progress. Finally, we discuss some limitations in TSGV through pointing out potential problems improperly resolved in the current evaluation protocols, 
which may push forwards more cutting edge research in TSGV. Besides, we also share our insights on several promising directions, including three typical tasks with new and practical settings based on TSGV.
\end{abstract}



\maketitle



\section{Introduction}

With the increasing development of multimedia technologies on mobile phones and other terminal devices, people have gained easier access to videos from all around the world. Compared with other mediums for information transmission and exchange like texts and images, videos contain more dynamic activities and are of richer semantics to convey complex while understandable information. Basically, one video is composed of a continuing sequence of frame images possibly accompanied by audio and subtitles. Moreover, the videos from online websites in the wild are also surrounded by multiple forms of natural language texts~(\eg, comments written by video viewers, video descriptions uploaded by 
creators, recommendation reasons edited by website editors). Thus, videos have natural advantages for multimedia intelligence exploration and research.
However, the raw videos are too redundant and of high information sparsity against the user-specific retrieval demands. Furthermore, it is also challenging to maintain and management these raw videos since they need to occupy a huge number of storage resources.
Therefore, the ability to quickly retrieve a specific video segment~(\ie, moment) from a long untrimmed video can allow users to locate highlighted moments of their interests conveniently and help information providers to optimize the storage fundamentally, thus being of great importance and interest in the research community. 

\begin{figure}[!tb]
\centering
\includegraphics[width=0.85\columnwidth]{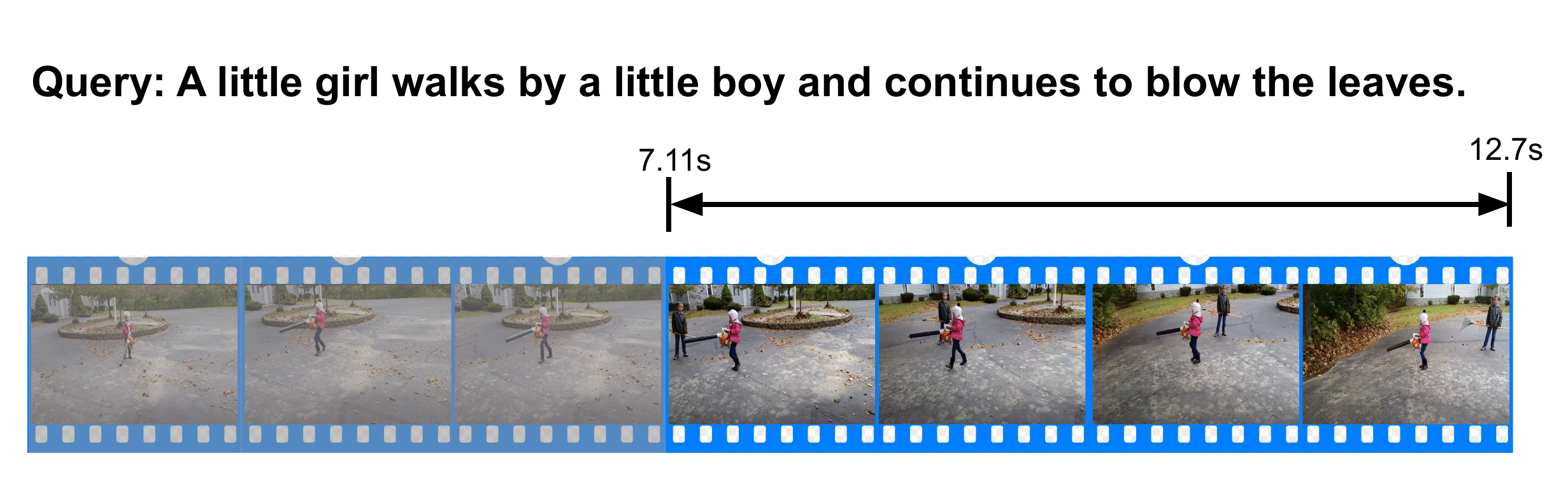}
\caption{An example of Temporal Sentence Grounding in Videos~(TSGV), \ie, to determine the start and end timestamps of the target video segment corresponding to the given sentence query.}
\label{fig:task_description}
\vspace{-0.5cm}
\end{figure}

Given the urge need in both academia and industry, a vast number of studies attempt to automatically capture the key information within a video, \eg, video summarization~\cite{ma2002user,yuan2017video,zhang2018retrospective}, video highlight detection~\cite{yao2016highlight,jiao2018three}.
More fundamentally, some works~\cite{wang2014action,karaman2014fast,singh2016multi,shou2016temporal,ma2016learning,yeung2016end,buch2019end,lin2017single} treat the task of detecting a video segment that performs a specific action as a video classification problem, denominating this type of task as \textit{action detection} or \textit{temporal action grounding~(or localization) in videos}~(TAGV). Though TAGV is able to extract effective information from the untrimmed videos to some extent, it is restricted by predefined action categories. Even the categorization is becoming more and more complicated, it is still not fully adequate to cover all kinds of interactive activities. Thus, it is natural to utilize natural language to describe those various and complex activities. Temporal Sentence Grounding in Videos~(TSGV) is such a task to match a descriptive sentence with one segment~(or moment) in an untrimmed video that is of the same semantics. As shown in Fig.~\ref{fig:task_description}, given the query ``A little girl walks by a little boy and continues to blow the leaves'' as input, the goal of TSGV is to predict the start and end points~(\ie, 7.11s to 12.7s) of the target segment within the whole video, and the predicted segment should contain the activities indicated by the input query. TSGV could serve as an intermediate task for various downstream vision-and-language tasks such as video question answering and video content retrieval. For example, related segments can be first grounded through the textual question and then analyzed for discovering the final answer to the input question. Also, by providing concise sentence summaries of videos, semantic coherent video segments can be grounded, retrieved and composed as the visual summaries of the original videos. Hence, it is worthwhile to go into a deep exploration in TSGV, which connects computer vision and natural language processing communities, as well as further promotes a variety of downstream applications. 



Compared with TAGV, TSGV is able to indicate more various and complex activities in videos via unrestricted natural language sentences, without being limited by predefined action categories. However, TSGV is much more challenging for the following reasons:
\begin{itemize}
    \item Both videos and sentence queries are temporal with rich semantic meanings. Therefore, matching the relationships between videos and sentences is quite complicated and needs to be modeled in a fine-grained manner for accurate temporal grounding.
    \item The target segments corresponded to the provided sentence queries are quite flexible in terms of spatial and temporal scales in videos. It will be computationally expensive to fetch candidate video segments of different lengths in different locations via sliding windows, followed by individually matching them with the sentence query. 
    Therefore, obtaining video segments with different temporal granularities to comprehensively cover the target segments efficiently also poses challenges for TSGV.
    \item Activities in a video often do not appear independently, instead they have internal semantic correlations and temporal dependencies on each other. Therefore, modelling the video context information, together with the inner logic relations among different video contents under the semantic guidance from sentence, 
    becomes an important and challenging step to ensure the accuracy of temporal grounding approaches.
\end{itemize}



Despite the above challenges, there exist many promising research works which bring continuous improvement in TSGV in the past few years, ranging from early two-stage matching-based methods
~\cite{gao2017tall,anne2017localizing,wu2018multi,liu2018attentive,ge2019mac}, end-to-end methods~\cite{yuan2019find,yuan2019semantic,zhang2019man,chen2020learning}, RL-based methods~\cite{wu2020tree,he2019read,hahn2019tripping},
to the recent weakly supervised setting that draws people's attention~\cite{mithun2019weakly,duan2018weakly}. 
Therefore, a systematic review for TSGV which summarizes the current works, analyzes their strengths and weaknesses, as well as promotes the future research directions becomes a necessity for the community. 
In this survey, we summarize the taxonomy of existing methods, present the evaluation protocols, critically reveal the potential problems based on the current benchmarking designs, and further identify promising research directions to promote the development of this field.


The remainder of this article is organized as follows: Sec.~\ref{2-method} gives a detailed taxonomy and analysis on the existing approaches. Sec.~\ref{3-datasets_and_metrics} reviews benchmark datasets and evaluation metrics, summarizing the current research progress via comprehensive performance comparisons. 
Sec.~\ref{4-discussion} contains a discussion of the hidden risks behind current evaluation setting and point  out promising research directions, followed by Sec.~\ref{5-conclusion} that concludes the whole paper.


\section{Method Overview}
\label{2-method}
\begin{figure}[!tb]
\centering
\includegraphics[width=0.75\columnwidth]{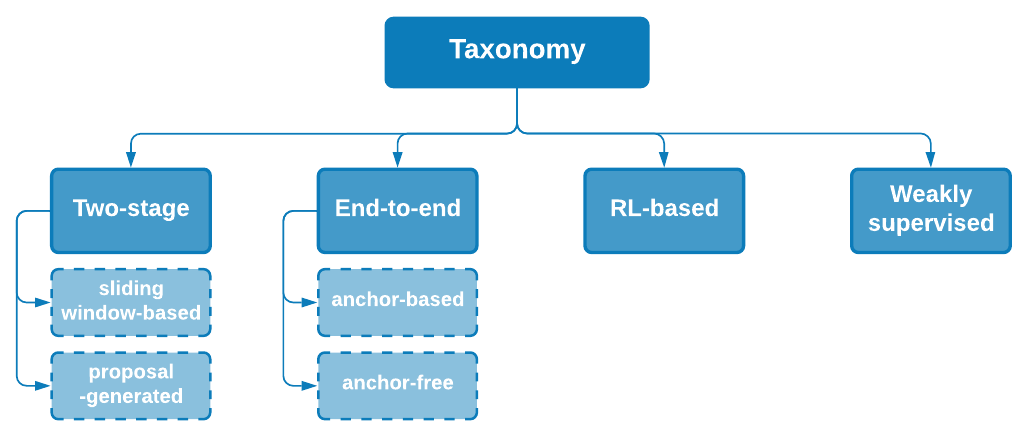}
\caption{The taxonomy of existing approaches, grouped into early two-stage methods, typical end-to-end methods, reinforcement learning~(RL)-based methods, and weakly supervised methods.}
\label{fig:taxonomy}
\vspace{-11pt}
\end{figure}

We establish the taxonomy of existing approaches based on their characteristics. As shown in Fig.~\ref{fig:taxonomy}, early works adopt a two-stage architecture, \ie, they first scan the whole video and pre-cut various candidate segments~(\ie, proposals or moments) via sliding window strategy or proposal generation network, and then rank the candidates according to the ranking scores produced by the cross-modal matching module. However, such a \textit{scan-and-localize} pipeline is time-consuming due to too much redundant computation of overlapping candidate segments, and the individual pairwise segment-query matching may also neglect the contextual video information. 

Considering the above concerns, some researchers start to solve TSGV in an end-to-end manner. It is unnecessary for such end-to-end models to pre-cut candidate moments as the inputs of the model. Instead, multi-scale candidate moments ended at each time step are maintained by LSTM sequentially or convolutional neural networks hierarchically, and such end-to-end methods are named anchor-based methods. Some other end-to-end methods predict the probabilities for each video unit~(\ie, frame-level or clip-level) being the start and end point of the target segment, or straightforwardly regress the target start and end coordinates based on the multimodal feature of the providing video and sentence query. These methods do not depend on any candidate proposal generation process, and are named anchor-free methods.

Besides, it is worth noting that some works resort to deep reinforcement learning techniques to address TSGV, taking the sentence localization problem as a sequential decision process, which are also of anchor-free. To reduce intensive labor for annotating the boundaries of groundtruth moments, weakly supervised methods with only video-level annotated descriptions have also emerged. In the following, we will present all the approaches above and perform a deep analysis of the characteristics for each type.

    \subsection{Two-stage method}
    For a two-stage method, the pre-segmenting of proposal candidates is conducted separately with the model computation. It takes the pre-segmented candidates and the sentence query as inputs of a cross-modal matching module for target segment localization. The two-stage methods can be grouped into two categories based on different ways to generate proposals.

\subsubsection{sliding window-based}
Early methods including MCN~\cite{anne2017localizing}, CTRL~\cite{gao2017tall}, ROLE~\cite{liu2018cross}, MCF~\cite{wu2018multi}, ACRN~\cite{liu2018attentive}, SLTA~\cite{jiang2019cross} and ACL-K\cite{ge2019mac}, adopt multi-scale sliding window sampling strategy for the generation of candidate proposals.

\begin{figure*}[!tb]
\centering
\subfloat[CTRL]{\includegraphics[width=0.48\columnwidth]{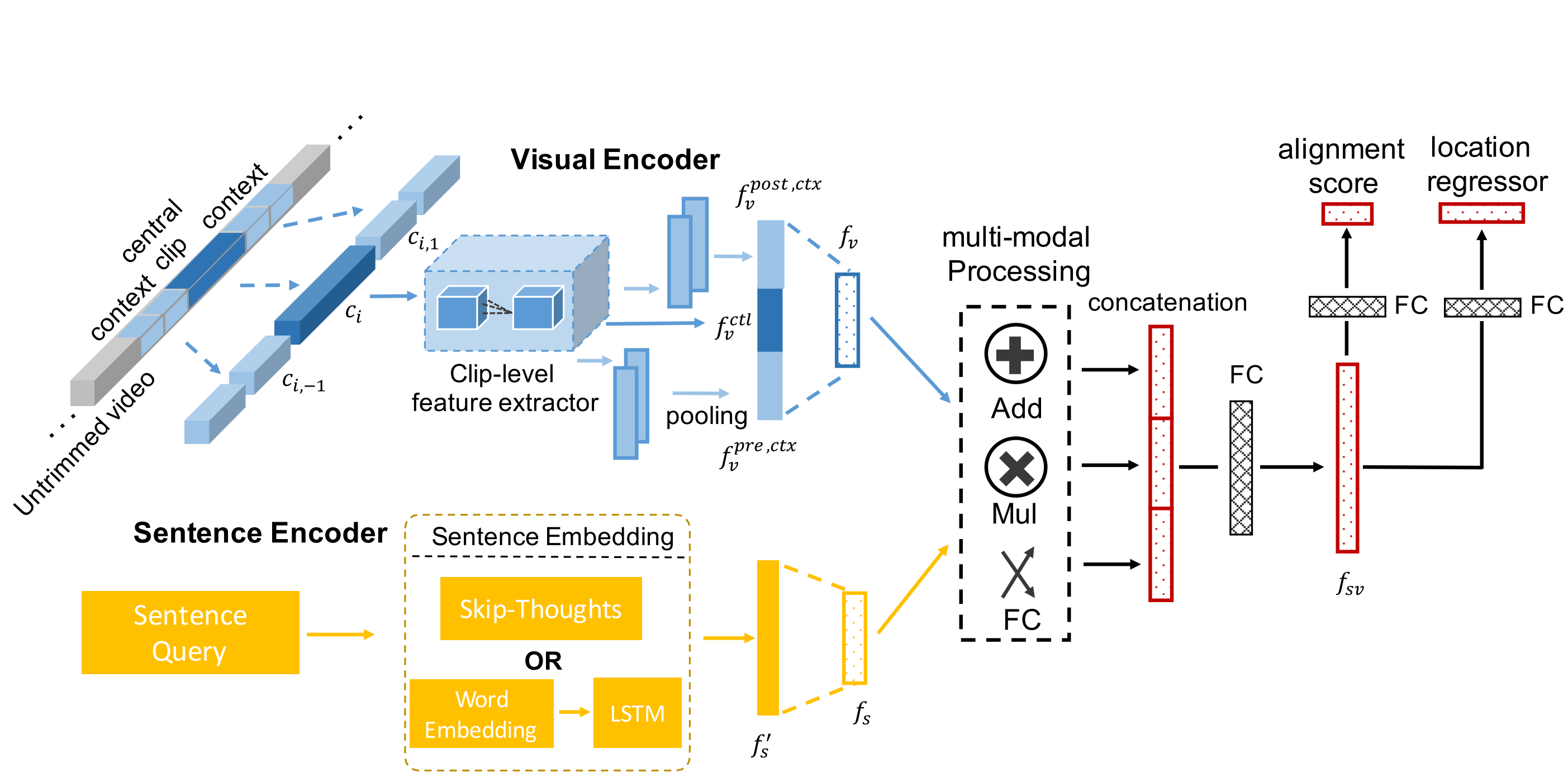}%
\label{fig:CTRL}}
\hfil
\subfloat[MCN]{\includegraphics[width=0.48\columnwidth]{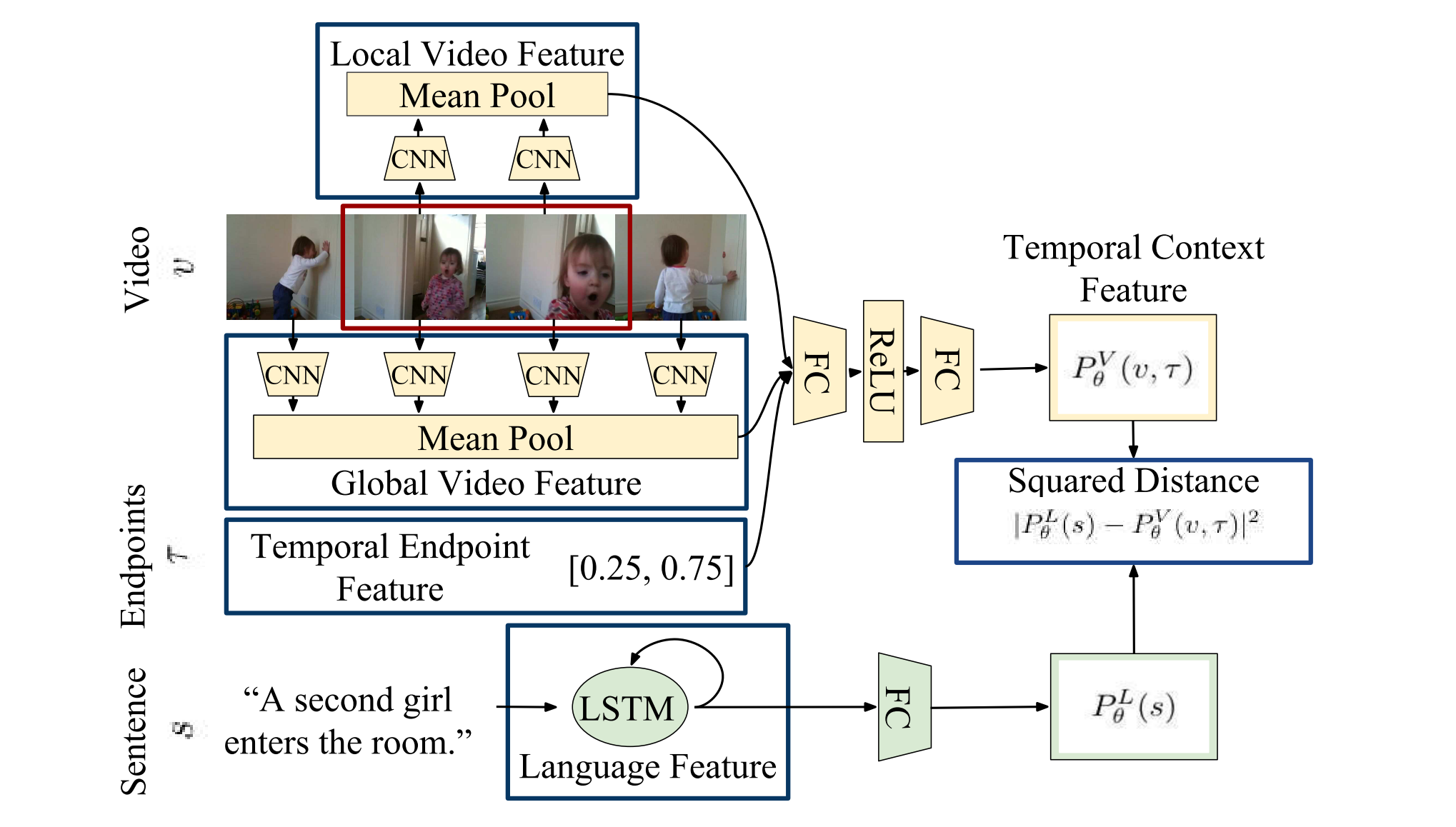}%
\label{fig:MCN}}
\caption{The Cross-modal Temporal Regression Localizer~(CTRL) and Moment Context Network~(MCN) frameworks, two pioneer works that firstly present TSGV task. CTRL uses a joint representation to get the final alignment score and refines the temporal boundaries by location regressor, while MCN tries to minimize the $\ell_2$ distance between the language and video representation vectors, figures from \cite{gao2017tall} and \cite{anne2017localizing}.}
\label{fig:CTRL_MCN}
\end{figure*}

There are two pioneering works MCN~\cite{anne2017localizing} and CTRL~\cite{gao2017tall} to define TSGV task and construct benchmark datasets. Firstly, Hendricks~\etal~\cite{anne2017localizing} propose MCN, which samples all the candidate moments (\ie segments) via sliding window mechanism, and then projects the video moment representation and query representation into a common embedding space. The $\ell_2$ distance between the sentence query and the corresponding target video moment in this space is minimized to supervise the model training~(\cf, Fig.~\ref{fig:MCN}). Specifically, MCN encourages the sentence query to be closer to the target moment than negative moments in a shared embedding space. 
Since the negative moments either come from other segments within the same video~(intra-video) or from different videos~(inter-video), MCN devises two similar but different ranking loss functions:
\begin{equation}
\begin{aligned}
\mathcal{L}_i^{intra}(\theta) &= \sum_{n \in \Gamma\setminus \tau^i}\mathcal{L}^R(D_\theta (s^i,\upsilon^i,\tau^i), D_\theta (s^i,\upsilon^i, n)) \,, \\
\mathcal{L}_i^{inter}(\theta) &= \sum_{j\neq i}\mathcal{L}^R(D_\theta (s^i,\upsilon^i,\tau^i), D_\theta (s^i,\upsilon^j,\tau^i)) \,,
\end{aligned}
\end{equation}
where $\mathcal{L}^R(x, y) = \text{max}(0, x-y+b)$, $b$ is a margin. As for training sample $i$, the intra-video ranking loss encourages sentence $i$ to be closer to the target moment at the location $\tau^i$ than the negative moments from other possible locations within the same video, while the inter-video ranking loss encourages sentence $i$ to be closer to the target one at location $\tau^i$ than the negative ones from other videos of the same location $\tau^i$. The intra-video ranking loss is able to differentiate between subtle difference within a video while the inter-video ranking loss can differentiate between broad semantic concepts.



At the same time, Gao~\etal~\cite{gao2017tall} propose CTRL, which is the first one to adapt R-CNN~\cite{girshick2014rich} methodology from object detection to the TSGV domain. Particularly, CTRL also leverages sliding window to obtain candidate segments of various lengths, and as shown in Fig.~\ref{fig:CTRL}, it exploits a multi-modal processing module to fuse the candidate segment representation with the sentence representation by three operators~(\ie, add, multiply, and full-connected layer). Then, CTRL feeds the fused representation into another fully-connected layer to predict the alignment score and location offsets between the candidate segment and the target segment. CTRL designs a multi-task loss function to train the model, including visual-semantic alignment loss and location regression loss:
\begin{align} \small
     L_{aln} &= \frac{1}{N}\sum_{i=0}^N[\alpha_c\log(1+\exp(-cs_{i,i}))   \nonumber 
     \\
     &+ \sum_{j=0,j\neq i}^N\alpha_w \log(1+\exp(cs_{i,j}))] \,, \\
     L_{reg} &= \frac{1}{N}\sum_{i=0}^N[R(t_{x,i}^*-t_{x,i})
    +R(t_{y,i}^*-t_{y,i})] \,,
\end{align}
where $L_{aln}$ is the visual-semantic alignment loss considering both aligned (video segment, query) pairs and misaligned pairs. 
$cs_{i,j}$ measures the alignment score between video segment $c_i$ and sentence $s_j$. The location regression loss $L_{reg}$ is only accounted for aligned pairs to predict the correct coordinates. $R$ is a smooth-L1 function. 

Compared to above CTRL that treats the query as a whole, Liu~\etal~\cite{liu2018cross} further make some improvements by decomposing the query and adaptively get the important textual components according to the temporal video context. 

Since CTRL overlooks the spatial-temporal information inside the moment and the query, Liu~\etal~\cite{liu2018attentive} further propose an attentive cross-modal retrieval network~(ACRN). With a memory attention network guided by the sentence query, ACRN adaptively assigns weights to the contextual moment representations for memorization to augment the moment representation. 
SLTA~\cite{jiang2019cross} also devises a spatial and language-temporal attention model to adaptively identify the relevant objects and interactions based on the query information. 


Wu and Han~\cite{wu2018multi} propose a multi-modal circulant fusion~(MCF) in contrast to the simple fusion ways employed in CTRL including element-wise product, element-wise sum, or concatenation. MCF extends the visual/textual vector to the circulant matrix, which can fully exploit the interactions of the visual and textual representations. By plugging MCF into CTRL, the grounding accuracy is further improved.

Previous works like CTRL, ACRN and MCF directly calculate the visual-semantic correlation without explicitly modelling the activity information within two modalities, and the candidate segments fairly sampled by sliding window may contain various meaningless noisy contents which do not contain any activity. Hence, Ge~\etal~\cite{ge2019mac} explicitly mine activity concepts from both visual and textual parts as prior knowledge to provide an actionness score for each candidate segment, reflecting how confident it contains activities, which enhances the localization accuracy. 

Despite the simplicity and effectiveness of such two-stage sliding window-based methods, they suffer from inefficient computation since there are too many overlapped areas re-computed due to the densely sampling process with predefined multi-scale sliding windows.

\subsubsection{proposal-generated}
Considering the inevitable drawbacks of sliding window-based methods, some approaches devote to reduce the number of proposal candidates, namely proposal-generated method. Such proposal-generated methods still adopt a two-stage scheme but avoid densely sliding window sampling through different kinds of proposal networks.

\begin{figure}[!t]
\centering
\subfloat[]{\includegraphics[width=0.48\columnwidth]{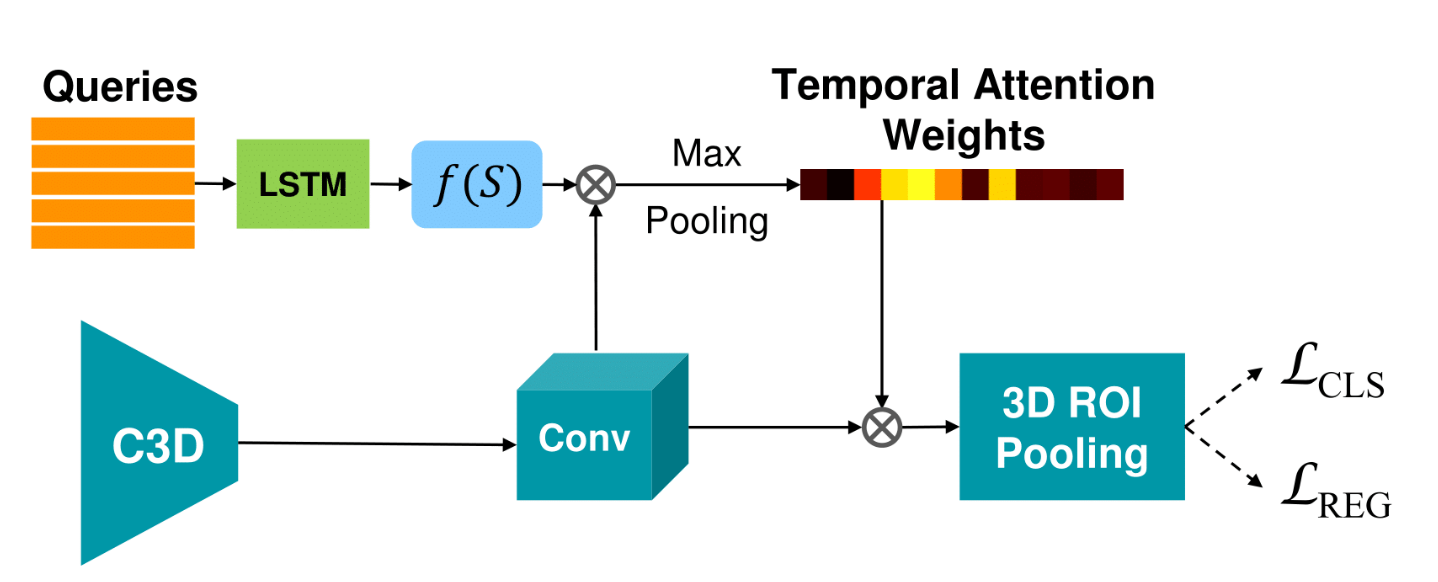}
\label{fig:QSPN-1}}
\hfil
\subfloat[]{\includegraphics[width=0.48\columnwidth]{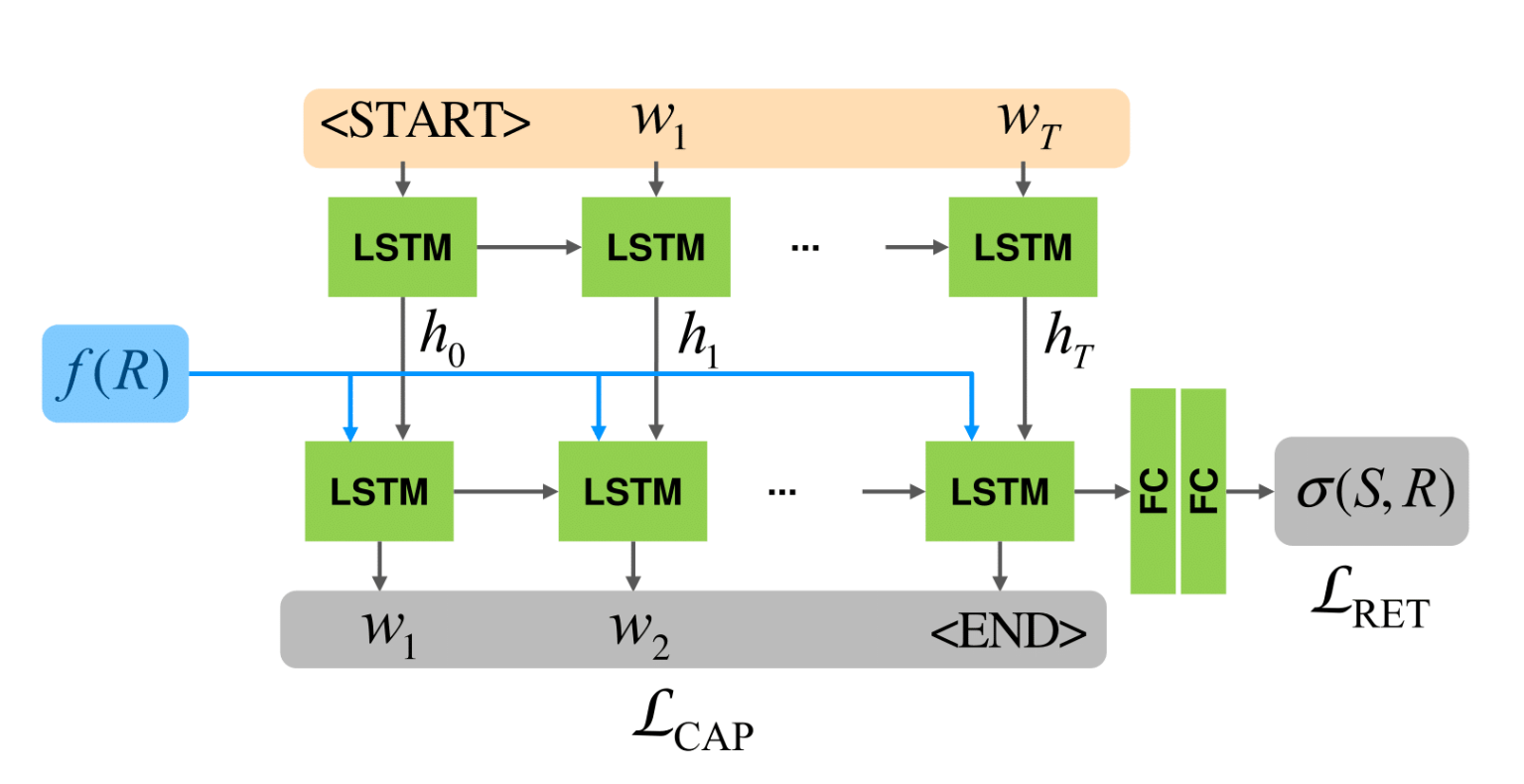}
\label{fig:QSPN-2}}
\caption{The structure of Query-guided Segment Proposal Network~(QSPN), including the query-guided segment proposal network~(\cf, \ref{fig:QSPN-1}) and a fine-grained early-fused similarity model for retrieval~(\cf, \ref{fig:QSPN-2}), figures from \cite{xu2019multilevel}.}
\label{fig:QSPN}
\vspace{-0.3cm}
\end{figure}

QSPN~\cite{xu2019multilevel} relieves such a computation burden by proposing temporal segments conditioned on the query so as to reduce the number of candidate segments~(\cf, Fig.~\ref{fig:QSPN}). Specifically, QSPN comprises of a query-guided segment proposal network~(SPN) to propose query-specific candidate segments, a fine-grained early-fused similarity model for retrieval and a multi-tasking loss combining retrieval task with an auxiliary captioning task.

As shown in Fig.~\ref{fig:QSPN-1}, the query-guided SPN first incorporates the query embeddings into the video features to get the attention weight for each temporal location, and further integrates the temporal attention weights into the convolutional process for video encoding to propose query-aware representations of candidate segments. Then as shown in Fig.~\ref{fig:QSPN-2}, the generated proposal visual feature from Fig.~\ref{fig:QSPN-1} is incorporated into the sentence embedding process at each time step of the second layer of the two-layer LSTM in a early fusion way. Then QSPN devises a triplet-based retrieval loss which is similar to MCN:

\begin{equation}
    \mathcal{L}_{RET} = \sum_{(S,R,R^\prime)}\text{max}\{0, \eta + \sigma(S,R^\prime) - \sigma(S,R)\} \,,
\end{equation}
where $(S,R)$ is the positive (sentence, segment) pair while $R^\prime$ is the sampled negative segment.
QSPN also devises an auxiliary captioning task which re-generate the query sentence from the retrieved video segment. The loss for captioning is as follows:
\begin{equation}
    \mathcal{L}_{CAP} = -\frac{1}{KT}\sum_{k=1}^K\sum_{t=1}^{T_k}\log P(w_t^k|f(R),h_{t-1}^{(2)},w_1^k,\ldots,w_{t-1}^k)\,,
\end{equation}
where a standard captioning loss is introduced to maximize the normalized log-likelihood of the words generated at all T unrolled time steps, over all K groundtruth matching sentence-segment pairs.

Similarly, SAP proposed by Chen and Jiang~\cite{chen2019semantic} integrates the semantic information of sentence queries into the generation process of activity proposals. Specifically, the visual concepts extracted from the query sentence and video frames are used to compute visual-semantic correlation score for every frame. Activity proposals are generated by grouping frames with high visual-semantic correlation scores.

Despite the success of such a two-stage pipeline, it also has some drawbacks. %
In order to achieve high localization accuracy~(\ie, the candidate pool should have at least one proposal that is close to the groundtruth moment), the duration and location distribution of the candidate moments should be diverse, thus inevitably increasing the number of candidates, which leads to inefficient computation of the subsequent matching process. 


    \subsection{End-to-end method}
The end-to-end model follows one single-pass pattern. We divide it into two types, \ie, anchor-based and anchor-free, based on whether the method uses anchors (\ie, proposals) to make predictions.

\subsubsection{anchor-based}
The representative anchor-based works include TGN~\cite{chen2018temporally}, CMIN~\cite{jiang2019cross}, SCDM~\cite{yuan2019semantic}, MAN~\cite{zhang2019man}, CBP~\cite{wang2020temporally}, CSMGAN~\cite{liu2020jointly}, 2D-TAN~\cite{zhang2020learning}, FIAN~\cite{qu2020fine}, SMIN~\cite{wang2021structured} and Zhang~\etal~\cite{zhang2021multi}.

\begin{figure}[!tb]
\centering
\includegraphics[width=0.58\textwidth]{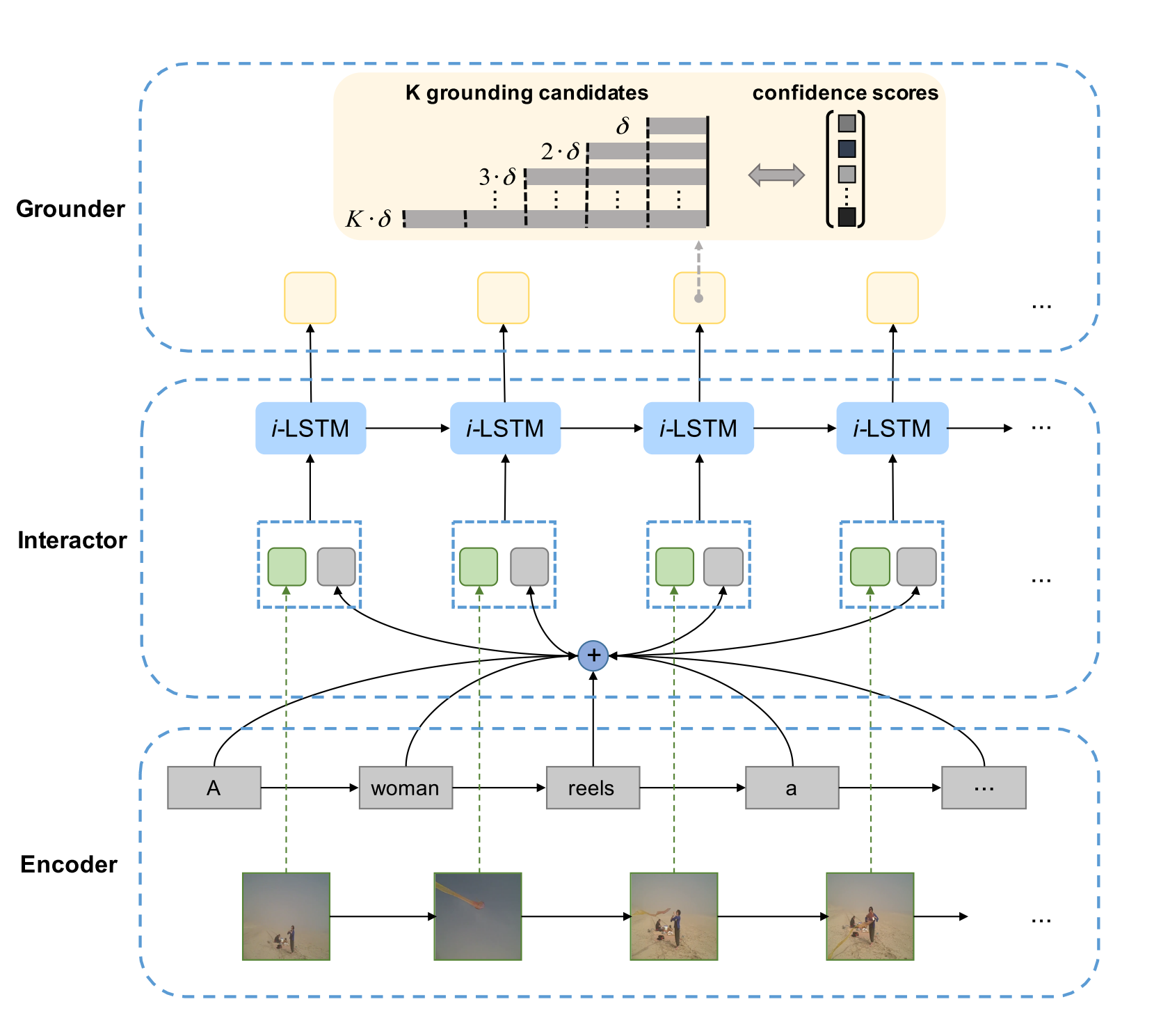}
\caption{The architecture of TGN, adopting a frame-by-word interaction single-stream framework, figure from \cite{chen2018temporally}.}
\label{fig:TGN}
\end{figure}

TGN~\cite{chen2018temporally} is one typical end-to-end deep architecture, which can localize the target moment in one single pass without handling heavily overlapped pre-segmented candidate moments. As shown in Fig.~\ref{fig:TGN}, TGN dynamically matches the sentence and video units via a sequential LSTM grounder with fine-grained frame-by-word interaction, and at each time step, the grounder would simultaneously score a group of candidate segments with different temporal scales ending at this time step.

CMIN~\cite{zhang2019cross} sequentially scores a set of candidate moments of multi-scale anchors like TGN but with a sequential BiGRU network, and refines the candidate moments with boundary regression. To further enhance the cross-modal matching, it devises a novel cross-modal interaction network~(CMIN), which first leverages a syntactic GCN to model the syntactic structure of queries, and captures long-range temporal dependencies of video context with a multi-head self-attention, then employs the fine-grained cross-modal multi-stage interaction module to produce the cross-modal features for following sequentially scoring.

Similarly, CBP~\cite{wang2020temporally} builds a single-stream model with sequential LSTM, which jointly predicts temporal anchors and boundaries at each time step for yield precise localization. Furthermore, to better detect semantic boundaries, CBP devises a self attention based module to collect contextual clues instead of simply concatenating the contextual features like \cite{gao2017tall,ge2019mac,anne2017localizing}. Based on interaction output of both language and video, it explicitly measures the contributions from different contextual elements. 

CSMGAN~\cite{liu2020jointly} also adopts such a single-pass scheme.
It builds a joint graph for modelling the cross-/self-modal relations via iterative message passing, to capture the high-order interactions between two modalities effectively. Each node of the graph aggregates the messages from its neighbor nodes in an edge-weighted manner and updates its state with both aggregated message and current state through ConvGRU.




Qu~\etal~\cite{qu2020fine} present a fine-grained iterative attention network~(FIAN), which devises a content-oriented strategy to generate candidate moments differing from the anchor-based methods with sequential RNNs mentioned above. 
FIAN employs a refined cross-modal guided attention~(CGA) block to capture the detailed cross-modal interactions, and further adopts a symmetrical iterative attention to generate both sentence-aware video and video-aware sentence representations, where the latter are explicitly facilitated to enhance the former and finally both parts contribute to a robust cross-modal feature.

\begin{figure}[!tb]
\centering
\includegraphics[width=0.8\textwidth]{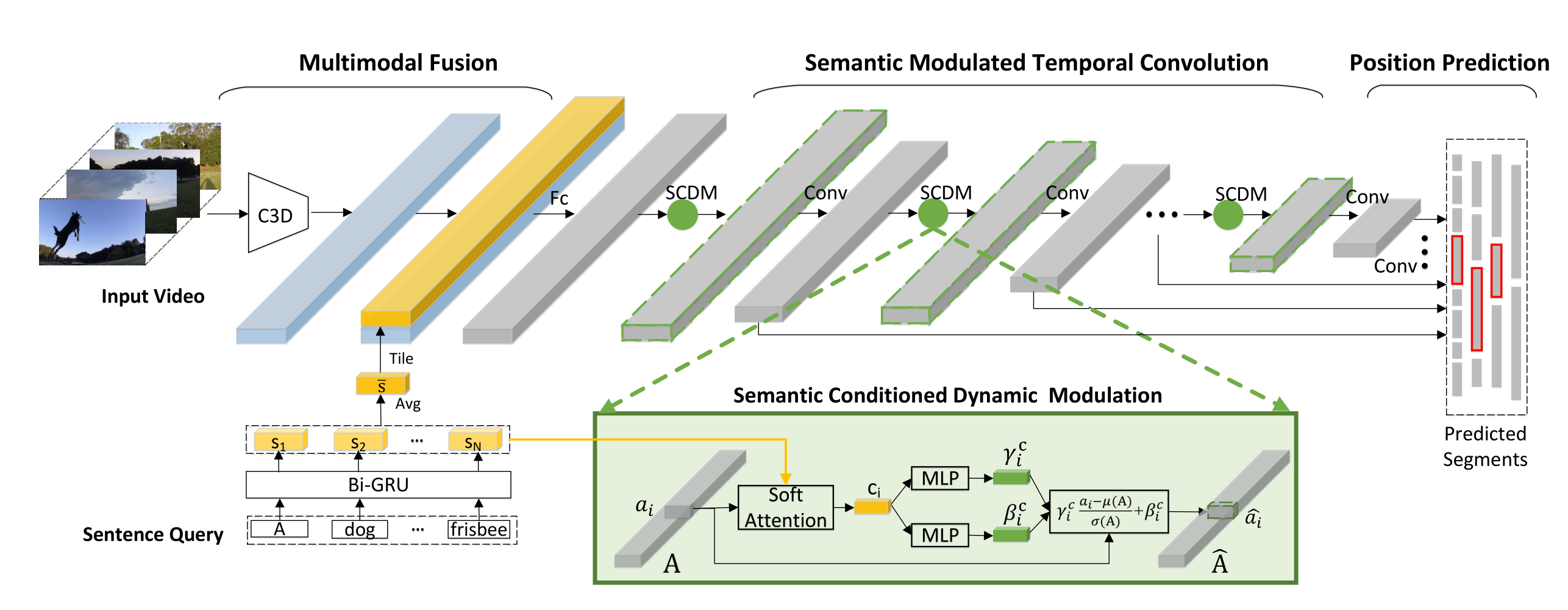}
\caption{The architecture of SCDM, which couples semantics-conditioned dynamic modulation with the temporal convolutional network, figure from \cite{yuan2019semantic}.}
\label{fig:SCDM}
\end{figure}

TGN establishes the temporal grounding architecture through a sequential LSTM network, while Yuan~\etal~\cite{yuan2019semantic} propose SCDM, which exploits a hierarchical temporal convolutional network to conduct target segment localization, and couples it with a semantics-conditioned dynamic modulation to fully leverage sentence semantics to compose the sentence-related video contents over time. As shown in Fig.~\ref{fig:SCDM}, the multimodal fusion module fuses the entire sentence and each video clip in a fine-grained manner. The fused representation is formulated as:
\begin{equation}
\mathbf{f}_t=\text{ReLU}(\mathbf{W}^f(\mathbf{v}_t||\bar{\mathbf{s}})+\mathbf{b}^f)\,.
\end{equation}
With such fused representations as inputs, the semantic modulated temporal convolution module further correlates sentence-related video contents in a temporal convolution procedure, dynamically modulating the temporal feature maps concerning the sentence. Specifically, for each temporal convolutional layer, the feature map is denoted as $\textbf{A} = \{\textbf{a}_i\}$. The feature unit $\textbf{a}_i$ will be modulated based on the modulation vectors $\gamma_i^c$ and $\beta_i^c$:
\begin{equation}
    \hat{\textbf{a}_i} = \gamma_i^c \cdot \frac{\textbf{a}_i - \mu(\textbf{A})}{\sigma(\textbf{A})} + \beta_i^c\,,
\end{equation}
where the modulation vectors are computed based on the sentence representation $\textbf{S}=\{\textbf{s}_n\}_{n=1}^N$:
\begin{equation}
    \begin{split}
        \rho_i^n &= \text{softmax}(\textbf{w}^T\text{tanh}(\textbf{W}^s \textbf{s}_n + \textbf{W}^a\textbf{a}_i + \textbf{b})) \,,
        \textbf{c}_i = \sum_{n=1}^N \rho_i^n\textbf{s}_n \,, \\
        \gamma_i^c &= \text{tanh}(\textbf{W}^{\gamma}\textbf{c}_i + \textbf{b}^{\gamma}) \,,
        \beta_i^c = \text{tanh}(\textbf{W}^{\beta}\textbf{c}_i + \textbf{b}^{\beta}) \,.\\
    \end{split}
\end{equation}
Finally, the position prediction module outputs the location offsets and overlap scores of candidate video segments based on the modulated features.

\begin{figure}[!tb]
\centering
\includegraphics[width=0.88\columnwidth]{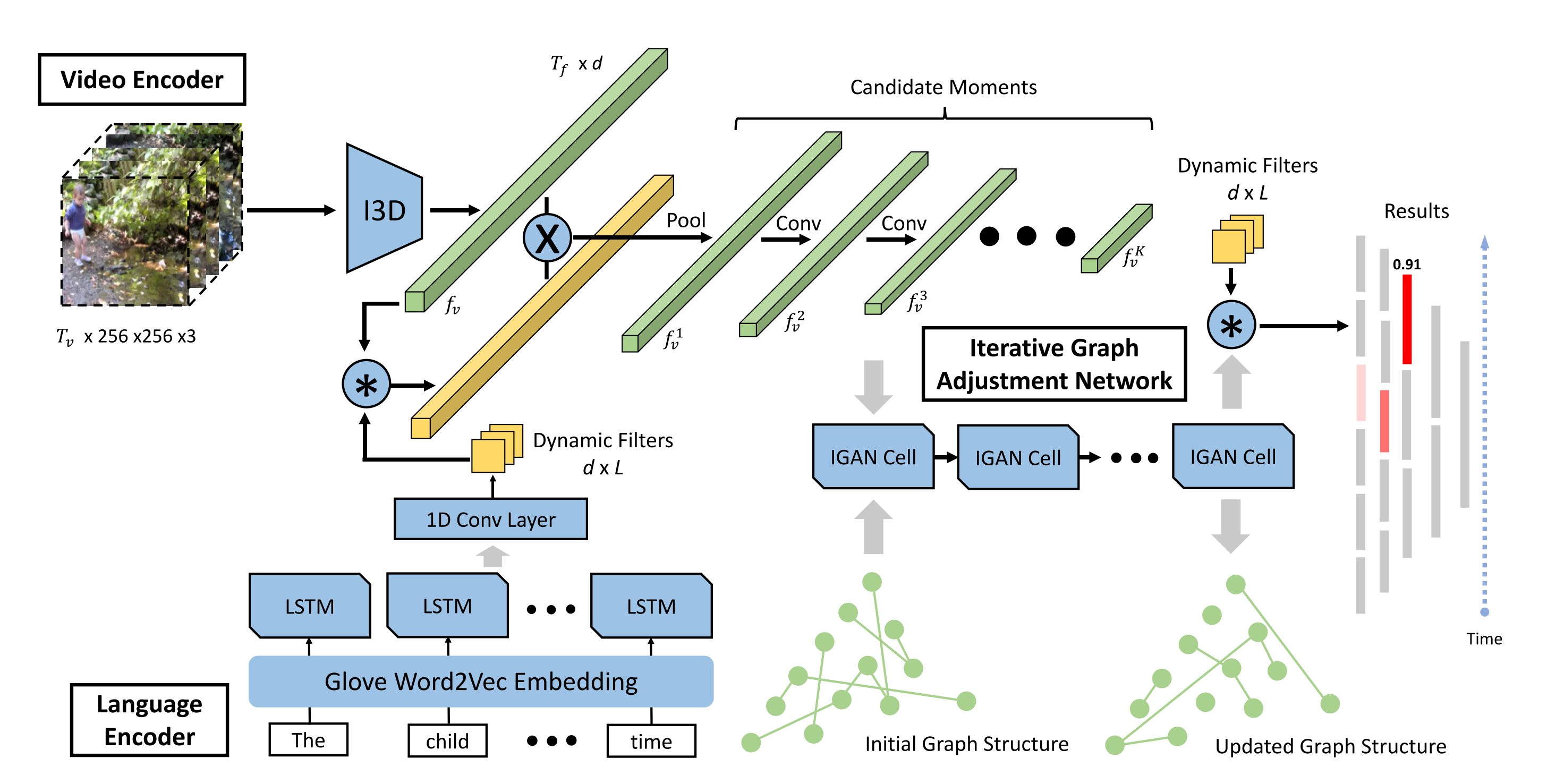}
\caption{The structure of Moment Alignment Network~(MAN). The multi-scale candidate moments are encoded by a hierarchical fully-convolutional network aligned with language semantics, and then iteratively updated by the stacked IGAN cells, figure from \cite{zhang2019man}.}
\label{fig:MAN}
\end{figure}

MAN~\cite{zhang2019man} also leverages temporal convolutional network to address the TSGV task, where the sentence query is integrated as dynamic filters into the convolutional process. Specifically, as shown in Fig.~\ref{fig:MAN}, MAN encodes the entire video stream using a hierarchical convolutional network to produce multi-scale candidate moment representations. The textual features are encoded as dynamic filters and convolved with such visual representations. Additionally, MAN exploits the graph-structured moment relation modelling adapted from Graph Convolution Network~(GCN)~\cite{kipf2016semi} for temporal reasoning to further improve the moment representations. 



\begin{figure}[!t]
\centering
\includegraphics[width=0.8\columnwidth]{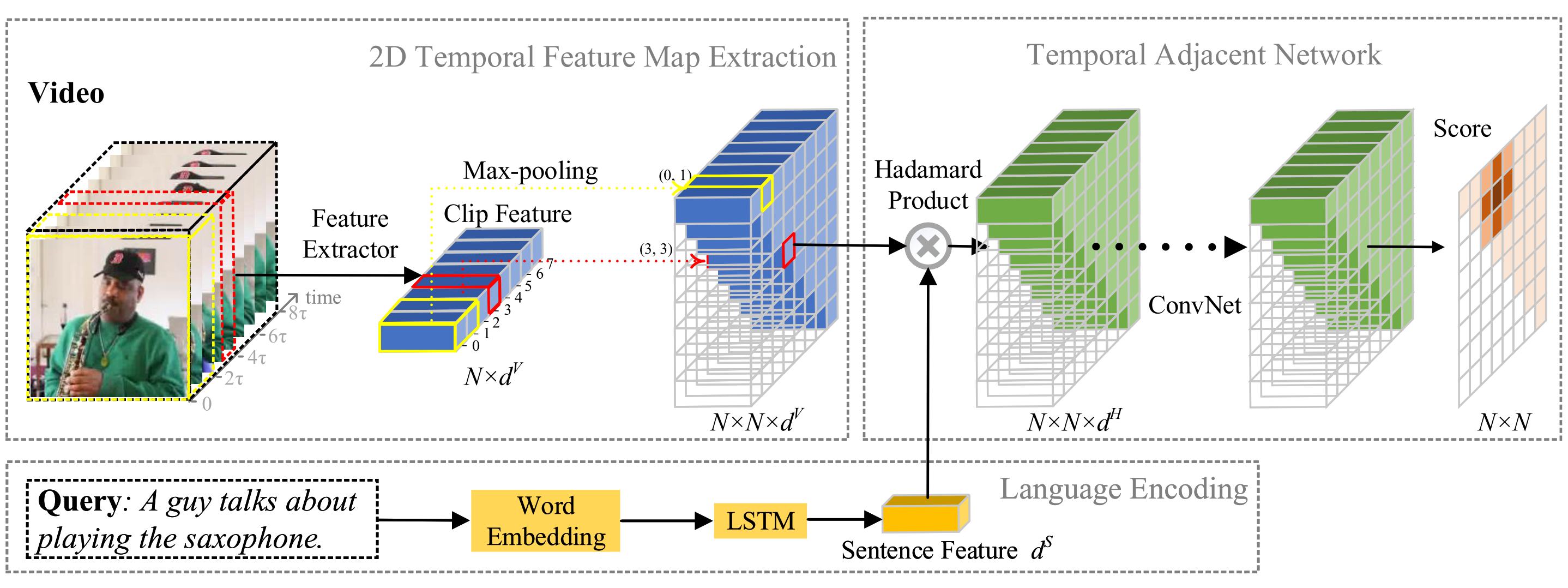}
\caption{The architecture of 2D temporal adjacent network~(2D-TAN), which consists of a text encoder for language representation, a 2D temporal feature map extractor for video representation and a temporal adjacent network for moment localization, figure from \cite{zhang2020learning}.}
\label{fig:2D-TAN}
\end{figure}

Both SCDM and MAN only consider 1D temporal feature maps, while the 2D-TAN~\cite{zhang2020learning} network models the temporal relations of video segments via a two-dimensional map. As shown in Fig.~\ref{fig:2D-TAN}, it firstly divides the video into evenly spaced video clips with duration $\tau$. The ($i,$ $j$)-th location on the 2D temporal map represents a candidate moment~(or anchor) from the time $i \tau $ to ($j$ + $1$)$\tau$. This kind of 2D temporal map covers diverse video moments with different lengths, while representing their adjacent relations. The proposed temporal adjacent network fuses the sentence representation with each of the candidate moment feature and then leverages convolutional neural network to embed the video context information, and finally predicts the confidence score of each candidate to be the final target segment. 2D-TAN adopts a binary cross entropy loss with a the scaled IoU as the supervision signal. The scaled IoU is controlled by two thresholds $t_{min}$ and $t_{max}$ as:
\begin{equation}
        y_i = \begin{cases}
                        0  &o_i \leq t_{min} \\
                        \dfrac{o_i-t_{min}}{t_{max}-t_{min}} & t_{min} < o_i < t_{max}\\
                        1 & o_i \geq t_{max}
                    \end{cases} \,,
\end{equation}
where $o_i$ is the temporal IoU between one candidate moment and the groundtruth moment. Thus, the loss function can be expressed as: 
\begin{equation}
L_{\text{2D-TAN}} = \frac{1}{C}\sum_{i=1}^C y_i \log p_i + (1-y_i) \log(1-p_i)\,,
\end{equation}
where $p_i$ is the predicted confidence score of a moment.

Wang~\etal~\cite{wang2021structured} propose a structured multi-level interaction network~(SMIN), which makes further modifications on the 2D temporal feature map as its proposal generation module. SMIN explores the inherent structure of moment, which can be disentangled into visual content and positional boundary parts for fine-grained cross-modal and intra-moment interaction. Zhang~\etal~\cite{zhang2021multi} also adopts the same proposal generation approach as that of 2D-TAN, designing a visual-language transformer backbone followed by a multi-stage aggregation module to get discriminative moment representations for more accurate moment localization.

Despite the superior performance anchor-based methods have achieved, the performance is sensitive with the heuristic rules manually designed~(\ie, the number and scales of anchors). As a result, such anchor-based methods can not adapt to the situation with variable video length. Meanwhile, although the pre-segmentation like two-stage methods is not required, it still essentially depends on the ranking of proposal candidates, which will also influence its efficiency.


\subsubsection{anchor-free}
Instead of ranking a vast number of proposal candidates, the anchor-free methods start from more fine-grained video units such as frames or clips, and aim to predict the probability for each frame/clip being the start and end point of the target segment, or directly regress the start and end points from the global view.
The typical methods include ABLR~\cite{yuan2019find}, L-Net~\cite{chen2019localizing}, LGI~\cite{mun2020local}, PMI~\cite{chen2020learning}, Rodriguez~\etal~\cite{rodriguez2020proposal}, DEBUG~\cite{lu2019debug}, GDP~\cite{chen2020rethinking}, HVTG~\cite{chen2020hierarchical}, DRN~\cite{zeng2020dense}, ExCL~\cite{ghosh2019excl}, and VSLNet~\cite{zhang2020span}.

\begin{figure}[!tb]
\centering
\includegraphics[width=0.8\textwidth]{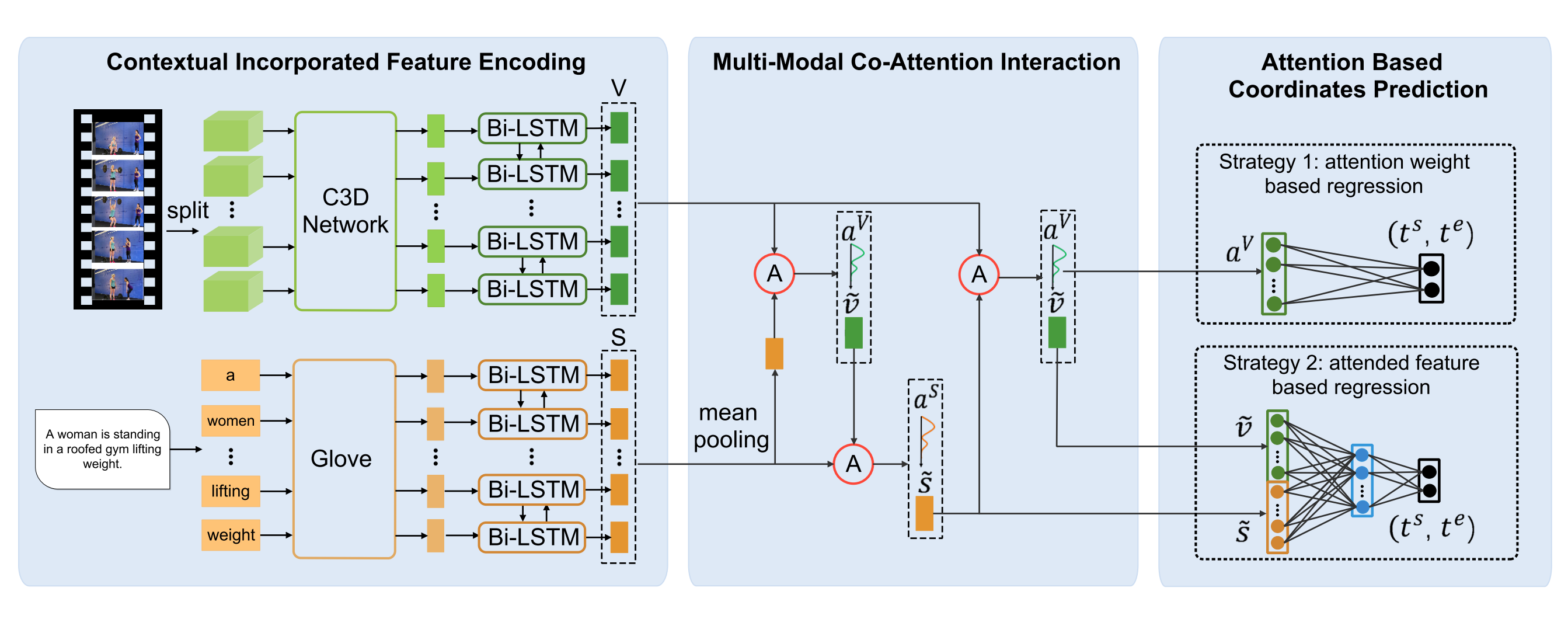}
\caption{The architecture of Attention Based Location Regression~(ABLR) model, which regresses the target coordinates with a multi-modal co-attention mechanism, figure from \cite{yuan2019find}.}
\label{fig:ABLR}
\end{figure}
Yuan~\etal propose ABLR~\cite{yuan2019find}, which solves TSGV from a global perspective without generating anchors. Specifically, as shown in Fig.~\ref{fig:ABLR}, to preserve the context information, ABLR first encodes both video and sentence via bidirectional LSTM networks. Then, a multi-modal co-attention mechanism is introduced to generate not only video attention which reflects the global video structure, but also sentence attention which highlights the crucial details for temporal localization. Finally, an attention-based coordinates prediction module is designed to regress the temporal coordinates (\ie~the starting timestamp $t^s$ and the ending timestamp $t^e$) of sentence query from the former output attentions. Meanwhile, there are two different regression strategies~(\ie, attention weight-based regression and attended feature-based regression) with the location regression loss $L_{reg}^{ablr}$: 
\begin{equation}
    L_{reg}^{ablr} =  \sum_{i=1}^K [R(\tilde{t}^{s}_i-t^{s}_i)+R(\tilde{t}^{e}_i-t^{e}_i)]\,,
\end{equation}
where $R$ is a smooth L1 function. Besides the location regression loss that aims to minimize the distance between the temporal coordinates of the predicted and the groundtruth segments, ABLR also designs an attention calibration loss $L_{cal}$ to get the video attentions more accurately:
\begin{equation}
L_{cal}=-\sum_{i=1}^K\frac{\sum_{j=1}^Mm_{i,j}\log(a_j^{V_i})}{\sum_{j=1}^Mm_{i,j}}\,.
\end{equation}
Here, $L_{cal}$ encourages the attention weights of the video clips within the groundtruth segment to be higher.


LGI~\cite{mun2020local} formulates the TSGV task as the attention-based location regression like ABLR. It further presents a more effective local-global video-text interaction module, which models the multi-level interactions between semantic phrases and video segments.

Chen~\etal~\cite{chen2020learning} propose pairwise modality interaction~(PMI) via a channel-gated modality interaction model to explicitly model the channel-level and sequence-level interactions in a pairwise fashion, which also directly predicts the boundaries. Specifically, a light-weight convolutional network is applied as the localization head to process the feature sequence and output the video-text relevance score and boundary prediction.

HVTG~\cite{chen2020hierarchical} also computes the frame-level relevance scores and make boundary prediction based on these scores. To perform the fine-grained interaction among the visual objects and between the visual object and the language query, HVTG devises a hierarchical visual-textual graph to encode the features. Objects in each video frame and words in the sentence query are considered as the graph nodes.

\begin{figure}[!tb]
\centering
\includegraphics[width=0.75\columnwidth]{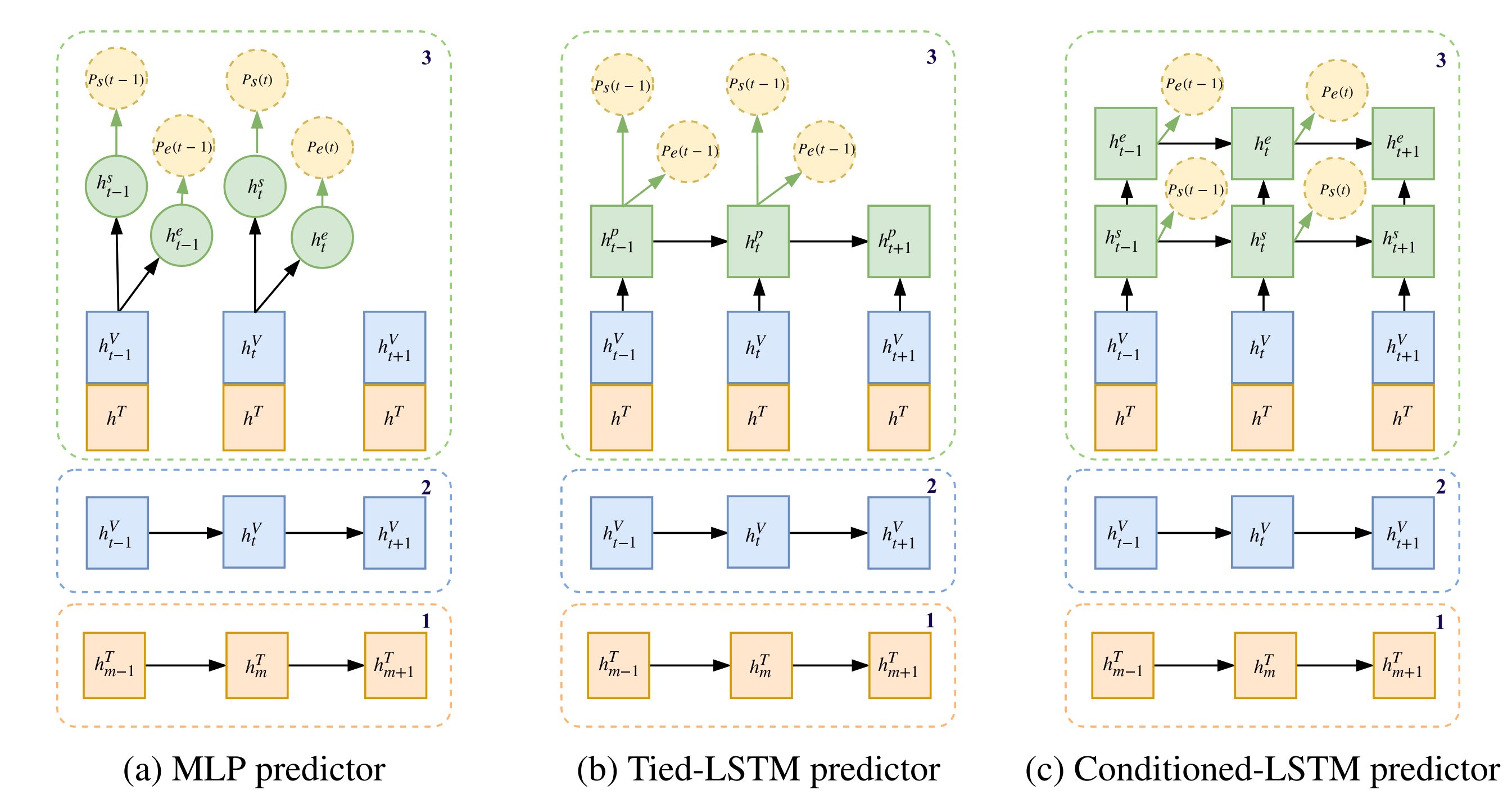}
\caption{The architecture of ExCL, consisting of three frame predictor variants, figure from \cite{ghosh2019excl}.}
\label{fig:ExCL}
\end{figure}

Unlike ABLR that regresses the coordinates of target moment directly, ExCL~\cite{ghosh2019excl} borrows the idea from the Reading Comprehension task~\cite{chen2017reading} in natural language processing area. The process of retrieving a video segment from the video is analogous to extract a text span from the passage. Specifically, as shown in Fig.~\ref{fig:ExCL}, ExCL employs three different variants of start-end frame predictor networks~(\ie, MLP, Tied-LSTM and Conditioned-LSTM) to predict start and end probabilities for each frame. The text sentence encoder~(noted in orange) and video encoder~(noted in blue) both use bidirectional LSTMs for feature encoding.
ExCL has two modes which depend on what the training objective is. ExCL-clf uses a classification loss, which is trained using negative log-likelihood loss:
\begin{equation}
    L(\theta)=-\frac{1}{N}\sum_i^N\log(P_{\text{start}}(t^i_s))+\log(P_{\text{end}}(t^i_e))\,,
\end{equation}
while ExCL-reg uses a regression loss for training, formulating start and end time prediction by computing an expectation over the probability distribution given by SoftMax outputs:
\begin{equation}
\begin{split}
    t_s&=\mathbb{E}_{P_{\text{start}}}[t] =\sum_{t_s=1}^Tt_sP_{\text{start}}(t_s) \\
    t_e&=\mathbb{E}_{P_{\text{start}}}[\mathbb{E}_{P_{\text{end}|\text{start}}}[t]] =\sum_{t_s=1}^TP_{\text{start}}(t_s)\sum_{t_e=1}^TP_{\text{end}|\text{start}}(t_e) \\
    P_{\text{end}|\text{start}}&=\text{SoftMax}(\mathbbm{1}[t_e\geq t_s]S_{\text{end}}(t))\,.
\end{split}
\end{equation}

VSLNet~\cite{zhang2020span} also employs a standard span-based Question Answering framework. VSLNet further distinguishes the differences between video sequence and text passage for better adaption to TSGV task. To address the differences, it designs a query-guided highlighting strategy to narrow down the search space to a smaller coarse highlight region.


L-Net~\cite{chen2019localizing} introduces a boundary model to predict the start and end boundaries, semantically localizing the video segment given the language query. It devises a cross-gated attended recurrent network to emphasize the relevant video parts while the irrelevant ones are gated out, and a cross-modal interactor for fine-grained interactions between two modalities.

Rodriguez~\etal~\cite{rodriguez2020proposal} also predicts start and end probabilities for each video unit. But they further model the uncertainty of boundary labels, using two Gaussian distributions as groundtruth probability distributions. The uncertainty of boundary labels results from the subjectivity of annotating process. Before the final localization, this model also adopts a dynamic filter-based guided attention mechanism to dynamically generate filters applied over video features given the sentence query, focusing on most relevant video part.

\begin{figure}[!tb]
\centering
\includegraphics[width=0.58\columnwidth]{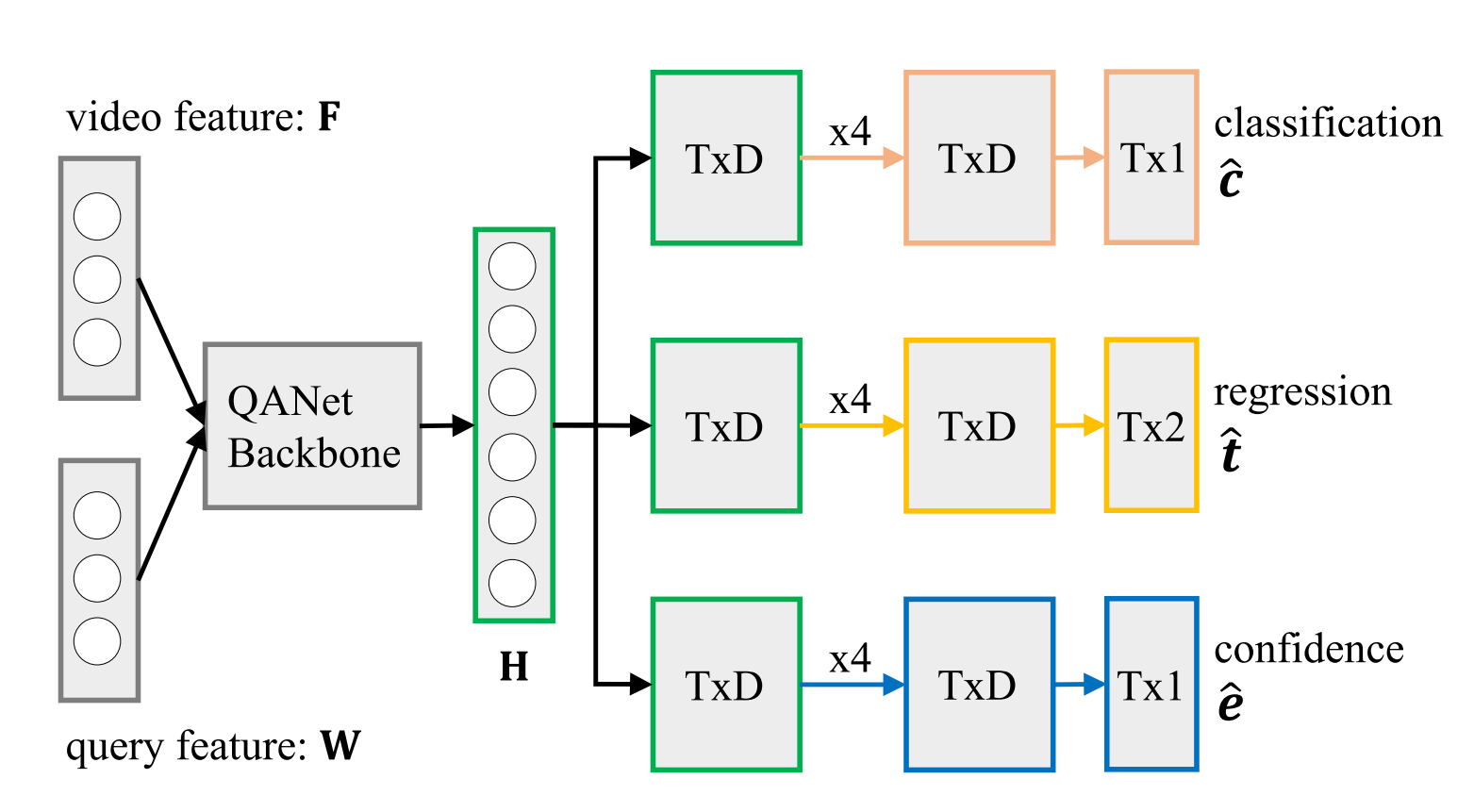}
\caption{The architecture of DEBUG, consisting of a backbone framework to model the multimodal interaction and a head module with three branches for dense regression, figure from \cite{lu2019debug}.}
\label{fig:DEBUG}
\vspace{-8pt}
\end{figure}

Lu~\etal~\cite{lu2019debug} propose a dense bottom-up grounding framework~(DEBUG), which localizes the target segment by predicting the distances to bidirectional temporal boundaries for all frames inside the groundtruth segment. In this way, all frames inside the groundtruth segment can be seen as positive samples, alleviating the severe imbalance issue caused by only regarding the groundtruth segment boundaries as positive samples. As shown in Fig.~\ref{fig:DEBUG}, a typical dense anchor-free model usually contains a backbone framework for multimodal feature encoding and a head network for frame-level predictions. Specifically, DEBUG adopts QANet as its backbone network which models the interaction between videos and queries, and designs three branches as head networks which aim to separately predict the classification score, boundary distances, and confidence score for each frame.

Similarly, DRN~\cite{zeng2020dense} and GDP~\cite{chen2020rethinking} also adopt such a dense anchor-free framework. For backbone, DRN uses a video-query interaction module to obtain fused hierarchical feature maps. For head network, DRN densely predicts the distances to boundaries, matching score and estimated IoU for each frame within the groundtruth segment.
Meanwhile, for backbone, GDP leverages a Graph-FPN layer which conducts graph convolution over all nodes in the scene space to enhance the integrated frame features. For head network, GDP predicts the distances from its location to the boundaries of target moment and a confidence score to rank its boundary prediction for each frame.

Compared with anchor-based methods, the anchor-free methods are obviously computation-efficient and robust to variable video duration. Despite these significant advantages, it is difficult for anchor-free methods to capture segment-level features for multimodal interactions.

Different from the aforementioned end-to-end methods which either samples from multi-scale anchors or directly regresses the final coordinates, some methods out of these patterns have emerged. 
The boundary proposal network~(BPNet)~\cite{xiao2021boundary} keeps the advantages of both anchor-based and anchor-free methods and avoids the defects, which generates proposals by anchor-free methods and then matches them with the sentence query in an anchor-based manner.
Wang~\etal~\cite{wang2020dual} propose a dual path interaction network~(DPIN) containing two branches~(\ie, a boundary prediction pathway for frame-level features and an alignment pathway for segment-level features) to complementarily localize the target moment.
Inspired from the dependency tree parsing task in natural language processing community, a biaffine-based architecture named context-aware biaffine localizing network~(CBLN)~\cite{liu2021context} has been proposed which can simultaneously score all possible pairs of start and end indices.

    \subsection{Reinforcement learning-based method}
    

As another kind of anchor-free approach, RL-based frameworks view such a task as a sequential decision process. The action space for each step is a set of handcraft-designed temporal transformations~(\eg, shifting, scaling). The typical methods include R-W-M~\cite{he2019read}, SM-RL~\cite{wang2019language}, TripNet~\cite{hahn2019tripping},  STRONG~\cite{cao2020strong}, TSP-PRL~\cite{wu2020tree} and AVMR~\cite{cao2020adversarial}.

\begin{figure*}[!tb]
\centering
\includegraphics[width=0.8\textwidth]{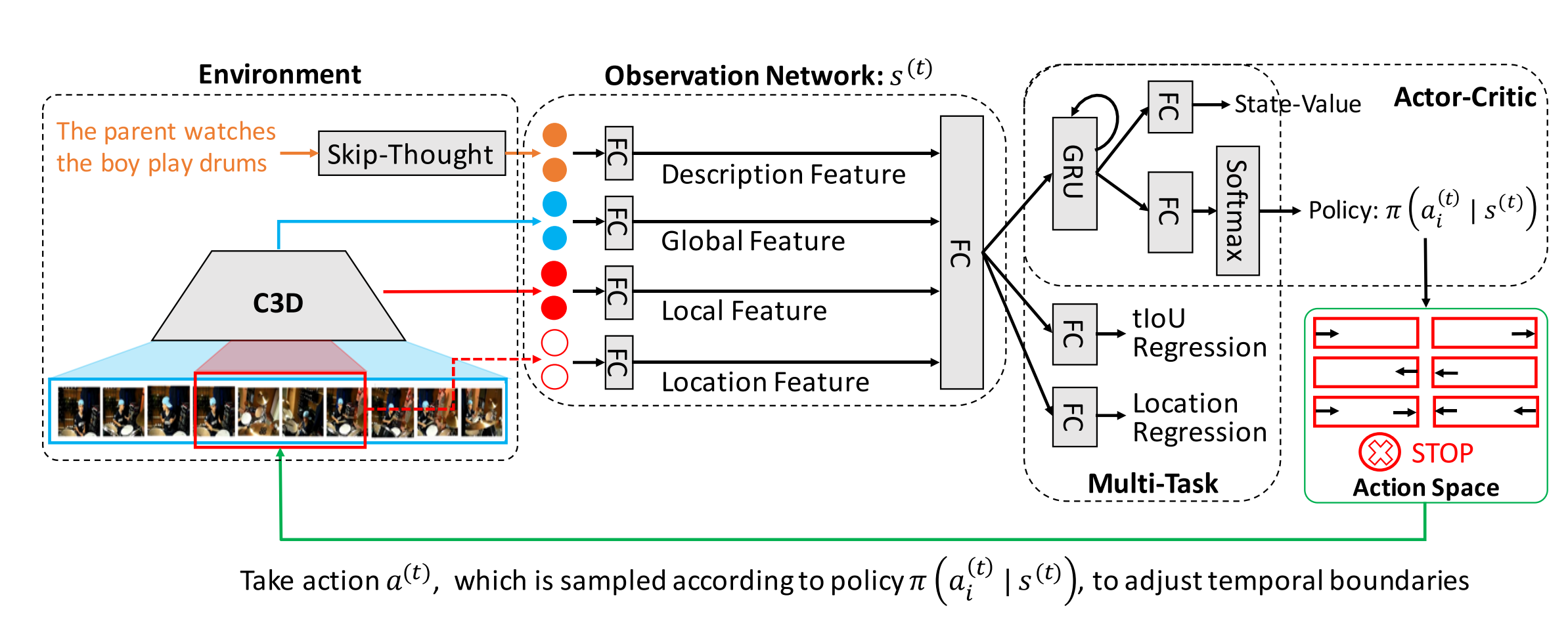}
\caption{The architecture of R-W-M framework. The action space includes 7 operators to adjust the temporal boundaries of current segment. Two regression supervised tasks are also leveraged, figure from \cite{he2019read}.}
\label{fig:RWM}
\end{figure*}

He~\etal~\cite{he2019read} first introduce deep reinforcement learning techniques to address the task of TSGV, which formulates TSGV as a sequential decision making problem. As depicted in Fig.~\ref{fig:RWM}, at each time step, the observation network outputs the current state of the environment for the actor-critic module to generate an action policy~(\ie, the probabilistic distribution of all the actions predefined in the action space), based on which the agent will perform an action to adjust the temporal boundaries. This iterative process will be ended when encountering the STOP action or reaching the maximum number of steps~(\ie, $T_{max}$).
Specifically, at each step, the current state vector is computed as:
\begin{equation}
    s^{(t)} = \Phi(E,V_G,V^{(t-1)}_L,L^{(t-1)})\,,
\end{equation}
where $s^{(t)}$ is generated by a FC layer whose inputs are the concatenated features including the segment-specific features~(\ie, the normalized boundary pair $L^{(t-1)}=[l_s^{(t-1)},l_e^{(t-1)}]$ and local segment C3D feature $V^{(t-1)}_L$) and global features~(\ie, the sentence embedding $E$ and entire video C3D feature $V_G$). Then the actor-critic module employs GRU to model the sequential decision making process. At each time step, GRU takes $s^{(t)}$ as input and the hidden state is used for policy~(denoted as $\pi(a_i^{(t)}|s^{(t)},\theta_{\pi})$) generation and state-value~(denoted as $v(s^{(t)}|\theta_v)$) estimation.
The reward for each step $r_t$ is designed to encourage a higher tIoU compared to that of the last step. The accumulated reward function is then defined as~($\gamma$ is a constant discount factor): 
\begin{equation}
    R_t = 
    \begin{cases}
    r_t + \gamma * v(s^{(t)}|\theta_v),& t=T_{max}   \\
    r_t + \gamma * R_{t+1},&   t=1,2,\ldots,T_{max}-1
    \end{cases}\,.
\end{equation}
Then they introduce the advantage function as objective which is approximated by the Mente Carlo sampling to get the policy gradient:
\begin{equation}
    \mathcal{L}^{\prime}_A(\theta_{\pi}) = - \sum_{t}(\log \pi(a_i^{(t)}|s^{(t)},\theta_{\pi}))(R_t - v(s^{(t)}|\theta_v))\,.
\end{equation}
They further leverage two supervised tasks~(\ie, tIoU regression and location regression) so the parameters can be updated from both policy gradient and supervised gradient to help the agent obtain more accurate information about the environment.

Wang~\etal~\cite{wang2019language} propose an RNN-based RL model which sequentially observes a selective set of video frames and finally obtains the temporal boundaries given the query. Cao~\etal~\cite{cao2020strong} firstly leverage the spatial scene tracking task, which utilizes a spatial-level RL for filtering out the information that is not relevant to the text query. The spatial-level RL can enhance the temporal-level RL for adjusting the temporal boundaries of the video.
TripNet~\cite{hahn2019tripping} uses gated attention to align textual and visual features, leading to improved accuracy. It incorporates a policy network for efficient search, which selects a fixed temporal bounding box moving around without watching the entire video.

\begin{figure*}[!tb]
\centering
\includegraphics[width=0.8\textwidth]{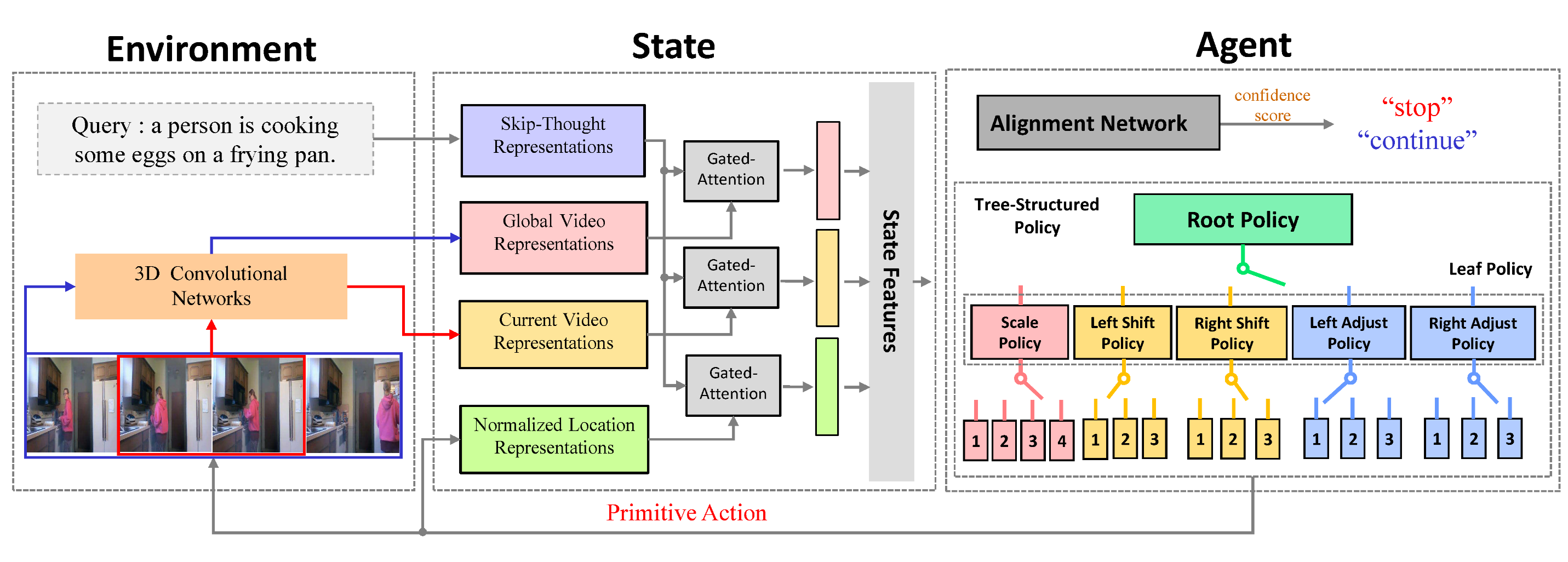}
\caption{The overall pipeline of TSP-PRL model, the action space for each step is a set of tree-structured primitive action transformations, figure from \cite{wu2020tree}.}
\label{fig:TSP-PRL}
\end{figure*}

TSP-PRL~\cite{wu2020tree} adopts a tree-structured policy that is different from conventional RL-based methods, inspired by a human's coarse-to-fine decision-making paradigm. 
As shown in Fig.~\ref{fig:TSP-PRL}, the agent receives the state from the environment~(video clips) and estimates a primitive action via tree-structured policy, including root policy and leaf policy. The action selection is depicted by a switch over the interface in the tree-structured policy. The alignment network will predict a confidence score to determine when to stop. 
Meanwhile, AVMR~\cite{cao2020adversarial} addresses TSGV under the adversarial learning paradigm, which designs a RL-based proposal generator to generate proposal candidates and employs Bayesian Personalized Ranking as a discriminator to rank these generated moment proposals in a pairwise manner.

    \subsection{Weakly supervised method}
    For the annotation of groundtruth data in TSGV, the annotators should read the query and watch the video first, and then determine the start and end points of the query-indicated segment in the video. Such a human-labored process is very time-consuming. Therefore, due to the labor-intensive groundtruth annotation procedure, some works start to extend TSGV to a weakly supervised scenario where the locations of groundtruth segments~(\ie, the start and end timestamps) are unavailable in the training stage. This is formally named as weakly supervised TSGV. 
The typical methods include WSDEC~\cite{duan2018weakly}, TGA~\cite{mithun2019weakly}, WSLLN~\cite{gao2019wslln}, SCN~\cite{lin2020weakly}, Chen~\etal~\cite{chen2020look}, VLANet~\cite{ma2020vlanet}, MARN~\cite{song2020weakly}, BAR~\cite{wu2020reinforcement}, RTBPN~\cite{zhang2020regularized}, CCL~\cite{zhang2020counterfactual}, EC-SL~\cite{chen2021towards}, LoGAN~\cite{tan2021logan} and CRM~\cite{huang2021cross}.
In general, weakly supervised methods for TSGV can be grouped into two categories~(\ie, MIL-based and reconstruction-based). One representative work will be illustrated in detail for each category, after which we will introduce the remaining. 


\begin{figure}[!tb]
\centering
\includegraphics[width=0.90\textwidth]{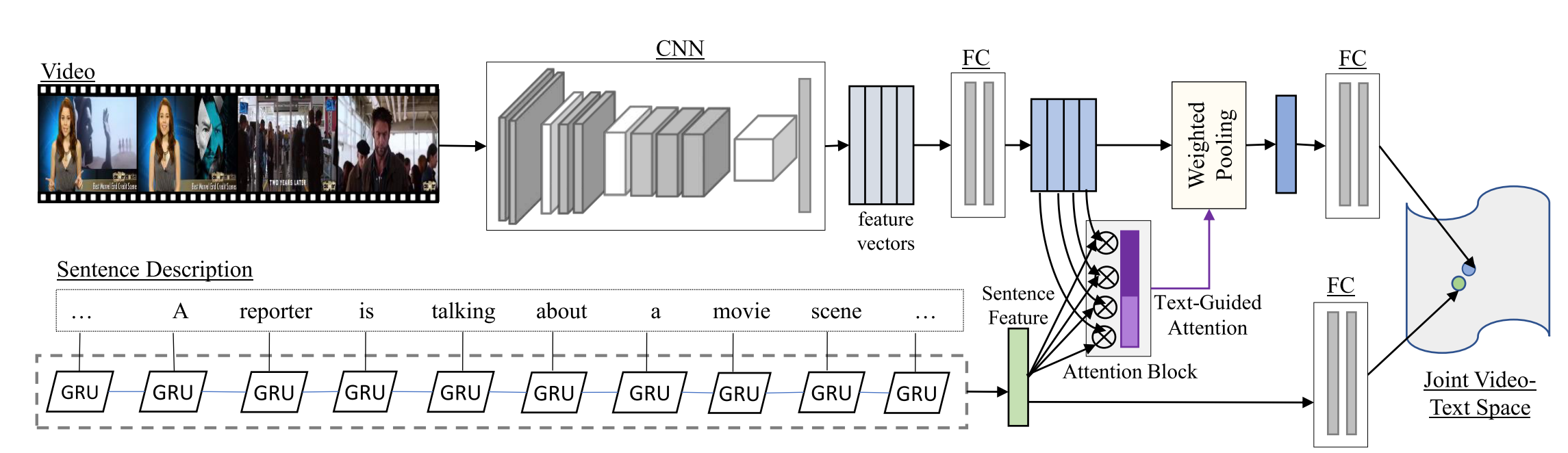}
\caption{The overall framework of TGA. It learns a joint embedding network to align the text and video features. The global video representation is generated by weighted pooling based on text-guided attentions, figure from \cite{mithun2019weakly}.}
\label{fig:TGA}
\vspace{-15pt}
\end{figure}

Some works~\cite{gao2019wslln,mithun2019weakly,tan2021logan,chen2020look} adopt multi-instance learning~(MIL) to address the weakly TSGV task. When temporal annotations are not available, the whole video is treated as a bag of instances with bag-level annotations, and the predictions for instances~(video segment proposals) are aggregated as the bag-level prediction. 

TGA~\cite{mithun2019weakly} is a typical MIL-based method which learns the visual-text alignment in the video level by maximizing the matching scores of the videos and their corresponding descriptions while minimizing the matching scores of the videos and the descriptions of others.
It presents text-guided attention~(TGA) to get text-specific global video representations, learning the joint representation of both the video and the video-level description. 
As illustrated in Fig.~\ref{fig:TGA}, TGA first employs a GRU for sentence embedding and a pretrained image encoder for extracting frame-level features. The similarity between $j^{th}$ sentence and the $k^{th}$ temporal feature within the $i^{th}$ video denoted as $s^i_{kj}$ is computed and a softmax opration is applied to get the text-guided attention weights for each temporal unit denoted as $a^i_{kj}$:
\begin{equation}
    s^i_{kj} = \frac{\mathbf{w}^{i^T}_j\mathbf{v}^i_k}{\left\|\mathbf{w}^{i}_j\right\|_2 \left\|\mathbf{v}^i_k\right\|_2} \,, 
    a^i_{kj} = \frac{\exp(s^i_{kj})}{\sum_{m=1}^{nv_i} \exp(s^i_{mj})}
    \,.
\end{equation}
Thus we could get the sentence-wise global video feature $\textbf{f}^i_j$:
\begin{equation}
    \textbf{f}^i_j = \sum_{k=1}^{nv_i} a^i_{kj}\mathbf{v}^i_k
    \,.
\end{equation}

WSLLN~\cite{gao2019wslln} is another MIL-based end-to-end weakly supervised language localization network conducting clip-sentence alignment and segment selection simultaneously. 
Huang~\etal~\cite{huang2021cross} present a cross-sentence relations mining~(CRM) method exploring the cross-sentence relations within paragraph-level scope to improve the per-sentence localization accuracy. 
A video-language alignment network~(VLANet) proposed by Ma~\etal~\cite{ma2020vlanet} prunes the irrelevant moment candidates with the Surrogate Proposal Module and utilizes multi-directional attention to get a sharper attention map for better multimodal alignment. It considers the multi-directional interactions between each surrogate proposal and query, devising the cascaded cross-modal attention~(CCA) module performing both intra- and inter-modality attention. VLANet also adopts a contrastive loss for clustering the videos and queries of the similar semantics. Wu~\etal~\cite{wu2020reinforcement} attempts to apply a RL-based model for weakly TSGV, which proposes a boundary adaptive refinement framework~(BAR) for achieving boundary-flexible and content-aware grounding results.  
Chen~\etal~\cite{chen2020look} propose a novel coarse-to-fine model based on MIL. First, the coarse stage selects a rough segment from a set of predefined sliding windows, which semantically corresponds to the given sentence. Afterwards, the fine stage mines the fine-grained matching relationship between each frame in the coarse segment and the sentence. It thereby refines the boundary of the coarse segment by grouping the frames and get a more precise grounding result.
Tan~\etal~\cite{tan2021logan} propose a Latent Graph Co-Attention Network~(LoGAN), a novel co-attention model that performs fine-grained semantic reasoning over an entire video. LoGAN is also a MIL-based method, which performs a similar frame-by-word interaction with the supervised method TGN~\cite{chen2018temporally} and adapts the graph-based method from another supervised method MAN~\cite{zhang2019man} for iterative frame representation update. 

Since MIL-based methods typically learn the visual-text alignment with a triplet loss, these methods heavily depend on the quality of randomly-selected negative samples, which are often easy to distinguish from the positive ones and cannot provide strong supervision signals.

\begin{figure*}[!tb]
\centering
\includegraphics[width=0.85\textwidth]{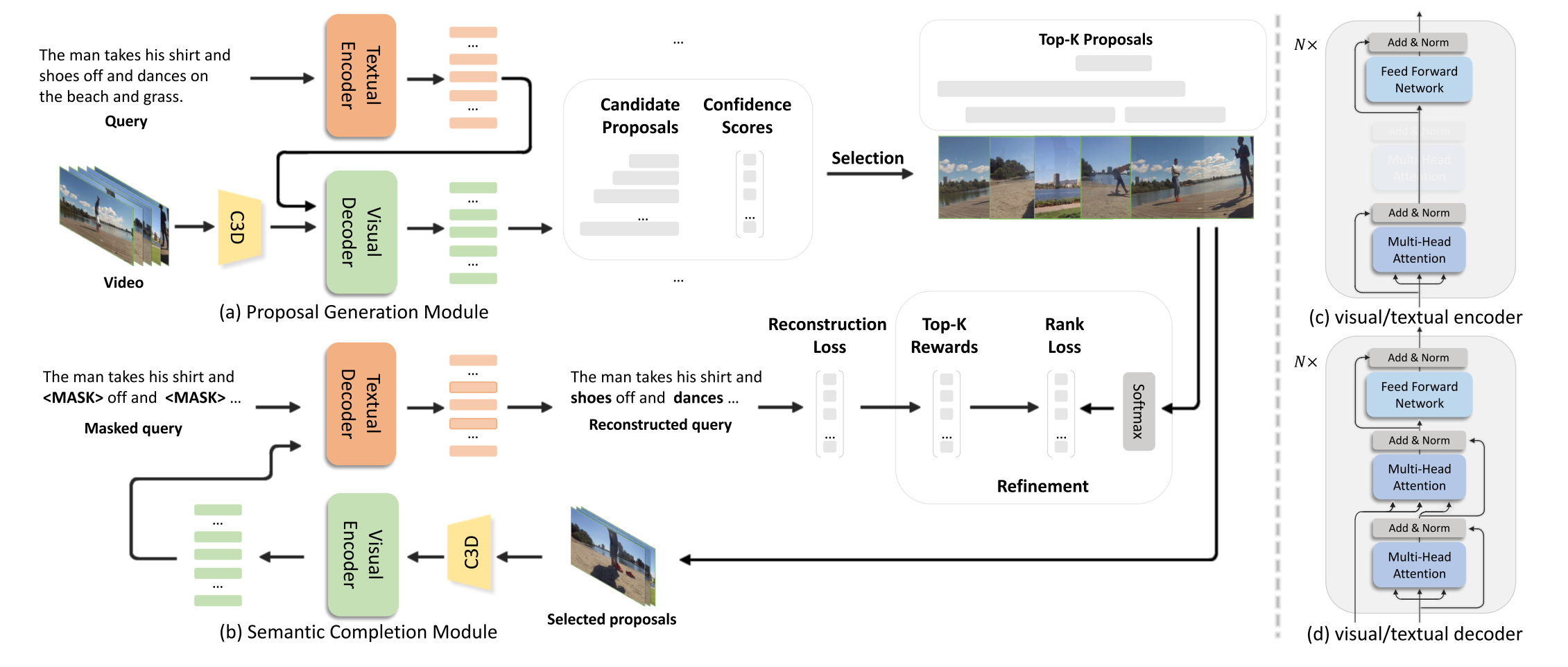}
\caption{The overall pipeline of Semantic Completion Network~(SCN), including a proposal generation module to score and select top-K proposals based on the cross-modal fusion representations and a semantic completion module to reconstruct query with masked words, figure from \cite{lin2020weakly}.}
\label{fig:SCN}
\end{figure*}

The reconstruction-based methods~\cite{lin2020weakly,duan2018weakly,song2020weakly,chen2021towards} attempt to reconstruct the given sentence query based on the selected video segments and use the intermediate results for sentence localization. Unlike MIL-based methods, the reconstruction-based methods learn the visual-textual alignment in an indirect way.
As depicted in Fig.~\ref{fig:SCN}, Lin~\etal~\cite{lin2020weakly} propose a semantic completion network~(SCN) to predict the masked important words within the query according to the visual context of generated and selected video proposals. Specifically, for each proposal $G^k$, denoted by $\hat{\mathbf{v}}^k=\{\mathbf{v}_i\}_{i=s_k}^{e_k}$, with the masked query representation $\hat{q}$, the energy word distribution $\mathbf{e}_i^k$ at $i^{th}$ time step can be computed as:
\begin{equation}
\begin{split}
\mathbf{f}^k &= \mathbf{Dec}_q(\hat{\mathbf{q}}, \mathbf{Enc}_v(\hat{\mathbf{v}}^k))\,, \\
\mathbf{e}_i^k &= \mathbf{W}_v\mathbf{f}_i^k + \mathbf{b}_v\,,
\end{split}
\end{equation}
where $\mathbf{f}^k=\{\mathbf{f}_i^k\}_{i=1}^{n_q}$ are the cross-modal semantic representations. $\mathbf{Dec}_q$ and $\mathbf{Enc}_v$ are respectively the textual decoder and visual encoder based on bi-directional Transformer~\cite{vaswani2017attention}.
Afterwards, the reconstruction loss can be computed by adding up all negative log-likelihood of masked words:
\begin{equation}
    \mathcal{L}_{rec}^k = - \sum_{i=1}^{n_q-1}\log p(\mathbf{w}_{i+1}|\hat{\mathbf{w}}_{1:i},\hat{\mathbf{v}}^k) = - \sum_{i=1}^{n_q-1}\log p(\mathbf{w}_{i+1}|\mathbf{e}_i^k)\,.
\end{equation}


Song~\etal~\cite{song2020weakly} present a Multi-Level Attentional Reconstruction Network~(MARN), which leverages the idea of attentional reconstruction. MARN uses proposal-level attentions to rank the segment candidates and refine them with clip-level attentions. 

Duan~\etal~\cite{duan2018weakly} formulate and address the problem of weakly supervised dense event captioning in videos~(\ie, to detect and describe all events of interest in a video), which is a dual problem of weakly supervised TSGV. It presents a cycle system to train the model which can solve such a pair of dual problems at the same time. In other words, weakly supervised TSGV can be 
regarded as an intermediate task in such a cycle system. 
Similar to \cite{duan2018weakly}, Chen~and~Jiang~\cite{chen2021towards} also employ a loop system for dense event captioning. They adopt a concept learner to construct an induced set of concept features to enhance the information passing between the sentence localizer and event captioner.

Besides, instead of proposing a reconstruction-based or MIL-based method, Zhang~\etal~\cite{zhang2020counterfactual} design a counterfactual contrastive learning paradigm to improve the visual-and-language grounding tasks.
A regularized two-branch proposal network~(RTBPN)~\cite{zhang2020regularized} is also presented to explore sufficient intra-sample confrontment with sharable two-branch proposal module for distinguishing the target moment from plausible negative moments.

\section{Datasets and evaluations}
\label{3-datasets_and_metrics}
In this section, we present benchmark datasets and evaluation metrics for TSGV, and provide detailed performance comparisons among the above mentioned approaches. 
    \subsection{Datasets}
    \begin{table}[!tb]
\centering
\caption{Statistics of the benchmark datasets.}
\label{table-datasets}
\begin{adjustbox}{width={\textwidth},totalheight={\textheight},keepaspectratio}%
\begin{tabular}{@{}cccccccc@{}}
\toprule
 & \# Videos & \# Moments & \# Queries & Aver. Video Duration & Aver. Query Length & Domain & Video Source \\ \midrule
DiDeMo       & 10464 & 26892 &  40543 
&   30s   &   -    & Open            & Flickr      \\
TACoS        & 127   & 7206     & 17344 & 300s &   8.86 
& Cooking         & Lab Kitchen \\
Charades-STA & 6672  & 11772 & 16124 &   30s   &   6.3    & Indoor Activity & Homes   \\
ActivityNet Captions  & 19209 & -     & 71942 & 180s & 13.48 & Open            & YouTube    \\ \bottomrule
\end{tabular}
\end{adjustbox}
\end{table}
Several datasets for TSGV from different scenarios with their distinct characteristics have been proposed in the past few years. There is no doubt that the effort of creating these datasets and designing corresponding evaluation metrics do promote the development of TSGV. Table~\ref{table-datasets} provides an overview about the statistics of public datasets, indicating the trend of involving more complicated activities and not being constrained in a narrow and specific scene~(\eg, kitchen). We will introduce them more concretely in the following. 

\textbf{DiDeMo}~\cite{anne2017localizing}.
This dataset is collected from Flickr, and consists of various human activities uploaded by personal users.
Hendricks~\etal~\cite{anne2017localizing} split and label video segments from original untrimmed videos by aggregating five-second clip units, which means the lengths of groundtruth segments are times of five seconds. They claim that this trick is for avoiding ambiguity of labeling and accelerating the validation process. 
However, such a length-fixed issue makes the retrieval task easier since it compresses the searching space into a set with limited candidates. The data split is also provided by \cite{anne2017localizing}, with 33008, 4180, and 4022 video-sentence pairs for training, validation, and test, respectively.

\textbf{TACoS}~\cite{regneri2013grounding}.
TACoS is built based on MPII-Compositive dataset~\cite{rohrbach2012script}. It contains 127 complex videos featuring cooking activities, and each video has several segments being annotated by sentence descriptions illustrating people's cooking actions. The average length of videos in TACoS is around 300s, which is much longer than that of other benchmark datasets. The total amount of sentence-segment pairs is 17,344 in this dataset, and 50\%, 25\%, 25\% of which are used for training, validation, and test, respectively.


\textbf{Charades-STA}~\cite{gao2017tall}. Charades-STA is built upon
Charades~\cite{sigurdsson2016hollywood}, which is originally collected for video activity recognition, and consists of 9848 videos depicting human daily indoor activities. Specifically, Charades contains 157 activity categories and 27,847 video-level sentence descriptions. Based on Charades, Gao~\etal~\cite{gao2017tall} construct Charades-STA with a semi-automatic pipeline, which parses the activity label out of the video description first and aligns the description with the original label-indicated temporal intervals. As such, the yielded (description, interval) pairs can be seen as the (sentence query, target segment) pairs for TSGV. Since the length of original description in Charades-STA is quite short, Gao~\etal~\cite{gao2017tall} further enhance the complexity of the description by combining consecutive descriptions into a more complex sentence for test. As a result, Charades-STA contains 13,898 sentence-segment pairs for training, 4,233 simple sentence-segment pairs~(6.3 words per sentence), and 1,378 complex sentence-segment pairs for test~(12.4 words per sentence).

\textbf{ActivityNet Captions}~\cite{krishna2017dense}.
ActivityNet Captions is originally proposed for dense video captioning, and the sentence-segment pairs in this dataset can naturally be utilized for TSGV. ActivityNet Captions contains the largest amount of videos, and it aligns videos with a series of temporally annotated sentence descriptions. On average, each of the 20k videos contains 3.65 temporally localized sentences, resulting in a total of 100k sentences. Each sentence has an average length of 13.48 words. The sentence length is also normally distributed. Since the official test set is withheld for competitions, most TSGV works merge the two available validation subsets ``val1'' and ``val2'' as the test set. In summary, there are 10,009 videos and 37,421 sentence-segment pairs in the training set, and 4,917 videos and 34,536 sentence-segment pairs in the test set.





    \subsection{Metrics}

\begin{figure}[!t]
\centering
\includegraphics[width=0.55\columnwidth]{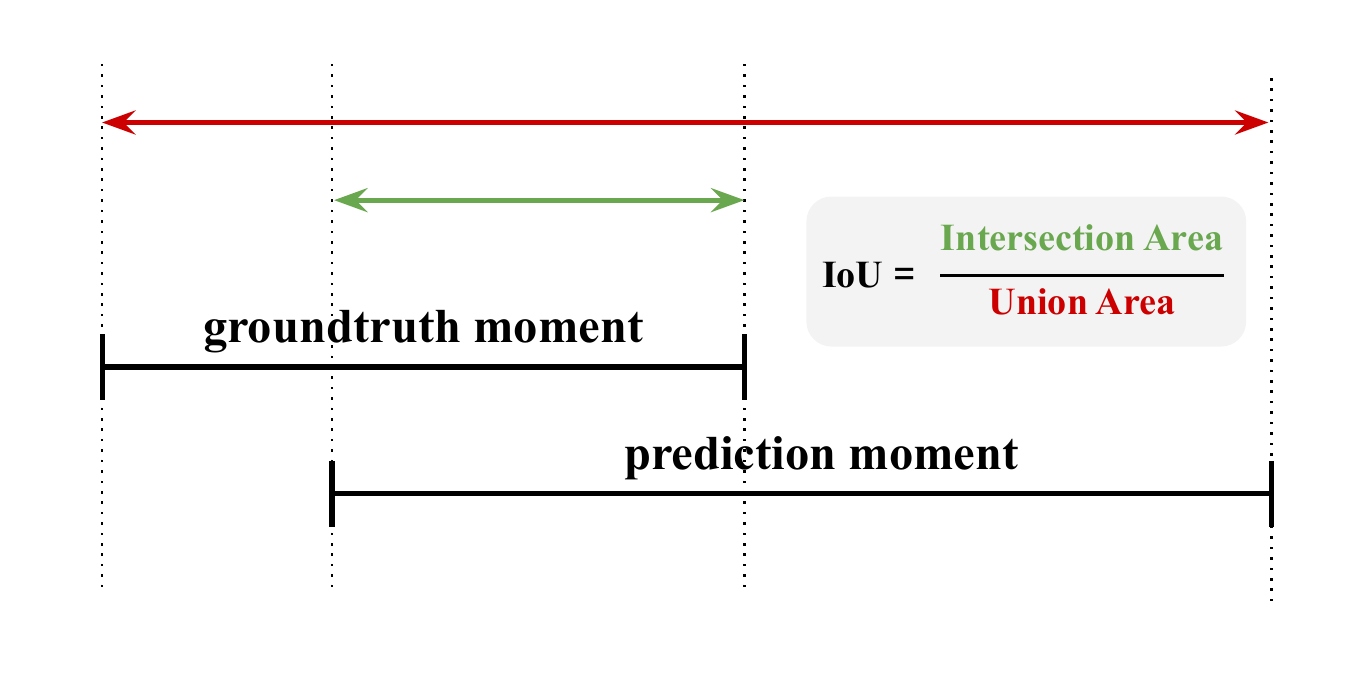}
\caption{The illustration of Temporal IoU~(Intersection over Union).}
\label{fig:IoU}
\end{figure}

There are two types of metrics for TSGV, \ie, R@$n$,IoU@$m$ and mIoU, both of which are first introduced for TSGV in \cite{gao2017tall}.
Since IoU~(Intersection over Union) is widely used in object detection to measure the similarity between two bounding boxes, similarly for TSGV, as illustrated in Fig.~\ref{fig:IoU}, many TSGV methods adopt temporal IoU to measure the similarity between the groundtruth moment and the predicted one. The ratio of intersection area over union area ranges from 0 to 1, and it will be equal to 1 when these two moments are totally overlapped.

Thereby, one of the metrics is mIoU~(\ie, mean IoU), a simple way to evaluate the results through averaging temporal IoUs of all samples. The other commonly-used metric is $\text{R@}n,\text{IoU@}m$~\cite{hu2016natural}. As for sample $i$, it is accounted as positive when there exists one segment out of top $n$ retrieved segments whose temporal IoU with the groundtruth segment is over $m$, which can be denoted as $r(n,m,q_i) = 1$. Otherwise, $r(n,m,q_i) = 0$. $\text{R@}n, \text{IoU@}m$ is the percentage of positive samples over all samples:
\begin{equation} \small \label{equa:metric}
    \text{R@}n, \text{IoU@}m = \frac{1}{N_q}\sum_i r(n,m,q_i) \,.
\end{equation}

The community is accustomed to setting $n\in\{1,5,10\}$ and $m\in\{0.3,0.5,0.7\}$. Usually, $n=1$ when the method adopts a proposal-free manner~(\ie, belongs to either anchor-free or RL-based frameworks). Moreover, it is worth noting that MCN~\cite{anne2017localizing} adopts a particular metric with the IoU threshold $m = 1$ since the groundtruth segments in DiDeMo is generated by aggregating the clip units of 5 seconds, and MCN also employs a matching-based method thus the predicted moment has chance to fully coincide with the target moment, satisfying such a extremely high IoU threshold.

    \subsection{Performance Comparison}
    \begin{table}[!tb]
\centering
\caption{The performance comparison of two-stage methods~(SW:sliding window-based, PG:proposal-generated).}
\label{two-stage-performance}
\begin{adjustbox}{width={0.95\textwidth},totalheight={\textheight},keepaspectratio}%
\begin{tabular}{@{}ccccc|ccc|ccc|ccc@{}}
\toprule
\multirow{2}{*}{Type} & \multirow{2}{*}{Method} & \multicolumn{3}{c|}{DiDeMo} & \multicolumn{3}{c|}{TACoS} & \multicolumn{3}{c|}{Charades-STA} & \multicolumn{3}{c}{ActivityNet Captions} \\ \cmidrule(l){3-14} 
 &  & 0.3 & 0.5 & 0.7 & 0.3 & 0.5 & 0.7 & 0.3 & 0.5 & 0.7 & 0.3 & 0.5 & 0.7 \\ \midrule
\multicolumn{1}{c|}{\multirow{7}{*}{SW}} & \multicolumn{1}{c|}{MCN~\cite{anne2017localizing}} & - & - & - & - & - & - & 13.57 & 4.05 & - & - & - & - \\
\multicolumn{1}{c|}{} & \multicolumn{1}{c|}{CTRL~\cite{gao2017tall}} & - & - & - & 18.32 & 13.3 & - & - & 23.63 & 8.89 & - & - & - \\
\multicolumn{1}{c|}{} & \multicolumn{1}{c|}{MCF~\cite{wu2018multi}} & - & - & - & 18.64 & 12.53 & - & - & - & - & - & - & - \\
\multicolumn{1}{c|}{} & \multicolumn{1}{c|}{ROLE~\cite{liu2018cross}} & 29.4 & 15.68 & - & - & - & - & 25.26 & 12.12 & - & - & - & - \\
\multicolumn{1}{c|}{} & \multicolumn{1}{c|}{ACRN~\cite{liu2018attentive}} & - & - & - & 19.52 & 14.62 & - & - & - & - & - & - & - \\
\multicolumn{1}{c|}{} & \multicolumn{1}{c|}{SLTA~\cite{jiang2019cross}} & - & 30.92 & 17.16 & 17.07 & 11.92 & - & 38.96 & 22.81 & 8.25 & - & - & - \\
\multicolumn{1}{c|}{} & \multicolumn{1}{c|}{ACL-K~\cite{ge2019mac}} & - & - & - & 24.17 & 20.01 & - & - & 30.48 & 12.2 & - & - & - \\ \midrule
\multicolumn{1}{c|}{\multirow{2}{*}{PG}} & \multicolumn{1}{c|}{QSPN~\cite{xu2019multilevel}} & - & - & - & - & - & - & 54.7 & 35.6 & 15.8 & 45.3 & 27.7 & 13.6 \\
\multicolumn{1}{c|}{} & \multicolumn{1}{c|}{SAP~\cite{chen2019semantic}} & - & - & - & - & 18.24 & - & - & 27.42 & 13.36 & - & - & - \\ \bottomrule
\end{tabular}
\end{adjustbox}
\end{table}

In this section, we give a thorough performance comparison of the aforementioned approaches based on four benchmark datasets. For convenience and fairness, we uniformly adopt $n = 1$ and $m \in \{0.3,0.5,0.7\}$ for the metric of R@$n$,IoU@$m$. Table~\ref{two-stage-performance} reports the experimental results of two-stage methods, Table~\ref{end-to-end-performance} is presented for end-to-end methods, Table~\ref{rl-weakly-performance} compares the performance of both RL-based and weakly supervised methods, and Table~\ref{didemo-performance} separately reports the experimental results on DiDeMo dataset with MCN-specific metrics.

\textbf{Two-stage method.} As shown in Table~\ref{two-stage-performance}, the overall performance of two-stage methods seems poorer than other approaches. The possible reasons lie in three folds: (1) Firstly, most of the two-stage methods combine video and sentence features coarsely, and neglect the fine-grained visual and textual interactions for accurate temporal sentence grounding in videos. (2) Secondly, separating the candidate segment generation and sentence-segment matching procedures will make the model unable to be globally optimized, which can also influence the overall performance. (3) Thirdly, establishing matching relationships between sentence queries and individual segments will make the local video content separate with the global video context, which may also hurt the temporal grounding accuracy.

Specifically, for the sliding window-based methods,
all the methods achieve the lowest grounding accuracy on the TACoS dataset compared to the other three datasets. The reason is that the cooking activities in TACoS take place in the same kitchen scene with only some slightly varied cooking objects (\eg~chopping board, knife, and bread). Thus, it is hard to do temporal location predictions for such fine-grained activities. Meanwhile, the lengths of videos in TACoS are also longer, which will greatly increase the target segment searching space and bring more difficulties. ACL-K outperforms the other sliding window-based methods by a large margin on the TACoS and Charades-STA datasets, proving the effectiveness of aligning the activity concepts mined from both textual and visual parts. MCN gets the most inferior results on the Charades-STA dataset, which shows that its simple multimodal matching and ranking strategy for candidate segments cannot deal well with the segments of various and flexible locations. However, CTRL, ACRN, ROLE, SLTA and ACL-K can adjust the candidate segment boundaries based on the model location offsets prediction, which can therefore improve the performances. All of the sliding window-based methods have not conducted experiments on the large-scale ActivityNet Captions dataset, which may due to the extremely expensive computation for multi-scale sliding window sampling. 

The proposal-generated methods achieve even better performance than the sliding window-based methods though the number of proposal candidates decreases. QSPN with query-guided segment proposal network and auxiliary captioning loss significantly outperforms other two-stage methods on the Charades dataset, verifying that unlike sliding window-based sampling, the presented query-guided proposal network is able to provide more effective candidate moments with finer temporal granularity. QSPN also conducts experiments on ActivityNet Captions that is comprised of richer scenes and achieves competitive results, which also proves the effectiveness of captioning supervision and query-guided proposals. Since the videos in Charades-STA dataset are of shorter lengths and contain less diverse activities, it is necessary to focus more on the metrics with higher IoU thresholds. SAP consistently outperforms other sliding-window based methods on Charades-STA with a higher IoU threshold, which attributes to its discriminative generated proposals and additional refinement process. 





\begin{table}[!tb]
\centering
\caption{The performance comparison of end-to-end frameworks~(AB:anchor-based,AF:anchor-free,OT:others).}
\label{end-to-end-performance}
\begin{adjustbox}{width={0.8\textwidth},totalheight={\textheight},keepaspectratio}%
\begin{tabular}{@{}ccccc|ccc|ccc@{}}
\toprule
\multirow{2}{*}{Type} & \multirow{2}{*}{Method} & \multicolumn{3}{c|}{TACoS} & \multicolumn{3}{c|}{Charades-STA} & \multicolumn{3}{c}{ActivityNet Captions} \\ \cmidrule(l){3-11} 
 &  & 0.3 & 0.5 & 0.7 & 0.3 & 0.5 & 0.7 & 0.3 & 0.5 & 0.7 \\ \midrule
\multicolumn{1}{c|}{\multirow{10}{*}{AB}} & \multicolumn{1}{c|}{TGN~\cite{chen2018temporally}} & 21.77 & 18.9 & - & - & - & - & 45.51 & 28.47 & - \\
\multicolumn{1}{c|}{} & \multicolumn{1}{c|}{MAN~\cite{zhang2019man}} & - & - & - & - & 46.53 & 22.72 & - & - & - \\
\multicolumn{1}{c|}{} & \multicolumn{1}{c|}{CMIN~\cite{zhang2019cross}} & 24.64 & 18.05 & - & - & - & - & 63.61 & 43.4 & 23.88 \\
\multicolumn{1}{c|}{} & \multicolumn{1}{c|}{SCDM~\cite{yuan2019semantic}} & 26.11 & 21.17 & - & - & 54.44 & 33.43 & 54.8 & 36.75 & 19.86 \\
\multicolumn{1}{c|}{} & \multicolumn{1}{c|}{CBP~\cite{wang2020temporally}} & 27.31 & 24.79 & 19.1 & - & 36.8 & 18.87 & 54.3 & 35.76 & 17.8 \\
\multicolumn{1}{c|}{} & \multicolumn{1}{c|}{2D-TAN~\cite{zhang2020learning}} & 37.29 & 25.32 & - & - & 39.7 & 23.31 & 59.45 & 44.51 & 26.54 \\
\multicolumn{1}{c|}{} & \multicolumn{1}{c|}{FIAN~\cite{qu2020fine}} & 33.87 & 28.58 & - & - & 58.55 & 37.72 & 64.1 & 47.9 & 29.81 \\
\multicolumn{1}{c|}{} & \multicolumn{1}{c|}{CSMGAN~\cite{liu2020jointly}} & 33.9 & 27.09 & - & - & - & - & 68.52 & 49.11 & 29.15 \\
\multicolumn{1}{c|}{} & \multicolumn{1}{c|}{SMIN~\cite{wang2021structured}} & 48.01 & 35.24 & - & - & 64.06 & 40.75 & - & 48.46 & 30.34 \\
\multicolumn{1}{c|}{} & \multicolumn{1}{c|}{Zhang~\etal~\cite{zhang2021multi}} & 48.79 & 37.57 & - & - & - & - & - & 48.02 & 31.78 \\ \midrule
\multicolumn{1}{c|}{\multirow{10}{*}{AF}} & \multicolumn{1}{c|}{ABLR~\cite{yuan2019find}} & 18.9 & 9.3 & - & - & - & - & 55.67 & 36.79 & - \\
\multicolumn{1}{c|}{} & \multicolumn{1}{c|}{DEBUG~\cite{lu2019debug}} & 23.45 & - & - & 54.95 & 37.39 & 17.69 & 55.91 & 39.72 & - \\
\multicolumn{1}{c|}{} & \multicolumn{1}{c|}{GDP~\cite{chen2020rethinking}} & 24.14 & - & - & 54.54 & 39.47 & 18.49 & 56.17 & 39.27 & - \\
\multicolumn{1}{c|}{} & \multicolumn{1}{c|}{PMI~\cite{chen2020learning}} &  & - & - & 55.48 & 39.73 & 19.27 & 59.69 & 38.28 & 17.83 \\
\multicolumn{1}{c|}{} & \multicolumn{1}{c|}{ExCL~\cite{ghosh2019excl}} & 44.4 & 27.8 & 14.6 & 61.4 & 41.2 & 21.3 & 62.1 & 41.6 & 23.9 \\
\multicolumn{1}{c|}{} & \multicolumn{1}{c|}{DRN~\cite{zeng2020dense}} & - & 23.17 & - & - & 45.4 & 26.4 & - & 42.49 & 22.25 \\
\multicolumn{1}{c|}{} & \multicolumn{1}{c|}{HVTG~\cite{chen2020hierarchical}} & - & - & - & 61.37 & 47.27 & 23.3 & 57.6 & 40.15 & 18.27 \\
\multicolumn{1}{c|}{} & \multicolumn{1}{c|}{Rodriguez~\etal~\cite{rodriguez2020proposal}} & 24.54 & 21.65 & 16.46 & 67.53 & 52.02 & 33.74 & 51.28 & 33.04 & 19.26 \\
\multicolumn{1}{c|}{} & \multicolumn{1}{c|}{LGI~\cite{mun2020local}} & - & - & - & 72.96 & 59.46 & 35.48 & 58.52 & 41.51 & 23.07 \\
\multicolumn{1}{c|}{} & \multicolumn{1}{c|}{VSLNet~\cite{zhang2020span}} & 29.61 & 24.27 & 20.03 & 70.46 & 54.19 & 35.22 & 63.16 & 43.22 & 26.16 \\ \midrule
\multicolumn{1}{c|}{\multirow{3}{*}{OT}} & \multicolumn{1}{c|}{BPNet~\cite{xiao2021boundary}} & 25.96 & 20.96 & 14.08 & 55.46 & 38.25 & 20.51 & 58.98 & 42.07 & 24.69 \\
\multicolumn{1}{c|}{} & \multicolumn{1}{c|}{DPIN~\cite{wang2020dual}} & 46.74 & 32.92 & - & - & 47.98 & 26.96 & 62.4 & 47.27 & 28.31 \\
\multicolumn{1}{c|}{} & \multicolumn{1}{c|}{CBLN~\cite{liu2021context}} & 38.98 & 27.65 & - & - & 61.13 & 38.22 & 66.34 & 48.12 & 27.6 \\ \bottomrule
\end{tabular}
\end{adjustbox}
\end{table}

\textbf{End-to-end method.}
For anchor-based methods, TGN achieves the lowest performance on TACoS and ActivityNet Captions datasets. CMIN also performs poorly on TACoS. The common inferior accuracy achieved by TGN, CMIN and CBP may attribute to their single-stream anchor-based localization framework. With sequential RNNs, they fail to reason complex cross-modal relations. Instead of employing RNN-based frameworks, both SCDM and MAN use convolutional neural networks to better capture fine-grained interactions and diverse video contents of different temporal granularities, which consistently achieve better performance. To make further improvement, 2D-TAN extends it to 2D feature maps to model the adjacent relations of various candidate moments of multi-anchors. SMIN and Zhang~\etal~\cite{zhang2021multi} that adopt such a similar 2D structure modelling the relationships of candidate moments, also achieve superior results out of anchor-based methods.
Specifically, Zhang~\etal~\cite{zhang2021multi} performs the best on TACoS while SMIN has surpassed other methods on Charades-STA, which also prove the effectiveness of 2D moment relationship modelling. Furthermore, CSMGAN, FIAN, SMIN and Zhang~\etal~\cite{zhang2021multi} all achieve superior results on ActivityNet Captions dataset. It is noted that although CSMGAN adopts the similar sequential RNN like TGN but it builds a joint graph for modeling the cross-/self-modal relations which can capture the high-order interactions between two modalities effectively, and FIAN employs a symmetrical iterative attention to obtain more robust cross-modal features for more accurate localization.

For anchor-free methods, reading comprehension-inspired methods including ExCL, VSLNet and Rodriguez~\etal~\cite{rodriguez2020proposal} outperform other anchor-free methods with a significant gap. Specifically, ExCL performs the best on TACoS and ActivityNet Captions dataset while VSLNet achieves the best performance on Charades-STA dataset, which proves that adopting such mature techniques in reading comprehension area for TSGV is available and effective. However, Rodriguez~\etal~\cite{rodriguez2020proposal} achieves the lowest performance on ActivityNet Captions. One possible reason is that the subjectivity of annotation is hardest to model for this challenging dataset.
The dense anchor-free methods including DRN, GDP and DEBUG outperform the early sparse regression network ABLR, justifying the importance of increasing the number of positive training samples. However, the additional regression-based methods including PMI, HVTG and LGI achieve superior performance on ActivityNet Captions dataset and LGI even performs best on Charades-STA dataset, which may result from more effective interaction between visual and textual contents. It is noted that L-Net has not been included in the table since the original paper~\cite{chen2019localizing} did not report the specific experimental values.

Additionally, other methods like BPNet, DPIN and CBLN which adopt neither anchor-based nor anchor-free achieve comparable results on three datasets. It is noted that CBLN achieves the best results out of all end-to-end methods on Charades-STA and ActivityNet Captions datasets, which quite highlights the superiority of combining the advances of both anchor-based and anchor-free and its special biaffine-based architecture.

\begin{table}[!t]
\centering
\caption{The performance comparison of RL-based and weakly supervised frameworks~(RL:RL-based,WS:weakly supervised).}
\label{rl-weakly-performance}
\begin{adjustbox}{width={0.8\textwidth},totalheight={\textheight},keepaspectratio}%
\begin{tabular}{@{}ccccc|ccc|ccc@{}}
\toprule
\multirow{2}{*}{Type} & \multirow{2}{*}{Method} & \multicolumn{3}{c|}{TACoS} & \multicolumn{3}{c|}{Charades-STA} & \multicolumn{3}{c}{ActivityNet Captions} \\ \cmidrule(l){3-11} 
 &  & 0.3 & 0.5 & 0.7 & 0.3 & 0.5 & 0.7 & 0.3 & 0.5 & 0.7 \\ \midrule
\multicolumn{1}{c|}{\multirow{6}{*}{RL}} & \multicolumn{1}{c|}{R-W-M~\cite{he2019read}} & - & - & - & - & 36.7 & - & - & 36.9 & - \\
\multicolumn{1}{c|}{} & \multicolumn{1}{c|}{SM-RL~\cite{wang2019language}} & 20.25 & 15.95 & - & - & 24.36 & 11.17 & - & - & - \\
\multicolumn{1}{c|}{} & \multicolumn{1}{c|}{TripNet~\cite{hahn2019tripping}} & - & - & - & 51.33 & 36.61 & 14.5 & 48.42 & 32.19 & 13.93 \\
\multicolumn{1}{c|}{} & \multicolumn{1}{c|}{TSP-PRL~\cite{wu2020tree}} & - & - & - & - & 45.45 & 24.75 & 56.02 & 38.82 & - \\
\multicolumn{1}{c|}{} & \multicolumn{1}{c|}{STRONG~\cite{cao2020strong}} & 72.14 & 49.73 & 18.29 & 78.1 & 50.14 & 19.3 & - & - & - \\
\multicolumn{1}{c|}{} & \multicolumn{1}{c|}{AVMR~\cite{cao2020adversarial}} & 72.16 & 49.13 & - & 77.72 & 54.59 & - & - & - & - \\ \midrule
\multicolumn{1}{c|}{\multirow{13}{*}{WS}} & \multicolumn{1}{c|}{WSDEC~\cite{duan2018weakly}} & - & - & - & - & - & - & 41.98 & 23.34 & - \\
\multicolumn{1}{c|}{} & \multicolumn{1}{c|}{TGA~\cite{mithun2019weakly}} & - & - & - & 32.14 & 19.94 & 8.84 & - & - & - \\
\multicolumn{1}{c|}{} & \multicolumn{1}{c|}{WSLLN~\cite{gao2019wslln}} & - & - & - & - & - & - & 42.8 & 22.7 & - \\
\multicolumn{1}{c|}{} & \multicolumn{1}{c|}{EC-SL~\cite{chen2021towards}} & - & - & - & - & - & - & 44.29 & 24.16 & - \\
\multicolumn{1}{c|}{} & \multicolumn{1}{c|}{SCN~\cite{lin2020weakly}} & - & - & - & 42.96 & 23.58 & 9.97 & 47.23 & 29.22 & - \\
\multicolumn{1}{c|}{} & \multicolumn{1}{c|}{Chen~\etal~\cite{chen2020look}} & - & - & - & 39.8 & 27.3 & 12.9 & 44.3 & 23.6 & - \\
\multicolumn{1}{c|}{} & \multicolumn{1}{c|}{VLANet~\cite{ma2020vlanet}} & - & - & - & 45.24 & 31.83 & 14.17 & - & - & - \\
\multicolumn{1}{c|}{} & \multicolumn{1}{c|}{MARN~\cite{song2020weakly}} & - & - & - & 48.55 & 31.94 & 14.81 & 47.01 & 29.95 & - \\
\multicolumn{1}{c|}{} & \multicolumn{1}{c|}{RTBPN~\cite{zhang2020regularized}} & - & - & - & 60.04 & 32.36 & 13.24 & 49.77 & 29.63 & - \\
\multicolumn{1}{c|}{} & \multicolumn{1}{c|}{BAR~\cite{wu2020reinforcement}} & - & - & - & 44.97 & 27.04 & 12.23 & 49.03 & 30.73 & - \\
\multicolumn{1}{c|}{} & \multicolumn{1}{c|}{CCL~\cite{zhang2020counterfactual}} & - & - & - & - & 33.21 & 15.68 & 50.12 & 31.07 & - \\
\multicolumn{1}{c|}{} & \multicolumn{1}{c|}{LoGAN~\cite{tan2021logan}} & - & - & - & 51.67 & 34.68 & 14.54 & - & - & - \\
\multicolumn{1}{c|}{} & \multicolumn{1}{c|}{CRM~\cite{huang2021cross}} & - & - & - & 53.66 & 34.76 & 16.37 & 55.26 & 32.19 & - \\ \bottomrule
\end{tabular}
\end{adjustbox}
\end{table}


\textbf{RL-based method.}
The upper part of Table~\ref{rl-weakly-performance} reports the performance of RL-based methods for TSGV. As we can see, TSP-PRL achieves promising performance on ActivityNet Captions, proving the effectiveness of borrowing the idea of the coarse-to-fine human-decision-making process. STRONG and AVMR achieves the best performance out of the RL-based frameworks on both TACoS and Charades-STA datasets, which proves the effectiveness of spatial RL for scene tracking and the employment of adversarial learning, respectively. R-W-M, TripNet and SM-RL achieve relative inferior performance. Specifically, SM-RL achieves lowest performance on Charades-STA. TripNet keeps the lowest performance on ActivityNet Captions.
Although RL-based methods can not reach the performance of end-to-end state-of-the-art methods, they offer brand-new thoughts to address the TSGV task and enhance the ability of interpretability.

\textbf{Weakly supervised method.}
The experimental results of Charades-STA and ActivityNet Captions datasets for weakly supervised methods are shown at the bottom part of Table~\ref{rl-weakly-performance}. The performance of DiDeMo for weakly supervised methods will be presented later. We cannot tell which framework~(\ie, MIL-based or reconstruction-based) has absolute advances according to the overall performance. Specifically, CRM achieves the best performance on Charades-STA and ActivityNet Captions datasets out of all weakly supervised methods. The results are also competitive compared with those of other fully supervised methods

\textbf{DiDeMo evaluation results with particular metrics.}
\begin{table}[!tb]
\centering
\caption{The evaluation results on DiDeMo~(The IoU threshold = 1).}
\label{didemo-performance}
\begin{adjustbox}{width={0.5\textwidth},totalheight={\textheight},keepaspectratio}%
\begin{tabular}{@{}c|c|ccc@{}}
\toprule
Type & Method & R@1 & R@5 & mIoU \\ \midrule
\multirow{3}{*}{Fully supervised} & TGN~\cite{chen2018temporally} & 24.28 & 71.43 & 38.62 \\
 & MCN~\cite{anne2017localizing} & 28.1 & 78.21 & 41.08 \\
 & MAN~\cite{zhang2019man} & 27.02 & 81.7 & 41.16 \\ \midrule
\multirow{5}{*}{Weakly supervised} & TGA~\cite{mithun2019weakly} & 12.19 & 39.74 & 24.92 \\
 & VLANet~\cite{ma2020vlanet} & 19.32 & 65.68 & 25.33 \\
 & WSLLN~\cite{gao2019wslln} & 19.4 & 53.1 & 25.4 \\
 & RTBPN~\cite{zhang2020regularized} & 20.79 & 60.26 & 29.81 \\
 & LoGAN~\cite{tan2021logan} & 39.2 & 64.04 & 38.28 \\ \bottomrule
\end{tabular}
\end{adjustbox}
\end{table}

As aforementioned, MCN~\cite{anne2017localizing} measures the results with the IoU threshold $m = 1$. 
Some works~\cite{chen2018temporally,zhang2019man,ma2020vlanet} also followed MCN using such metrics. We supplementally list the evaluation results~(\ie, R@1,m@1 and R@5,m@1) on DiDeMo at Table~\ref{didemo-performance} grouped by whether the method belongs to fully-supervised or weakly supervised. 
Specifically, LoGAN achieves the best performance among the weakly supervised methods while TGA~\cite{mithun2019weakly} achieves the worst. As for fully supervised methods, the performance achieved by MCN and MAN is inferior to that of TGN. 

\section{Discussions}
\label{4-discussion}
In this section, we discuss the limitations of current benchmarks and point out several promising research directions for TSGV.
Firstly, we comprehensively divide these limitations into three categories, \ie, the temporal annotation biases and ambiguous groundtruth annotations in public datasets, and the problematic evaluation metrics. These limitations may heavily mislead the TSGV approaches since each proposed method should be evaluated with these benchmarks. Meanwhile, we also present a couple of recent efforts
to address these issues with proposing new datasets/metrics or proposing new methods. Then, we point out some promising research directions of TSGV including three typical tasks, \ie, large-scale video corpus moment retrieval, spatio-temporal localization, and audio-enhanced localization.
We hope these research advances can provide more insights for future TSGV explorations, and thus further promote the development in this area.
    \subsection{Limitations of Current Benchmarks}
    \label{4-1-limitations}

Despite the promising results which have been made in TSGV, there are also some recent works~\cite{yuan2021closer,otani2020uncovering} doubting the quality of current datasets and metrics: (1) The joint distributions of starting and ending timestamps of target video segments are extremely similar in the training and test splits of current datasets. Without truly modelling the video and sentence data, and just fitting such distribution or biases in the training set, some tricky models can still achieve good results and even outperform some well-designed methods. (2) The annotation of groundtruth segment location for TSGV is ambiguous and subjective, and may influence the model evaluation. (3) Current evaluation metrics are easily deceived by the above annotation biases in current datasets, and cannot measure the model performance effectively. Since TSGV is heavily driven by these datasets and evaluation metrics, such problematic benchmarks will influence the research progress of TSGV, and further mislead this research direction. In the following, we will detail the limitations on existing datasets and evaluation metrics, and present some recent solutions to address these issues.

\begin{figure}[!t]
\centering
\includegraphics[width=\columnwidth]{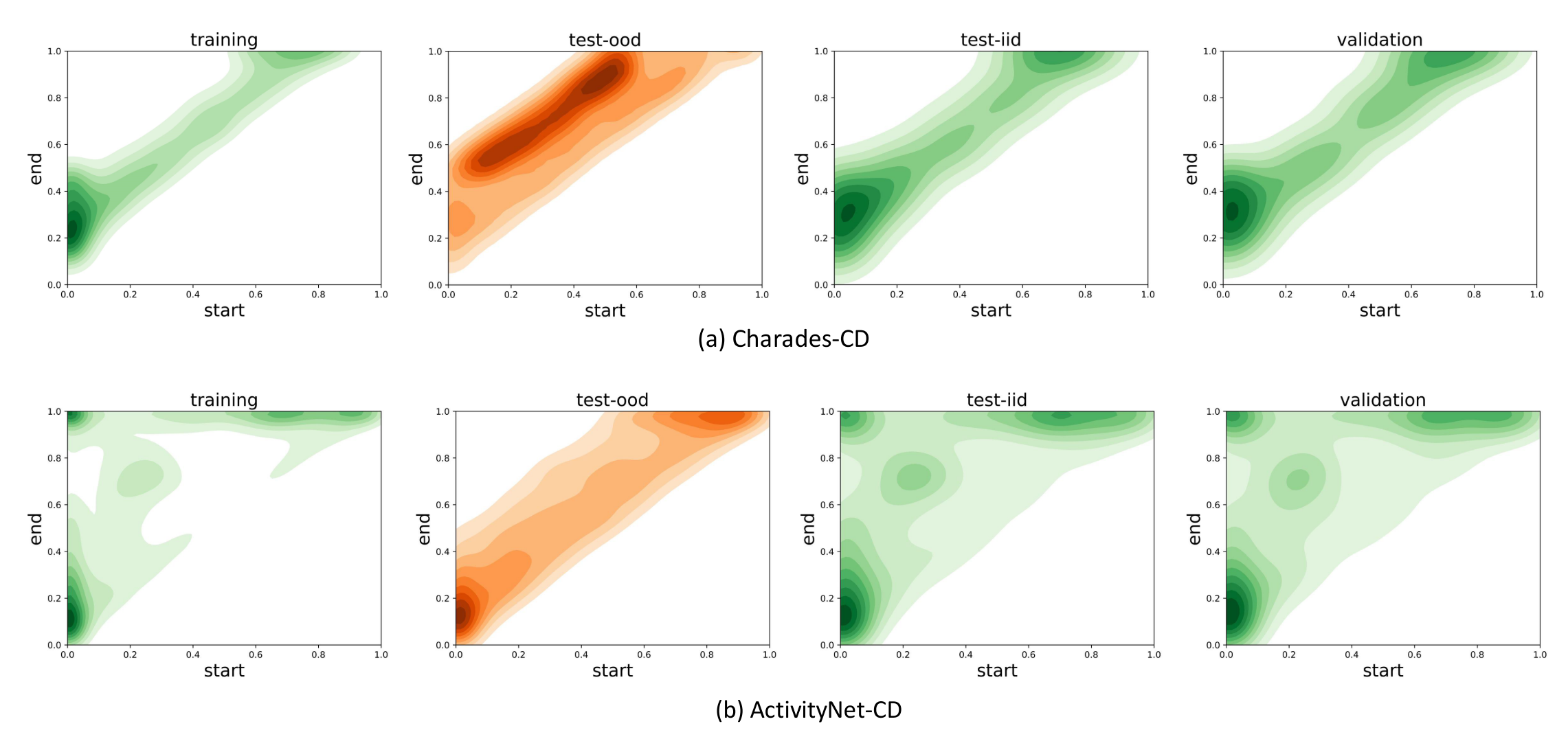}
\caption{The joint distribution of the normalized start-end timestamps for re-organized data splits~(\ie, training, test-ood, test-iid and validation) in Charades-CD~(modified from Charades-STA) and ActivityNet-CD~(modified from ActivityNet Captions), figure from \cite{yuan2021closer}.}
\label{fig:OOD-new-split}
\end{figure}


\begin{figure}[!t]
\centering
\includegraphics[width=0.9\columnwidth]{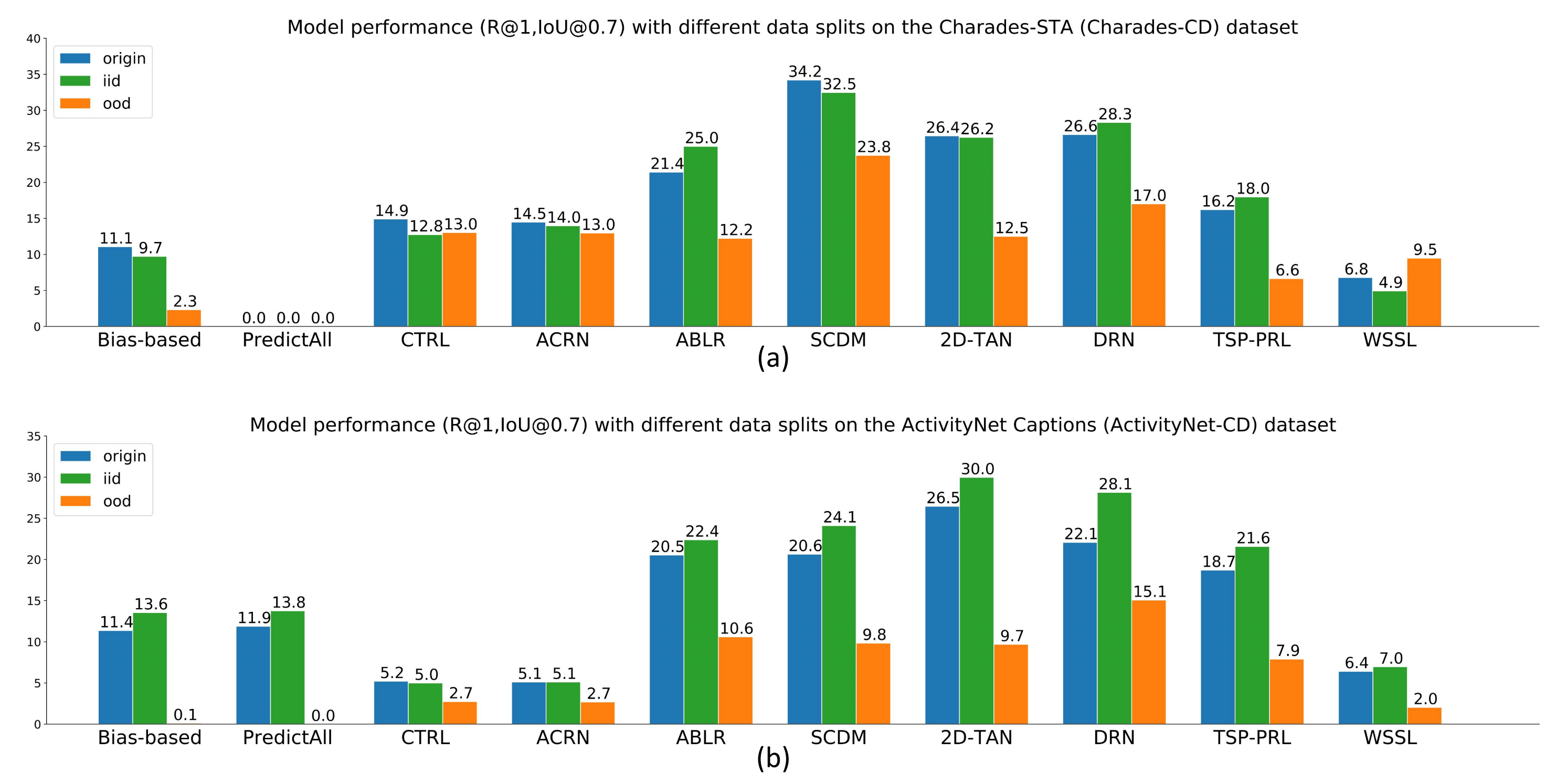}
\caption{Performance of SOTA TSGV methods on re-organized data splits, figure from \cite{yuan2021closer}.}
\label{fig:performance-drop}
\end{figure}

\textbf{Annotation distribution biases in datasets.} 
Some recent studies~\cite{yuan2021closer,otani2020uncovering} attempt to visualize the temporal location distribution of groundtruth segments, finding joint distributions of starting and ending timestamps of groundtruth segments identical in training and test sets with obvious distribution biases. They design some simple model-free methods, 
for example, a bias-based method~\cite{yuan2021closer}, which samples locations from the observed training distribution and takes them as predicted locations of target segments at inference stage. This bias-based method can achieve good performance even surpassing some well-designed deep models, without any valid visual and textual inputs. Further, as shown in Fig.~\ref{fig:OOD-new-split}, Yuan~\etal~\cite{yuan2021closer} re-organize two benchmark datasets to create two different test sets: one test set follows the identical temporal location distribution with the training set, namely test-iid, and the other test set that has quite different distribution with the training set, namely test-ood. After comparing the experimental results of various baseline methods on these two test sets, they find that for almost all methods, the performance on test-ood drops significantly~(\cf, Fig.~\ref{fig:performance-drop}), which indicates that existing methods are heavily influenced by temporal annotation biases 
and do not truly model the semantic matching relationship between videos and texts. Thus, it is crucial for future works to construct de-biased datasets and build robust models unaffected by biases. Recently, there have been some attempts to address this issue. Yang~\etal~\cite{yang2021deconfounded} design a causal-inspired framework based on CTRL and 2D-TAN, which attempts to eliminate the spurious correlation between the input and prediction caused by hidden confounder~(\ie, the temporal location of moments).


\begin{figure}[!t]
\centering
\includegraphics[width=0.85\columnwidth]{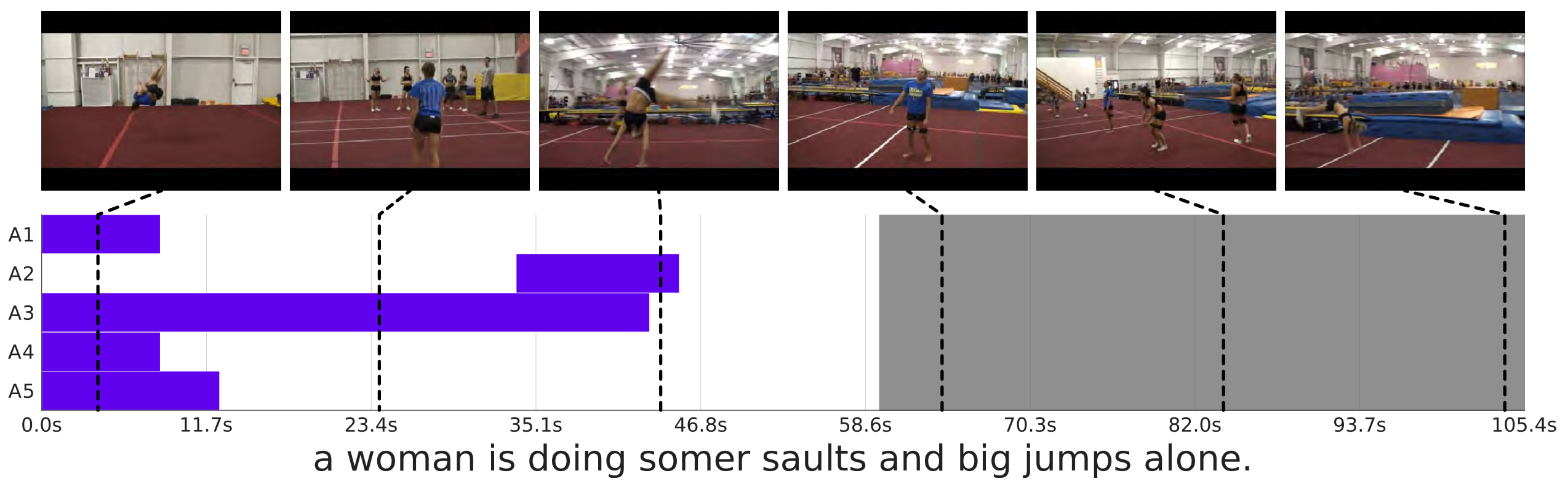}
\caption{The re-annotation example for ActivityNet Captions. Five annotators annotate five different positive segments~(shown as blue bars), all of which match the given query. While the original groundtruth segment is represented as grey bar, figure from \cite{otani2020uncovering}.}
\label{fig:annotation-ambiguity}
\end{figure}

\textbf{Ambiguity of groundtruth annotation.}
One recent study~\cite{otani2020uncovering} also mentions the ambiguous and inconsistent annotations among current TSGV datasets. Annotating the target segment location of the provided sentence query is a quite subjective task. In some cases, one query can be matched with multiple segments in videos, or different annotators will make different decisions on the grounded location of the sentence query. Therefore, only using one single groundtruth to evaluate the temporal grounding results is problematic. Otani~\etal~\cite{otani2020uncovering} suggest to re-annotate the benchmark datasets with multiple groundtruth moments for one given sentence query if exists, as shown in Fig.~\ref{fig:annotation-ambiguity}, they ask five annotators to re-annotate a video from ActivityNet Captions given the query ``a woman is doing somersaults and big jumps alone''. These five re-annotated segments corresponding to the query are totally different and do not overlap with the groundtruth segment, justifying the ambiguity and subjectivity of groundtruth annotations. They further present two alternative evaluation metrics that take multiple annotated groundtruth moments into consideration.

\textbf{Limitation of evaluation metrics.}
Besides the temporal annotation biases in current dataset, Yuan~\etal~\cite{yuan2021closer} also find that some characteristics of the datasets may have negative effects on model evaluation. Most of previous TSGV methods~\cite{chen2018temporally,liu2018cross,xu2019multilevel,yuan2019find,zhang2020learning} report their scores on some small IoU thresholds like $m\in\{0.1,0.3,0.5\}$. However, as shown in Fig.~\ref{fig:OOD-new-split}(b), for ActivityNet Captions dataset, a substantial proportion of groundtruth moments are of quite long lengths. Statistically, 40\%, 20\%, and 10\% of sentence queries refer to a moment occupying over 30\%, 50\%, and 70\% of the length of the whole video, respectively. Such annotation biases can obviously increase the chance of correct prediction under small IoU thresholds. Taking an extreme case as example, if the groundtruth moment is the whole video, any predictions with duration longer than $0.3$ can achieve R@1,IoU@0.3=1. Thus, the metric $\text{R@n},\text{IoU@}m$ with small $m$ is unreliable for current biased annotated datasets.
Therefore, to alleviate the above effects, they present a new metric namely $\text{discounted-R@}n,\text{IoU@}m$. This new metric considers that the hit score~(\ie, $r(n,m,q_i)$) for each positive sample $i$ should not be limited to $\{0,1\}$. It can be a real number $\in[0,1]$ depending on the relative distances between the predicted and groundtruth boundaries. The formal definition for each sample $i$ is as follows:
\begin{equation}
r(n,m,q_i) = 
(1-\text{nDis}(g^s_i,p^s_i)) \times (1-\text{nDis}(g^e_i,p^e_i)) \,,
\end{equation}
where the nDis operation calculates the distance between the groundtruth and predicted boundaries normalized to $[0,1]$ by the video length. $(g_i^s, g_i^e)$/$(p_i^s, p_i^e)$ indicates the (start,end) timestamps of the groundtruth/predicted segment for sample $i$. The remaining computation for $\text{discounted-R@}n,\text{IoU@}m$ is the same with $\text{R@}n,\text{IoU@}m$~(\cf Equation~\ref{equa:metric}). 

    \subsection{Promising Research Directions}
    \label{4-2-directions}
    We point out some promising research directions, including three TSGV-related tasks based on TSGV. 
        \subsubsection{Large-scale video corpus moment retrieval}
        Large-scale video corpus moment retrieval~(VCMR) is a research direction extended from TSGV that has been explored over the past few years~\cite{escorcia2019temporal,lei2020tvr,zhang2020hierarchical,zhang2021video}.
It has more application value since it can retrieve the target segment semantically corresponding to a given text query from a large-scale video corpus~(\ie, a collection of untrimmed and unsegmented videos) rather than from a single video. As compared with TSGV, VCMR has higher efficiency requirements since it not only needs to retrieve a specific segment from one single video but also locates the target video from a video corpus. 

Escorcia~\etal~\cite{escorcia2019temporal} first extend TSGV to VCMR, introducing a model named Clip Alignment with Language~(CAL) to 
align the query feature with a sequence of uniformly partitioned clips for moment composing. Lei~\etal~\cite{lei2020tvr} introduce a new dataset for VCMR called TVR, which is comprised of videos and their associated subtitle texts. A Cross-modal Moment Localization~(XML) network with a novel convolutional start-end detector module is also proposed to produce moment predictions in a late fusion manner. Zhang~\etal~\cite{zhang2020hierarchical} present a hierarchical multi-modal encoder~(HAMMER) to capture both coarse- and fine-grained semantic information from the videos and train the model with three sub-tasks~(\ie, video retrieval, segment temporal localization, and masked language modeling). Zhang~\etal~\cite{zhang2021video} introduce contrastive learning for VCMR, designing a retrieval and localization network with contrastive learning~(ReLoCLNet). %
        \subsubsection{Spatio-temporal localization}
Spatial-temporal sentence grounding in videos is another extension from TSGV which mainly localizes the referring object/instance as a continuing spatial-temporal tube~(\ie, a sequence of bounding boxes) extracted from an untrimmed video via a natural language description.
Since fine-grained labeling process of localizing a tube~(\ie, annotate a spatial region for each frame in videos) for STSGV is labor-intensive and complicated, Chen~\etal~\cite{chen2019weakly} propose to solve this task in a weakly-supervised manner which only needs video-level descriptions, with a newly-constructed VID-sentence dataset.  
Besides, VOGNet~\cite{sadhu2020video} commits to address the task of video object grounding, which grounds objects in videos referred to the natural language descriptions, and constructs a new dataset called ActivityNet-SRL. Tang~\etal~\cite{tang2021human} employ visual transformer to solve a similar task which aims to localize a spatio-temporal tube of the target person from an untrimmed video based on a given textural description with a newly-constructed HC-STVG dataset.

        \subsubsection{Audio-enhanced localization}
        The current inputs for TSGV only contain the given sentence along with the untrimmed video. However, the audio signals are not effectively exploited, which may provide extra guidance for video localization, \eg, the loud noise while using electronics in the kitchen or cheers from the audience when the football player kicks a goal. Such various forms of sounds do offer auxiliary but essential clues for more precise localization of the target moments, which has not been explored yet. Moreover, what people speak in videos can be converted into text with the Automated Speech Recognition~(ASR) technique. The converted text also provides relevant information for the cross-modal alignment between video and the text query. Nowadays, there has been many works~\cite{hori2019end,xu2019semantic}
in visual-and-language area with audio-enhanced auxiliary proving its effectiveness for performance improvements. Thus, it is a promising future direction to embed the audio information for the TSGV task.

\section{Conclusion}
\label{5-conclusion}
Temporal Sentence Grounding in Videos (TSGV) is a fundamental and challenging task connecting computer vision and natural language processing communities. It is also worth exploring since it can be seen as an intermediate task for some downstream video understanding applications such as video question answering, video summarization and video content retrieval.

In this survey, we take a systematic and insightful overview of the current research progress of the TSGV task, 
by categorizing existing approaches, benchmark datasets and evaluation metrics.
The identified limitations of current benchmarks as well as our careful thoughts on promising research directions are also provided to researchers, aiming to further promote the development for TSGV.
For future works, we suggest that i) more efforts should be made on proposing unbiased datasets and reliable metrics to better evaluate new methods for TSGV, and 
ii) models that are more robust and able to generalize well in dynamic scenarios should be paid with more attentions.

\bibliographystyle{ACM-Reference-Format}
\bibliography{references}


\begin{thebibliography}{86}


\ifx \showCODEN    \undefined \def \showCODEN     #1{\unskip}     \fi
\ifx \showDOI      \undefined \def \showDOI       #1{#1}\fi
\ifx \showISBNx    \undefined \def \showISBNx     #1{\unskip}     \fi
\ifx \showISBNxiii \undefined \def \showISBNxiii  #1{\unskip}     \fi
\ifx \showISSN     \undefined \def \showISSN      #1{\unskip}     \fi
\ifx \showLCCN     \undefined \def \showLCCN      #1{\unskip}     \fi
\ifx \shownote     \undefined \def \shownote      #1{#1}          \fi
\ifx \showarticletitle \undefined \def \showarticletitle #1{#1}   \fi
\ifx \showURL      \undefined \def \showURL       {\relax}        \fi
\providecommand\bibfield[2]{#2}
\providecommand\bibinfo[2]{#2}
\providecommand\natexlab[1]{#1}
\providecommand\showeprint[2][]{arXiv:#2}

\bibitem[\protect\citeauthoryear{Buch, Escorcia, Ghanem, Fei{-}Fei, and
  Niebles}{Buch et~al\mbox{.}}{2017}]%
        {buch2019end}
\bibfield{author}{\bibinfo{person}{Shyamal Buch}, \bibinfo{person}{Victor
  Escorcia}, \bibinfo{person}{Bernard Ghanem}, \bibinfo{person}{Li Fei{-}Fei},
  {and} \bibinfo{person}{Juan~Carlos Niebles}.}
  \bibinfo{year}{2017}\natexlab{}.
\newblock \showarticletitle{End-to-End, Single-Stream Temporal Action Detection
  in Untrimmed Videos}. In \bibinfo{booktitle}{\emph{British Machine Vision
  Conference 2017, {BMVC} 2017, London, UK, September 4-7, 2017}}.
  \bibinfo{publisher}{{BMVA} Press}.
\newblock
\urldef\tempurl%
\url{https://www.dropbox.com/s/9n90etsu6jubiax/0144.pdf?dl=1}
\showURL{%
\tempurl}


\bibitem[\protect\citeauthoryear{Cao, Zeng, Liu, He, Wang, and Qin}{Cao
  et~al\mbox{.}}{2020a}]%
        {cao2020strong}
\bibfield{author}{\bibinfo{person}{Da Cao}, \bibinfo{person}{Yawen Zeng},
  \bibinfo{person}{Meng Liu}, \bibinfo{person}{Xiangnan He},
  \bibinfo{person}{Meng Wang}, {and} \bibinfo{person}{Zheng Qin}.}
  \bibinfo{year}{2020}\natexlab{a}.
\newblock \showarticletitle{{STRONG:} Spatio-Temporal Reinforcement Learning
  for Cross-Modal Video Moment Localization}. In \bibinfo{booktitle}{\emph{{MM}
  '20: The 28th {ACM} International Conference on Multimedia, Virtual Event /
  Seattle, WA, USA, October 12-16, 2020}}. \bibinfo{pages}{4162--4170}.
\newblock
\urldef\tempurl%
\url{https://doi.org/10.1145/3394171.3413840}
\showDOI{\tempurl}


\bibitem[\protect\citeauthoryear{Cao, Zeng, Wei, Nie, Hong, and Qin}{Cao
  et~al\mbox{.}}{2020b}]%
        {cao2020adversarial}
\bibfield{author}{\bibinfo{person}{Da Cao}, \bibinfo{person}{Yawen Zeng},
  \bibinfo{person}{Xiaochi Wei}, \bibinfo{person}{Liqiang Nie},
  \bibinfo{person}{Richang Hong}, {and} \bibinfo{person}{Zheng Qin}.}
  \bibinfo{year}{2020}\natexlab{b}.
\newblock \showarticletitle{Adversarial Video Moment Retrieval by Jointly
  Modeling Ranking and Localization}. In \bibinfo{booktitle}{\emph{{MM} '20:
  The 28th {ACM} International Conference on Multimedia, Virtual Event /
  Seattle, WA, USA, October 12-16, 2020}}. \bibinfo{pages}{898--906}.
\newblock
\urldef\tempurl%
\url{https://doi.org/10.1145/3394171.3413841}
\showDOI{\tempurl}


\bibitem[\protect\citeauthoryear{Chen, Fisch, Weston, and Bordes}{Chen
  et~al\mbox{.}}{2017}]%
        {chen2017reading}
\bibfield{author}{\bibinfo{person}{Danqi Chen}, \bibinfo{person}{Adam Fisch},
  \bibinfo{person}{Jason Weston}, {and} \bibinfo{person}{Antoine Bordes}.}
  \bibinfo{year}{2017}\natexlab{}.
\newblock \showarticletitle{Reading {W}ikipedia to Answer Open-Domain
  Questions}. In \bibinfo{booktitle}{\emph{Proceedings of the 55th Annual
  Meeting of the Association for Computational Linguistics (Volume 1: Long
  Papers)}}. \bibinfo{publisher}{Association for Computational Linguistics},
  \bibinfo{address}{Vancouver, Canada}, \bibinfo{pages}{1870--1879}.
\newblock
\urldef\tempurl%
\url{https://doi.org/10.18653/v1/P17-1171}
\showDOI{\tempurl}


\bibitem[\protect\citeauthoryear{Chen, Chen, Ma, Jie, and Chua}{Chen
  et~al\mbox{.}}{2018}]%
        {chen2018temporally}
\bibfield{author}{\bibinfo{person}{Jingyuan Chen}, \bibinfo{person}{Xinpeng
  Chen}, \bibinfo{person}{Lin Ma}, \bibinfo{person}{Zequn Jie}, {and}
  \bibinfo{person}{Tat-Seng Chua}.} \bibinfo{year}{2018}\natexlab{}.
\newblock \showarticletitle{Temporally Grounding Natural Sentence in Video}. In
  \bibinfo{booktitle}{\emph{Proceedings of the 2018 Conference on Empirical
  Methods in Natural Language Processing}}. \bibinfo{publisher}{Association for
  Computational Linguistics}, \bibinfo{address}{Brussels, Belgium},
  \bibinfo{pages}{162--171}.
\newblock
\urldef\tempurl%
\url{https://doi.org/10.18653/v1/D18-1015}
\showDOI{\tempurl}


\bibitem[\protect\citeauthoryear{Chen, Ma, Chen, Jie, and Luo}{Chen
  et~al\mbox{.}}{2019a}]%
        {chen2019localizing}
\bibfield{author}{\bibinfo{person}{Jingyuan Chen}, \bibinfo{person}{Lin Ma},
  \bibinfo{person}{Xinpeng Chen}, \bibinfo{person}{Zequn Jie}, {and}
  \bibinfo{person}{Jiebo Luo}.} \bibinfo{year}{2019}\natexlab{a}.
\newblock \showarticletitle{Localizing Natural Language in Videos}. In
  \bibinfo{booktitle}{\emph{The Thirty-Third {AAAI} Conference on Artificial
  Intelligence, {AAAI} 2019, The Thirty-First Innovative Applications of
  Artificial Intelligence Conference, {IAAI} 2019, The Ninth {AAAI} Symposium
  on Educational Advances in Artificial Intelligence, {EAAI} 2019, Honolulu,
  Hawaii, USA, January 27 - February 1, 2019}}. \bibinfo{publisher}{{AAAI}
  Press}, \bibinfo{pages}{8175--8182}.
\newblock
\urldef\tempurl%
\url{https://doi.org/10.1609/aaai.v33i01.33018175}
\showDOI{\tempurl}


\bibitem[\protect\citeauthoryear{Chen, Lu, Tang, Xiao, Zhang, Tan, and Li}{Chen
  et~al\mbox{.}}{2020b}]%
        {chen2020rethinking}
\bibfield{author}{\bibinfo{person}{Long Chen}, \bibinfo{person}{Chujie Lu},
  \bibinfo{person}{Siliang Tang}, \bibinfo{person}{Jun Xiao},
  \bibinfo{person}{Dong Zhang}, \bibinfo{person}{Chilie Tan}, {and}
  \bibinfo{person}{Xiaolin Li}.} \bibinfo{year}{2020}\natexlab{b}.
\newblock \showarticletitle{Rethinking the Bottom-Up Framework for Query-Based
  Video Localization}. In \bibinfo{booktitle}{\emph{The Thirty-Fourth {AAAI}
  Conference on Artificial Intelligence, {AAAI} 2020, The Thirty-Second
  Innovative Applications of Artificial Intelligence Conference, {IAAI} 2020,
  The Tenth {AAAI} Symposium on Educational Advances in Artificial
  Intelligence, {EAAI} 2020, New York, NY, USA, February 7-12, 2020}}.
  \bibinfo{publisher}{{AAAI} Press}, \bibinfo{pages}{10551--10558}.
\newblock
\urldef\tempurl%
\url{https://aaai.org/ojs/index.php/AAAI/article/view/6627}
\showURL{%
\tempurl}


\bibitem[\protect\citeauthoryear{Chen, Jiang, Liu, and Jiang}{Chen
  et~al\mbox{.}}{2020a}]%
        {chen2020learning}
\bibfield{author}{\bibinfo{person}{Shaoxiang Chen}, \bibinfo{person}{Wenhao
  Jiang}, \bibinfo{person}{Wei Liu}, {and} \bibinfo{person}{Yu-Gang Jiang}.}
  \bibinfo{year}{2020}\natexlab{a}.
\newblock \showarticletitle{Learning modality interaction for temporal sentence
  localization and event captioning in videos}. In
  \bibinfo{booktitle}{\emph{European Conference on Computer Vision}}. Springer,
  \bibinfo{pages}{333--351}.
\newblock


\bibitem[\protect\citeauthoryear{Chen and Jiang}{Chen and Jiang}{2019}]%
        {chen2019semantic}
\bibfield{author}{\bibinfo{person}{Shaoxiang Chen} {and}
  \bibinfo{person}{Yu{-}Gang Jiang}.} \bibinfo{year}{2019}\natexlab{}.
\newblock \showarticletitle{Semantic Proposal for Activity Localization in
  Videos via Sentence Query}. In \bibinfo{booktitle}{\emph{The Thirty-Third
  {AAAI} Conference on Artificial Intelligence, {AAAI} 2019, The Thirty-First
  Innovative Applications of Artificial Intelligence Conference, {IAAI} 2019,
  The Ninth {AAAI} Symposium on Educational Advances in Artificial
  Intelligence, {EAAI} 2019, Honolulu, Hawaii, USA, January 27 - February 1,
  2019}}. \bibinfo{publisher}{{AAAI} Press}, \bibinfo{pages}{8199--8206}.
\newblock
\urldef\tempurl%
\url{https://doi.org/10.1609/aaai.v33i01.33018199}
\showDOI{\tempurl}


\bibitem[\protect\citeauthoryear{Chen and Jiang}{Chen and Jiang}{2020}]%
        {chen2020hierarchical}
\bibfield{author}{\bibinfo{person}{Shaoxiang Chen} {and}
  \bibinfo{person}{Yu-Gang Jiang}.} \bibinfo{year}{2020}\natexlab{}.
\newblock \showarticletitle{Hierarchical Visual-Textual Graph for Temporal
  Activity Localization via Language}. In \bibinfo{booktitle}{\emph{European
  Conference on Computer Vision}}. Springer, \bibinfo{pages}{601--618}.
\newblock


\bibitem[\protect\citeauthoryear{Chen and Jiang}{Chen and Jiang}{2021}]%
        {chen2021towards}
\bibfield{author}{\bibinfo{person}{Shaoxiang Chen} {and}
  \bibinfo{person}{Yu-Gang Jiang}.} \bibinfo{year}{2021}\natexlab{}.
\newblock \showarticletitle{Towards Bridging Event Captioner and Sentence
  Localizer for Weakly Supervised Dense Event Captioning}. In
  \bibinfo{booktitle}{\emph{Proceedings of the IEEE/CVF Conference on Computer
  Vision and Pattern Recognition}}. \bibinfo{pages}{8425--8435}.
\newblock


\bibitem[\protect\citeauthoryear{Chen, Ma, Luo, Tang, and Wong}{Chen
  et~al\mbox{.}}{2020c}]%
        {chen2020look}
\bibfield{author}{\bibinfo{person}{Zhenfang Chen}, \bibinfo{person}{Lin Ma},
  \bibinfo{person}{Wenhan Luo}, \bibinfo{person}{Peng Tang}, {and}
  \bibinfo{person}{Kwan-Yee~K Wong}.} \bibinfo{year}{2020}\natexlab{c}.
\newblock \showarticletitle{Look closer to ground better: Weakly-supervised
  temporal grounding of sentence in video}.
\newblock \bibinfo{journal}{\emph{arXiv preprint arXiv:2001.09308}}
  (\bibinfo{year}{2020}).
\newblock
\urldef\tempurl%
\url{https://arxiv.org/abs/2001.09308}
\showURL{%
\tempurl}


\bibitem[\protect\citeauthoryear{Chen, Ma, Luo, and Wong}{Chen
  et~al\mbox{.}}{2019b}]%
        {chen2019weakly}
\bibfield{author}{\bibinfo{person}{Zhenfang Chen}, \bibinfo{person}{Lin Ma},
  \bibinfo{person}{Wenhan Luo}, {and} \bibinfo{person}{Kwan-Yee~Kenneth Wong}.}
  \bibinfo{year}{2019}\natexlab{b}.
\newblock \showarticletitle{Weakly-Supervised Spatio-Temporally Grounding
  Natural Sentence in Video}. In \bibinfo{booktitle}{\emph{Proceedings of the
  57th Annual Meeting of the Association for Computational Linguistics}}.
  \bibinfo{publisher}{Association for Computational Linguistics},
  \bibinfo{address}{Florence, Italy}, \bibinfo{pages}{1884--1894}.
\newblock
\urldef\tempurl%
\url{https://doi.org/10.18653/v1/P19-1183}
\showDOI{\tempurl}


\bibitem[\protect\citeauthoryear{Duan, Huang, Gan, Wang, Zhu, and Huang}{Duan
  et~al\mbox{.}}{2018}]%
        {duan2018weakly}
\bibfield{author}{\bibinfo{person}{Xuguang Duan}, \bibinfo{person}{Wen{-}bing
  Huang}, \bibinfo{person}{Chuang Gan}, \bibinfo{person}{Jingdong Wang},
  \bibinfo{person}{Wenwu Zhu}, {and} \bibinfo{person}{Junzhou Huang}.}
  \bibinfo{year}{2018}\natexlab{}.
\newblock \showarticletitle{Weakly Supervised Dense Event Captioning in
  Videos}. In \bibinfo{booktitle}{\emph{Advances in Neural Information
  Processing Systems 31: Annual Conference on Neural Information Processing
  Systems 2018, NeurIPS 2018, December 3-8, 2018, Montr{\'{e}}al, Canada}},
  \bibfield{editor}{\bibinfo{person}{Samy Bengio}, \bibinfo{person}{Hanna~M.
  Wallach}, \bibinfo{person}{Hugo Larochelle}, \bibinfo{person}{Kristen
  Grauman}, \bibinfo{person}{Nicol{\`{o}} Cesa{-}Bianchi}, {and}
  \bibinfo{person}{Roman Garnett}} (Eds.). \bibinfo{pages}{3063--3073}.
\newblock
\urldef\tempurl%
\url{https://proceedings.neurips.cc/paper/2018/hash/49af6c4e558a7569d80eee2e035e2bd7-Abstract.html}
\showURL{%
\tempurl}


\bibitem[\protect\citeauthoryear{Escorcia, Soldan, Sivic, Ghanem, and
  Russell}{Escorcia et~al\mbox{.}}{2019}]%
        {escorcia2019temporal}
\bibfield{author}{\bibinfo{person}{Victor Escorcia}, \bibinfo{person}{Mattia
  Soldan}, \bibinfo{person}{Josef Sivic}, \bibinfo{person}{Bernard Ghanem},
  {and} \bibinfo{person}{Bryan Russell}.} \bibinfo{year}{2019}\natexlab{}.
\newblock \showarticletitle{Temporal localization of moments in video
  collections with natural language}.
\newblock \bibinfo{journal}{\emph{arXiv preprint arXiv:1907.12763}}
  (\bibinfo{year}{2019}).
\newblock
\urldef\tempurl%
\url{https://arxiv.org/abs/1907.12763}
\showURL{%
\tempurl}


\bibitem[\protect\citeauthoryear{Gao, Sun, Yang, and Nevatia}{Gao
  et~al\mbox{.}}{2017}]%
        {gao2017tall}
\bibfield{author}{\bibinfo{person}{Jiyang Gao}, \bibinfo{person}{Chen Sun},
  \bibinfo{person}{Zhenheng Yang}, {and} \bibinfo{person}{Ram Nevatia}.}
  \bibinfo{year}{2017}\natexlab{}.
\newblock \showarticletitle{{TALL:} Temporal Activity Localization via Language
  Query}. In \bibinfo{booktitle}{\emph{{IEEE} International Conference on
  Computer Vision, {ICCV} 2017, Venice, Italy, October 22-29, 2017}}.
  \bibinfo{publisher}{{IEEE} Computer Society}, \bibinfo{pages}{5277--5285}.
\newblock
\urldef\tempurl%
\url{https://doi.org/10.1109/ICCV.2017.563}
\showDOI{\tempurl}


\bibitem[\protect\citeauthoryear{Gao, Davis, Socher, and Xiong}{Gao
  et~al\mbox{.}}{2019}]%
        {gao2019wslln}
\bibfield{author}{\bibinfo{person}{Mingfei Gao}, \bibinfo{person}{Larry Davis},
  \bibinfo{person}{Richard Socher}, {and} \bibinfo{person}{Caiming Xiong}.}
  \bibinfo{year}{2019}\natexlab{}.
\newblock \showarticletitle{{WSLLN}:Weakly Supervised Natural Language
  Localization Networks}. In \bibinfo{booktitle}{\emph{Proceedings of the 2019
  Conference on Empirical Methods in Natural Language Processing and the 9th
  International Joint Conference on Natural Language Processing
  (EMNLP-IJCNLP)}}. \bibinfo{publisher}{Association for Computational
  Linguistics}, \bibinfo{address}{Hong Kong, China},
  \bibinfo{pages}{1481--1487}.
\newblock
\urldef\tempurl%
\url{https://doi.org/10.18653/v1/D19-1157}
\showDOI{\tempurl}


\bibitem[\protect\citeauthoryear{Ge, Gao, Chen, and Nevatia}{Ge
  et~al\mbox{.}}{2019}]%
        {ge2019mac}
\bibfield{author}{\bibinfo{person}{Runzhou Ge}, \bibinfo{person}{Jiyang Gao},
  \bibinfo{person}{Kan Chen}, {and} \bibinfo{person}{Ram Nevatia}.}
  \bibinfo{year}{2019}\natexlab{}.
\newblock \showarticletitle{Mac: Mining activity concepts for language-based
  temporal localization}. In \bibinfo{booktitle}{\emph{2019 IEEE Winter
  Conference on Applications of Computer Vision (WACV)}}. IEEE,
  \bibinfo{pages}{245--253}.
\newblock


\bibitem[\protect\citeauthoryear{Ghosh, Agarwal, Parekh, and Hauptmann}{Ghosh
  et~al\mbox{.}}{2019}]%
        {ghosh2019excl}
\bibfield{author}{\bibinfo{person}{Soham Ghosh}, \bibinfo{person}{Anuva
  Agarwal}, \bibinfo{person}{Zarana Parekh}, {and} \bibinfo{person}{Alexander
  Hauptmann}.} \bibinfo{year}{2019}\natexlab{}.
\newblock \showarticletitle{{E}x{CL}: {E}xtractive {C}lip {L}ocalization
  {U}sing {N}atural {L}anguage {D}escriptions}. In
  \bibinfo{booktitle}{\emph{Proceedings of the 2019 Conference of the North
  {A}merican Chapter of the Association for Computational Linguistics: Human
  Language Technologies, Volume 1 (Long and Short Papers)}}.
  \bibinfo{publisher}{Association for Computational Linguistics},
  \bibinfo{address}{Minneapolis, Minnesota}, \bibinfo{pages}{1984--1990}.
\newblock
\urldef\tempurl%
\url{https://doi.org/10.18653/v1/N19-1198}
\showDOI{\tempurl}


\bibitem[\protect\citeauthoryear{Girshick, Donahue, Darrell, and
  Malik}{Girshick et~al\mbox{.}}{2014}]%
        {girshick2014rich}
\bibfield{author}{\bibinfo{person}{Ross~B. Girshick}, \bibinfo{person}{Jeff
  Donahue}, \bibinfo{person}{Trevor Darrell}, {and} \bibinfo{person}{Jitendra
  Malik}.} \bibinfo{year}{2014}\natexlab{}.
\newblock \showarticletitle{Rich Feature Hierarchies for Accurate Object
  Detection and Semantic Segmentation}. In \bibinfo{booktitle}{\emph{2014
  {IEEE} Conference on Computer Vision and Pattern Recognition, {CVPR} 2014,
  Columbus, OH, USA, June 23-28, 2014}}. \bibinfo{publisher}{{IEEE} Computer
  Society}, \bibinfo{pages}{580--587}.
\newblock
\urldef\tempurl%
\url{https://doi.org/10.1109/CVPR.2014.81}
\showDOI{\tempurl}


\bibitem[\protect\citeauthoryear{Hahn, Kadav, Rehg, and Graf}{Hahn
  et~al\mbox{.}}{2020}]%
        {hahn2019tripping}
\bibfield{author}{\bibinfo{person}{Meera Hahn}, \bibinfo{person}{Asim Kadav},
  \bibinfo{person}{James~M. Rehg}, {and} \bibinfo{person}{Hans~Peter Graf}.}
  \bibinfo{year}{2020}\natexlab{}.
\newblock \showarticletitle{Tripping through time: Efficient Localization of
  Activities in Videos}. In \bibinfo{booktitle}{\emph{31st British Machine
  Vision Conference 2020, {BMVC} 2020, Virtual Event, UK, September 7-10,
  2020}}. \bibinfo{publisher}{{BMVA} Press}.
\newblock
\urldef\tempurl%
\url{https://www.bmvc2020-conference.com/assets/papers/0549.pdf}
\showURL{%
\tempurl}


\bibitem[\protect\citeauthoryear{He, Zhao, Huang, Li, Liu, and Wen}{He
  et~al\mbox{.}}{2019}]%
        {he2019read}
\bibfield{author}{\bibinfo{person}{Dongliang He}, \bibinfo{person}{Xiang Zhao},
  \bibinfo{person}{Jizhou Huang}, \bibinfo{person}{Fu Li},
  \bibinfo{person}{Xiao Liu}, {and} \bibinfo{person}{Shilei Wen}.}
  \bibinfo{year}{2019}\natexlab{}.
\newblock \showarticletitle{Read, Watch, and Move: Reinforcement Learning for
  Temporally Grounding Natural Language Descriptions in Videos}. In
  \bibinfo{booktitle}{\emph{The Thirty-Third {AAAI} Conference on Artificial
  Intelligence, {AAAI} 2019, The Thirty-First Innovative Applications of
  Artificial Intelligence Conference, {IAAI} 2019, The Ninth {AAAI} Symposium
  on Educational Advances in Artificial Intelligence, {EAAI} 2019, Honolulu,
  Hawaii, USA, January 27 - February 1, 2019}}. \bibinfo{publisher}{{AAAI}
  Press}, \bibinfo{pages}{8393--8400}.
\newblock
\urldef\tempurl%
\url{https://doi.org/10.1609/aaai.v33i01.33018393}
\showDOI{\tempurl}


\bibitem[\protect\citeauthoryear{Hendricks, Wang, Shechtman, Sivic, Darrell,
  and Russell}{Hendricks et~al\mbox{.}}{2017}]%
        {anne2017localizing}
\bibfield{author}{\bibinfo{person}{Lisa~Anne Hendricks},
  \bibinfo{person}{Oliver Wang}, \bibinfo{person}{Eli Shechtman},
  \bibinfo{person}{Josef Sivic}, \bibinfo{person}{Trevor Darrell}, {and}
  \bibinfo{person}{Bryan~C. Russell}.} \bibinfo{year}{2017}\natexlab{}.
\newblock \showarticletitle{Localizing Moments in Video with Natural Language}.
  In \bibinfo{booktitle}{\emph{{IEEE} International Conference on Computer
  Vision, {ICCV} 2017, Venice, Italy, October 22-29, 2017}}.
  \bibinfo{publisher}{{IEEE} Computer Society}, \bibinfo{pages}{5804--5813}.
\newblock
\urldef\tempurl%
\url{https://doi.org/10.1109/ICCV.2017.618}
\showDOI{\tempurl}


\bibitem[\protect\citeauthoryear{Hori, AlAmri, Wang, Wichern, Hori, Cherian,
  Marks, Cartillier, Lopes, Das, Essa, Batra, and Parikh}{Hori
  et~al\mbox{.}}{2019}]%
        {hori2019end}
\bibfield{author}{\bibinfo{person}{Chiori Hori}, \bibinfo{person}{Huda AlAmri},
  \bibinfo{person}{Jue Wang}, \bibinfo{person}{Gordon Wichern},
  \bibinfo{person}{Takaaki Hori}, \bibinfo{person}{Anoop Cherian},
  \bibinfo{person}{Tim~K. Marks}, \bibinfo{person}{Vincent Cartillier},
  \bibinfo{person}{Raphael~Gontijo Lopes}, \bibinfo{person}{Abhishek Das},
  \bibinfo{person}{Irfan Essa}, \bibinfo{person}{Dhruv Batra}, {and}
  \bibinfo{person}{Devi Parikh}.} \bibinfo{year}{2019}\natexlab{}.
\newblock \showarticletitle{End-to-end Audio Visual Scene-aware Dialog Using
  Multimodal Attention-based Video Features}. In
  \bibinfo{booktitle}{\emph{{IEEE} International Conference on Acoustics,
  Speech and Signal Processing, {ICASSP} 2019, Brighton, United Kingdom, May
  12-17, 2019}}. \bibinfo{publisher}{{IEEE}}, \bibinfo{pages}{2352--2356}.
\newblock
\urldef\tempurl%
\url{https://doi.org/10.1109/ICASSP.2019.8682583}
\showDOI{\tempurl}


\bibitem[\protect\citeauthoryear{Hu, Xu, Rohrbach, Feng, Saenko, and
  Darrell}{Hu et~al\mbox{.}}{2016}]%
        {hu2016natural}
\bibfield{author}{\bibinfo{person}{Ronghang Hu}, \bibinfo{person}{Huazhe Xu},
  \bibinfo{person}{Marcus Rohrbach}, \bibinfo{person}{Jiashi Feng},
  \bibinfo{person}{Kate Saenko}, {and} \bibinfo{person}{Trevor Darrell}.}
  \bibinfo{year}{2016}\natexlab{}.
\newblock \showarticletitle{Natural Language Object Retrieval}. In
  \bibinfo{booktitle}{\emph{2016 {IEEE} Conference on Computer Vision and
  Pattern Recognition, {CVPR} 2016, Las Vegas, NV, USA, June 27-30, 2016}}.
  \bibinfo{publisher}{{IEEE} Computer Society}, \bibinfo{pages}{4555--4564}.
\newblock
\urldef\tempurl%
\url{https://doi.org/10.1109/CVPR.2016.493}
\showDOI{\tempurl}


\bibitem[\protect\citeauthoryear{Huang, Liu, Gong, and Jin}{Huang
  et~al\mbox{.}}{2021}]%
        {huang2021cross}
\bibfield{author}{\bibinfo{person}{Jiabo Huang}, \bibinfo{person}{Yang Liu},
  \bibinfo{person}{Shaogang Gong}, {and} \bibinfo{person}{Hailin Jin}.}
  \bibinfo{year}{2021}\natexlab{}.
\newblock \showarticletitle{Cross-Sentence Temporal and Semantic Relations in
  Video Activity Localisation}.
\newblock \bibinfo{journal}{\emph{arXiv preprint arXiv:2107.11443}}
  (\bibinfo{year}{2021}).
\newblock
\urldef\tempurl%
\url{https://arxiv.org/abs/2107.11443}
\showURL{%
\tempurl}


\bibitem[\protect\citeauthoryear{Jiang, Huang, Yang, and Yuan}{Jiang
  et~al\mbox{.}}{2019}]%
        {jiang2019cross}
\bibfield{author}{\bibinfo{person}{Bin Jiang}, \bibinfo{person}{Xin Huang},
  \bibinfo{person}{Chao Yang}, {and} \bibinfo{person}{Junsong Yuan}.}
  \bibinfo{year}{2019}\natexlab{}.
\newblock \showarticletitle{Cross-modal video moment retrieval with spatial and
  language-temporal attention}. In \bibinfo{booktitle}{\emph{Proceedings of the
  2019 on international conference on multimedia retrieval}}.
  \bibinfo{pages}{217--225}.
\newblock


\bibitem[\protect\citeauthoryear{Jiao, Li, Huang, Yang, Liu, and Zhang}{Jiao
  et~al\mbox{.}}{2018}]%
        {jiao2018three}
\bibfield{author}{\bibinfo{person}{Yifan Jiao}, \bibinfo{person}{Zhetao Li},
  \bibinfo{person}{Shucheng Huang}, \bibinfo{person}{Xiaoshan Yang},
  \bibinfo{person}{Bin Liu}, {and} \bibinfo{person}{Tianzhu Zhang}.}
  \bibinfo{year}{2018}\natexlab{}.
\newblock \showarticletitle{Three-dimensional attention-based deep ranking
  model for video highlight detection}.
\newblock \bibinfo{journal}{\emph{IEEE Transactions on Multimedia}}
  \bibinfo{volume}{20}, \bibinfo{number}{10} (\bibinfo{year}{2018}),
  \bibinfo{pages}{2693--2705}.
\newblock


\bibitem[\protect\citeauthoryear{Karaman, Seidenari, and Del~Bimbo}{Karaman
  et~al\mbox{.}}{2014}]%
        {karaman2014fast}
\bibfield{author}{\bibinfo{person}{Svebor Karaman}, \bibinfo{person}{Lorenzo
  Seidenari}, {and} \bibinfo{person}{Alberto Del~Bimbo}.}
  \bibinfo{year}{2014}\natexlab{}.
\newblock \showarticletitle{Fast saliency based pooling of fisher encoded dense
  trajectories}. In \bibinfo{booktitle}{\emph{ECCV THUMOS Workshop}},
  Vol.~\bibinfo{volume}{1}. \bibinfo{pages}{5}.
\newblock


\bibitem[\protect\citeauthoryear{Kipf and Welling}{Kipf and Welling}{2017}]%
        {kipf2016semi}
\bibfield{author}{\bibinfo{person}{Thomas~N. Kipf} {and} \bibinfo{person}{Max
  Welling}.} \bibinfo{year}{2017}\natexlab{}.
\newblock \showarticletitle{Semi-Supervised Classification with Graph
  Convolutional Networks}. In \bibinfo{booktitle}{\emph{5th International
  Conference on Learning Representations, {ICLR} 2017, Toulon, France, April
  24-26, 2017, Conference Track Proceedings}}.
  \bibinfo{publisher}{OpenReview.net}.
\newblock
\urldef\tempurl%
\url{https://openreview.net/forum?id=SJU4ayYgl}
\showURL{%
\tempurl}


\bibitem[\protect\citeauthoryear{Krishna, Hata, Ren, Fei{-}Fei, and
  Niebles}{Krishna et~al\mbox{.}}{2017}]%
        {krishna2017dense}
\bibfield{author}{\bibinfo{person}{Ranjay Krishna}, \bibinfo{person}{Kenji
  Hata}, \bibinfo{person}{Frederic Ren}, \bibinfo{person}{Li Fei{-}Fei}, {and}
  \bibinfo{person}{Juan~Carlos Niebles}.} \bibinfo{year}{2017}\natexlab{}.
\newblock \showarticletitle{Dense-Captioning Events in Videos}. In
  \bibinfo{booktitle}{\emph{{IEEE} International Conference on Computer Vision,
  {ICCV} 2017, Venice, Italy, October 22-29, 2017}}. \bibinfo{publisher}{{IEEE}
  Computer Society}, \bibinfo{pages}{706--715}.
\newblock
\urldef\tempurl%
\url{https://doi.org/10.1109/ICCV.2017.83}
\showDOI{\tempurl}


\bibitem[\protect\citeauthoryear{Lei, Yu, Berg, and Bansal}{Lei
  et~al\mbox{.}}{2020}]%
        {lei2020tvr}
\bibfield{author}{\bibinfo{person}{Jie Lei}, \bibinfo{person}{Licheng Yu},
  \bibinfo{person}{Tamara~L Berg}, {and} \bibinfo{person}{Mohit Bansal}.}
  \bibinfo{year}{2020}\natexlab{}.
\newblock \showarticletitle{Tvr: A large-scale dataset for video-subtitle
  moment retrieval}. In \bibinfo{booktitle}{\emph{Computer Vision--ECCV 2020:
  16th European Conference, Glasgow, UK, August 23--28, 2020, Proceedings, Part
  XXI 16}}. Springer, \bibinfo{pages}{447--463}.
\newblock


\bibitem[\protect\citeauthoryear{Lin, Zhao, and Shou}{Lin
  et~al\mbox{.}}{2017}]%
        {lin2017single}
\bibfield{author}{\bibinfo{person}{Tianwei Lin}, \bibinfo{person}{Xu Zhao},
  {and} \bibinfo{person}{Zheng Shou}.} \bibinfo{year}{2017}\natexlab{}.
\newblock \showarticletitle{Single Shot Temporal Action Detection}. In
  \bibinfo{booktitle}{\emph{Proceedings of the 2017 {ACM} on Multimedia
  Conference, {MM} 2017, Mountain View, CA, USA, October 23-27, 2017}}.
  \bibinfo{pages}{988--996}.
\newblock
\urldef\tempurl%
\url{https://doi.org/10.1145/3123266.3123343}
\showDOI{\tempurl}


\bibitem[\protect\citeauthoryear{Lin, Zhao, Zhang, Wang, and Liu}{Lin
  et~al\mbox{.}}{2020}]%
        {lin2020weakly}
\bibfield{author}{\bibinfo{person}{Zhijie Lin}, \bibinfo{person}{Zhou Zhao},
  \bibinfo{person}{Zhu Zhang}, \bibinfo{person}{Qi Wang}, {and}
  \bibinfo{person}{Huasheng Liu}.} \bibinfo{year}{2020}\natexlab{}.
\newblock \showarticletitle{Weakly-Supervised Video Moment Retrieval via
  Semantic Completion Network}. In \bibinfo{booktitle}{\emph{The Thirty-Fourth
  {AAAI} Conference on Artificial Intelligence, {AAAI} 2020, The Thirty-Second
  Innovative Applications of Artificial Intelligence Conference, {IAAI} 2020,
  The Tenth {AAAI} Symposium on Educational Advances in Artificial
  Intelligence, {EAAI} 2020, New York, NY, USA, February 7-12, 2020}}.
  \bibinfo{publisher}{{AAAI} Press}, \bibinfo{pages}{11539--11546}.
\newblock
\urldef\tempurl%
\url{https://aaai.org/ojs/index.php/AAAI/article/view/6820}
\showURL{%
\tempurl}


\bibitem[\protect\citeauthoryear{Liu, Qu, Dong, Zhou, Cheng, Wei, Xu, and
  Xie}{Liu et~al\mbox{.}}{2021}]%
        {liu2021context}
\bibfield{author}{\bibinfo{person}{Daizong Liu}, \bibinfo{person}{Xiaoye Qu},
  \bibinfo{person}{Jianfeng Dong}, \bibinfo{person}{Pan Zhou},
  \bibinfo{person}{Yu Cheng}, \bibinfo{person}{Wei Wei},
  \bibinfo{person}{Zichuan Xu}, {and} \bibinfo{person}{Yulai Xie}.}
  \bibinfo{year}{2021}\natexlab{}.
\newblock \showarticletitle{Context-aware Biaffine Localizing Network for
  Temporal Sentence Grounding}. In \bibinfo{booktitle}{\emph{Proceedings of the
  IEEE/CVF Conference on Computer Vision and Pattern Recognition}}.
  \bibinfo{pages}{11235--11244}.
\newblock


\bibitem[\protect\citeauthoryear{Liu, Qu, Liu, Dong, Zhou, and Xu}{Liu
  et~al\mbox{.}}{2020}]%
        {liu2020jointly}
\bibfield{author}{\bibinfo{person}{Daizong Liu}, \bibinfo{person}{Xiaoye Qu},
  \bibinfo{person}{Xiao{-}Yang Liu}, \bibinfo{person}{Jianfeng Dong},
  \bibinfo{person}{Pan Zhou}, {and} \bibinfo{person}{Zichuan Xu}.}
  \bibinfo{year}{2020}\natexlab{}.
\newblock \showarticletitle{Jointly Cross- and Self-Modal Graph Attention
  Network for Query-Based Moment Localization}. In
  \bibinfo{booktitle}{\emph{{MM} '20: The 28th {ACM} International Conference
  on Multimedia, Virtual Event / Seattle, WA, USA, October 12-16, 2020}}.
  \bibinfo{pages}{4070--4078}.
\newblock
\urldef\tempurl%
\url{https://doi.org/10.1145/3394171.3414026}
\showDOI{\tempurl}


\bibitem[\protect\citeauthoryear{Liu, Wang, Nie, He, Chen, and Chua}{Liu
  et~al\mbox{.}}{2018a}]%
        {liu2018attentive}
\bibfield{author}{\bibinfo{person}{Meng Liu}, \bibinfo{person}{Xiang Wang},
  \bibinfo{person}{Liqiang Nie}, \bibinfo{person}{Xiangnan He},
  \bibinfo{person}{Baoquan Chen}, {and} \bibinfo{person}{Tat{-}Seng Chua}.}
  \bibinfo{year}{2018}\natexlab{a}.
\newblock \showarticletitle{Attentive Moment Retrieval in Videos}. In
  \bibinfo{booktitle}{\emph{The 41st International {ACM} {SIGIR} Conference on
  Research {\&} Development in Information Retrieval, {SIGIR} 2018, Ann Arbor,
  MI, USA, July 08-12, 2018}}, \bibfield{editor}{\bibinfo{person}{Kevyn
  Collins{-}Thompson}, \bibinfo{person}{Qiaozhu Mei}, \bibinfo{person}{Brian~D.
  Davison}, \bibinfo{person}{Yiqun Liu}, {and} \bibinfo{person}{Emine Yilmaz}}
  (Eds.). \bibinfo{publisher}{{ACM}}, \bibinfo{pages}{15--24}.
\newblock
\urldef\tempurl%
\url{https://doi.org/10.1145/3209978.3210003}
\showDOI{\tempurl}


\bibitem[\protect\citeauthoryear{Liu, Wang, Nie, Tian, Chen, and Chua}{Liu
  et~al\mbox{.}}{2018b}]%
        {liu2018cross}
\bibfield{author}{\bibinfo{person}{Meng Liu}, \bibinfo{person}{Xiang Wang},
  \bibinfo{person}{Liqiang Nie}, \bibinfo{person}{Qi Tian},
  \bibinfo{person}{Baoquan Chen}, {and} \bibinfo{person}{Tat{-}Seng Chua}.}
  \bibinfo{year}{2018}\natexlab{b}.
\newblock \showarticletitle{Cross-modal Moment Localization in Videos}. In
  \bibinfo{booktitle}{\emph{2018 {ACM} Multimedia Conference on Multimedia
  Conference, {MM} 2018, Seoul, Republic of Korea, October 22-26, 2018}}.
  \bibinfo{pages}{843--851}.
\newblock
\urldef\tempurl%
\url{https://doi.org/10.1145/3240508.3240549}
\showDOI{\tempurl}


\bibitem[\protect\citeauthoryear{Lu, Chen, Tan, Li, and Xiao}{Lu
  et~al\mbox{.}}{2019}]%
        {lu2019debug}
\bibfield{author}{\bibinfo{person}{Chujie Lu}, \bibinfo{person}{Long Chen},
  \bibinfo{person}{Chilie Tan}, \bibinfo{person}{Xiaolin Li}, {and}
  \bibinfo{person}{Jun Xiao}.} \bibinfo{year}{2019}\natexlab{}.
\newblock \showarticletitle{{DEBUG}: A Dense Bottom-Up Grounding Approach for
  Natural Language Video Localization}. In
  \bibinfo{booktitle}{\emph{Proceedings of the 2019 Conference on Empirical
  Methods in Natural Language Processing and the 9th International Joint
  Conference on Natural Language Processing (EMNLP-IJCNLP)}}.
  \bibinfo{publisher}{Association for Computational Linguistics},
  \bibinfo{address}{Hong Kong, China}, \bibinfo{pages}{5144--5153}.
\newblock
\urldef\tempurl%
\url{https://doi.org/10.18653/v1/D19-1518}
\showDOI{\tempurl}


\bibitem[\protect\citeauthoryear{Ma, Yoon, Kim, Lee, Kang, and Yoo}{Ma
  et~al\mbox{.}}{2020}]%
        {ma2020vlanet}
\bibfield{author}{\bibinfo{person}{Minuk Ma}, \bibinfo{person}{Sunjae Yoon},
  \bibinfo{person}{Junyeong Kim}, \bibinfo{person}{Youngjoon Lee},
  \bibinfo{person}{Sunghun Kang}, {and} \bibinfo{person}{Chang~D Yoo}.}
  \bibinfo{year}{2020}\natexlab{}.
\newblock \showarticletitle{Vlanet: Video-language alignment network for
  weakly-supervised video moment retrieval}. In
  \bibinfo{booktitle}{\emph{European Conference on Computer Vision}}. Springer,
  \bibinfo{pages}{156--171}.
\newblock


\bibitem[\protect\citeauthoryear{Ma, Sigal, and Sclaroff}{Ma
  et~al\mbox{.}}{2016}]%
        {ma2016learning}
\bibfield{author}{\bibinfo{person}{Shugao Ma}, \bibinfo{person}{Leonid Sigal},
  {and} \bibinfo{person}{Stan Sclaroff}.} \bibinfo{year}{2016}\natexlab{}.
\newblock \showarticletitle{Learning Activity Progression in LSTMs for Activity
  Detection and Early Detection}. In \bibinfo{booktitle}{\emph{2016 {IEEE}
  Conference on Computer Vision and Pattern Recognition, {CVPR} 2016, Las
  Vegas, NV, USA, June 27-30, 2016}}. \bibinfo{publisher}{{IEEE} Computer
  Society}, \bibinfo{pages}{1942--1950}.
\newblock
\urldef\tempurl%
\url{https://doi.org/10.1109/CVPR.2016.214}
\showDOI{\tempurl}


\bibitem[\protect\citeauthoryear{Ma, Lu, Zhang, and Li}{Ma
  et~al\mbox{.}}{2002}]%
        {ma2002user}
\bibfield{author}{\bibinfo{person}{Yu-Fei Ma}, \bibinfo{person}{Lie Lu},
  \bibinfo{person}{Hong-Jiang Zhang}, {and} \bibinfo{person}{Mingjing Li}.}
  \bibinfo{year}{2002}\natexlab{}.
\newblock \showarticletitle{A user attention model for video summarization}. In
  \bibinfo{booktitle}{\emph{Proceedings of the tenth ACM international
  conference on Multimedia}}. \bibinfo{pages}{533--542}.
\newblock


\bibitem[\protect\citeauthoryear{Mithun, Paul, and Roy{-}Chowdhury}{Mithun
  et~al\mbox{.}}{2019}]%
        {mithun2019weakly}
\bibfield{author}{\bibinfo{person}{Niluthpol~Chowdhury Mithun},
  \bibinfo{person}{Sujoy Paul}, {and} \bibinfo{person}{Amit~K.
  Roy{-}Chowdhury}.} \bibinfo{year}{2019}\natexlab{}.
\newblock \showarticletitle{Weakly Supervised Video Moment Retrieval From Text
  Queries}. In \bibinfo{booktitle}{\emph{{IEEE} Conference on Computer Vision
  and Pattern Recognition, {CVPR} 2019, Long Beach, CA, USA, June 16-20,
  2019}}. \bibinfo{publisher}{Computer Vision Foundation / {IEEE}},
  \bibinfo{pages}{11592--11601}.
\newblock
\urldef\tempurl%
\url{https://doi.org/10.1109/CVPR.2019.01186}
\showDOI{\tempurl}


\bibitem[\protect\citeauthoryear{Mun, Cho, and Han}{Mun et~al\mbox{.}}{2020}]%
        {mun2020local}
\bibfield{author}{\bibinfo{person}{Jonghwan Mun}, \bibinfo{person}{Minsu Cho},
  {and} \bibinfo{person}{Bohyung Han}.} \bibinfo{year}{2020}\natexlab{}.
\newblock \showarticletitle{Local-Global Video-Text Interactions for Temporal
  Grounding}. In \bibinfo{booktitle}{\emph{2020 {IEEE/CVF} Conference on
  Computer Vision and Pattern Recognition, {CVPR} 2020, Seattle, WA, USA, June
  13-19, 2020}}. \bibinfo{publisher}{{IEEE}}, \bibinfo{pages}{10807--10816}.
\newblock
\urldef\tempurl%
\url{https://doi.org/10.1109/CVPR42600.2020.01082}
\showDOI{\tempurl}


\bibitem[\protect\citeauthoryear{Otani, Nakashima, Rahtu, and
  Heikkil{\"{a}}}{Otani et~al\mbox{.}}{2020}]%
        {otani2020uncovering}
\bibfield{author}{\bibinfo{person}{Mayu Otani}, \bibinfo{person}{Yuta
  Nakashima}, \bibinfo{person}{Esa Rahtu}, {and} \bibinfo{person}{Janne
  Heikkil{\"{a}}}.} \bibinfo{year}{2020}\natexlab{}.
\newblock \showarticletitle{Uncovering Hidden Challenges in Query-Based Video
  Moment Retrieval}. In \bibinfo{booktitle}{\emph{31st British Machine Vision
  Conference 2020, {BMVC} 2020, Virtual Event, UK, September 7-10, 2020}}.
  \bibinfo{publisher}{{BMVA} Press}.
\newblock
\urldef\tempurl%
\url{https://www.bmvc2020-conference.com/assets/papers/0306.pdf}
\showURL{%
\tempurl}


\bibitem[\protect\citeauthoryear{Qu, Tang, Zou, Cheng, Dong, Zhou, and Xu}{Qu
  et~al\mbox{.}}{2020}]%
        {qu2020fine}
\bibfield{author}{\bibinfo{person}{Xiaoye Qu}, \bibinfo{person}{Pengwei Tang},
  \bibinfo{person}{Zhikang Zou}, \bibinfo{person}{Yu Cheng},
  \bibinfo{person}{Jianfeng Dong}, \bibinfo{person}{Pan Zhou}, {and}
  \bibinfo{person}{Zichuan Xu}.} \bibinfo{year}{2020}\natexlab{}.
\newblock \showarticletitle{Fine-grained Iterative Attention Network for
  Temporal Language Localization in Videos}. In \bibinfo{booktitle}{\emph{{MM}
  '20: The 28th {ACM} International Conference on Multimedia, Virtual Event /
  Seattle, WA, USA, October 12-16, 2020}}. \bibinfo{pages}{4280--4288}.
\newblock
\urldef\tempurl%
\url{https://doi.org/10.1145/3394171.3414053}
\showDOI{\tempurl}


\bibitem[\protect\citeauthoryear{Regneri, Rohrbach, Wetzel, Thater, Schiele,
  and Pinkal}{Regneri et~al\mbox{.}}{2013}]%
        {regneri2013grounding}
\bibfield{author}{\bibinfo{person}{Michaela Regneri}, \bibinfo{person}{Marcus
  Rohrbach}, \bibinfo{person}{Dominikus Wetzel}, \bibinfo{person}{Stefan
  Thater}, \bibinfo{person}{Bernt Schiele}, {and} \bibinfo{person}{Manfred
  Pinkal}.} \bibinfo{year}{2013}\natexlab{}.
\newblock \showarticletitle{Grounding Action Descriptions in Videos}.
\newblock \bibinfo{journal}{\emph{Transactions of the Association for
  Computational Linguistics}}  \bibinfo{volume}{1} (\bibinfo{year}{2013}),
  \bibinfo{pages}{25--36}.
\newblock
\urldef\tempurl%
\url{https://doi.org/10.1162/tacl_a_00207}
\showDOI{\tempurl}


\bibitem[\protect\citeauthoryear{Rodriguez, Marrese-Taylor, Saleh, Li, and
  Gould}{Rodriguez et~al\mbox{.}}{2020}]%
        {rodriguez2020proposal}
\bibfield{author}{\bibinfo{person}{Cristian Rodriguez}, \bibinfo{person}{Edison
  Marrese-Taylor}, \bibinfo{person}{Fatemeh~Sadat Saleh},
  \bibinfo{person}{Hongdong Li}, {and} \bibinfo{person}{Stephen Gould}.}
  \bibinfo{year}{2020}\natexlab{}.
\newblock \showarticletitle{Proposal-free temporal moment localization of a
  natural-language query in video using guided attention}. In
  \bibinfo{booktitle}{\emph{Proceedings of the IEEE/CVF Winter Conference on
  Applications of Computer Vision}}. \bibinfo{pages}{2464--2473}.
\newblock


\bibitem[\protect\citeauthoryear{Rohrbach, Regneri, Andriluka, Amin, Pinkal,
  and Schiele}{Rohrbach et~al\mbox{.}}{2012}]%
        {rohrbach2012script}
\bibfield{author}{\bibinfo{person}{Marcus Rohrbach}, \bibinfo{person}{Michaela
  Regneri}, \bibinfo{person}{Mykhaylo Andriluka}, \bibinfo{person}{Sikandar
  Amin}, \bibinfo{person}{Manfred Pinkal}, {and} \bibinfo{person}{Bernt
  Schiele}.} \bibinfo{year}{2012}\natexlab{}.
\newblock \showarticletitle{Script data for attribute-based recognition of
  composite activities}. In \bibinfo{booktitle}{\emph{European conference on
  computer vision}}. Springer, \bibinfo{pages}{144--157}.
\newblock


\bibitem[\protect\citeauthoryear{Sadhu, Chen, and Nevatia}{Sadhu
  et~al\mbox{.}}{2020}]%
        {sadhu2020video}
\bibfield{author}{\bibinfo{person}{Arka Sadhu}, \bibinfo{person}{Kan Chen},
  {and} \bibinfo{person}{Ram Nevatia}.} \bibinfo{year}{2020}\natexlab{}.
\newblock \showarticletitle{Video Object Grounding Using Semantic Roles in
  Language Description}. In \bibinfo{booktitle}{\emph{2020 {IEEE/CVF}
  Conference on Computer Vision and Pattern Recognition, {CVPR} 2020, Seattle,
  WA, USA, June 13-19, 2020}}. \bibinfo{publisher}{{IEEE}},
  \bibinfo{pages}{10414--10424}.
\newblock
\urldef\tempurl%
\url{https://doi.org/10.1109/CVPR42600.2020.01043}
\showDOI{\tempurl}


\bibitem[\protect\citeauthoryear{Shou, Wang, and Chang}{Shou
  et~al\mbox{.}}{2016}]%
        {shou2016temporal}
\bibfield{author}{\bibinfo{person}{Zheng Shou}, \bibinfo{person}{Dongang Wang},
  {and} \bibinfo{person}{Shih{-}Fu Chang}.} \bibinfo{year}{2016}\natexlab{}.
\newblock \showarticletitle{Temporal Action Localization in Untrimmed Videos
  via Multi-stage CNNs}. In \bibinfo{booktitle}{\emph{2016 {IEEE} Conference on
  Computer Vision and Pattern Recognition, {CVPR} 2016, Las Vegas, NV, USA,
  June 27-30, 2016}}. \bibinfo{publisher}{{IEEE} Computer Society},
  \bibinfo{pages}{1049--1058}.
\newblock
\urldef\tempurl%
\url{https://doi.org/10.1109/CVPR.2016.119}
\showDOI{\tempurl}


\bibitem[\protect\citeauthoryear{Sigurdsson, Varol, Wang, Farhadi, Laptev, and
  Gupta}{Sigurdsson et~al\mbox{.}}{2016}]%
        {sigurdsson2016hollywood}
\bibfield{author}{\bibinfo{person}{Gunnar~A Sigurdsson},
  \bibinfo{person}{G{\"u}l Varol}, \bibinfo{person}{Xiaolong Wang},
  \bibinfo{person}{Ali Farhadi}, \bibinfo{person}{Ivan Laptev}, {and}
  \bibinfo{person}{Abhinav Gupta}.} \bibinfo{year}{2016}\natexlab{}.
\newblock \showarticletitle{Hollywood in homes: Crowdsourcing data collection
  for activity understanding}. In \bibinfo{booktitle}{\emph{European Conference
  on Computer Vision}}. Springer, \bibinfo{pages}{510--526}.
\newblock


\bibitem[\protect\citeauthoryear{Singh, Marks, Jones, Tuzel, and Shao}{Singh
  et~al\mbox{.}}{2016}]%
        {singh2016multi}
\bibfield{author}{\bibinfo{person}{Bharat Singh}, \bibinfo{person}{Tim~K.
  Marks}, \bibinfo{person}{Michael~J. Jones}, \bibinfo{person}{Oncel Tuzel},
  {and} \bibinfo{person}{Ming Shao}.} \bibinfo{year}{2016}\natexlab{}.
\newblock \showarticletitle{A Multi-stream Bi-directional Recurrent Neural
  Network for Fine-Grained Action Detection}. In \bibinfo{booktitle}{\emph{2016
  {IEEE} Conference on Computer Vision and Pattern Recognition, {CVPR} 2016,
  Las Vegas, NV, USA, June 27-30, 2016}}. \bibinfo{publisher}{{IEEE} Computer
  Society}, \bibinfo{pages}{1961--1970}.
\newblock
\urldef\tempurl%
\url{https://doi.org/10.1109/CVPR.2016.216}
\showDOI{\tempurl}


\bibitem[\protect\citeauthoryear{Song, Wang, Ma, Yu, and Yu}{Song
  et~al\mbox{.}}{2020}]%
        {song2020weakly}
\bibfield{author}{\bibinfo{person}{Yijun Song}, \bibinfo{person}{Jingwen Wang},
  \bibinfo{person}{Lin Ma}, \bibinfo{person}{Zhou Yu}, {and}
  \bibinfo{person}{Jun Yu}.} \bibinfo{year}{2020}\natexlab{}.
\newblock \showarticletitle{Weakly-supervised multi-level attentional
  reconstruction network for grounding textual queries in videos}.
\newblock \bibinfo{journal}{\emph{arXiv preprint arXiv:2003.07048}}
  (\bibinfo{year}{2020}).
\newblock
\urldef\tempurl%
\url{https://arxiv.org/abs/2003.07048}
\showURL{%
\tempurl}


\bibitem[\protect\citeauthoryear{Tan, Xu, Saenko, and Plummer}{Tan
  et~al\mbox{.}}{2021}]%
        {tan2021logan}
\bibfield{author}{\bibinfo{person}{Reuben Tan}, \bibinfo{person}{Huijuan Xu},
  \bibinfo{person}{Kate Saenko}, {and} \bibinfo{person}{Bryan~A Plummer}.}
  \bibinfo{year}{2021}\natexlab{}.
\newblock \showarticletitle{Logan: Latent graph co-attention network for
  weakly-supervised video moment retrieval}. In
  \bibinfo{booktitle}{\emph{Proceedings of the IEEE/CVF Winter Conference on
  Applications of Computer Vision}}. \bibinfo{pages}{2083--2092}.
\newblock


\bibitem[\protect\citeauthoryear{Tang, Liao, Liu, Li, Jin, Jiang, Yu, and
  Xu}{Tang et~al\mbox{.}}{2021}]%
        {tang2021human}
\bibfield{author}{\bibinfo{person}{Zongheng Tang}, \bibinfo{person}{Yue Liao},
  \bibinfo{person}{Si Liu}, \bibinfo{person}{Guanbin Li},
  \bibinfo{person}{Xiaojie Jin}, \bibinfo{person}{Hongxu Jiang},
  \bibinfo{person}{Qian Yu}, {and} \bibinfo{person}{Dong Xu}.}
  \bibinfo{year}{2021}\natexlab{}.
\newblock \showarticletitle{Human-centric spatio-temporal video grounding with
  visual transformers}.
\newblock \bibinfo{journal}{\emph{IEEE Transactions on Circuits and Systems for
  Video Technology}} (\bibinfo{year}{2021}).
\newblock


\bibitem[\protect\citeauthoryear{Vaswani, Shazeer, Parmar, Uszkoreit, Jones,
  Gomez, Kaiser, and Polosukhin}{Vaswani et~al\mbox{.}}{2017}]%
        {vaswani2017attention}
\bibfield{author}{\bibinfo{person}{Ashish Vaswani}, \bibinfo{person}{Noam
  Shazeer}, \bibinfo{person}{Niki Parmar}, \bibinfo{person}{Jakob Uszkoreit},
  \bibinfo{person}{Llion Jones}, \bibinfo{person}{Aidan~N. Gomez},
  \bibinfo{person}{Lukasz Kaiser}, {and} \bibinfo{person}{Illia Polosukhin}.}
  \bibinfo{year}{2017}\natexlab{}.
\newblock \showarticletitle{Attention is All you Need}. In
  \bibinfo{booktitle}{\emph{Advances in Neural Information Processing Systems
  30: Annual Conference on Neural Information Processing Systems 2017, December
  4-9, 2017, Long Beach, CA, {USA}}},
  \bibfield{editor}{\bibinfo{person}{Isabelle Guyon}, \bibinfo{person}{Ulrike
  von Luxburg}, \bibinfo{person}{Samy Bengio}, \bibinfo{person}{Hanna~M.
  Wallach}, \bibinfo{person}{Rob Fergus}, \bibinfo{person}{S.~V.~N.
  Vishwanathan}, {and} \bibinfo{person}{Roman Garnett}} (Eds.).
  \bibinfo{pages}{5998--6008}.
\newblock
\urldef\tempurl%
\url{https://proceedings.neurips.cc/paper/2017/hash/3f5ee243547dee91fbd053c1c4a845aa-Abstract.html}
\showURL{%
\tempurl}


\bibitem[\protect\citeauthoryear{Wang, Zha, Chen, Xiong, and Luo}{Wang
  et~al\mbox{.}}{2020b}]%
        {wang2020dual}
\bibfield{author}{\bibinfo{person}{Hao Wang}, \bibinfo{person}{Zheng{-}Jun
  Zha}, \bibinfo{person}{Xuejin Chen}, \bibinfo{person}{Zhiwei Xiong}, {and}
  \bibinfo{person}{Jiebo Luo}.} \bibinfo{year}{2020}\natexlab{b}.
\newblock \showarticletitle{Dual Path Interaction Network for Video Moment
  Localization}. In \bibinfo{booktitle}{\emph{{MM} '20: The 28th {ACM}
  International Conference on Multimedia, Virtual Event / Seattle, WA, USA,
  October 12-16, 2020}}. \bibinfo{pages}{4116--4124}.
\newblock
\urldef\tempurl%
\url{https://doi.org/10.1145/3394171.3413975}
\showDOI{\tempurl}


\bibitem[\protect\citeauthoryear{Wang, Zha, Li, Liu, and Luo}{Wang
  et~al\mbox{.}}{2021}]%
        {wang2021structured}
\bibfield{author}{\bibinfo{person}{Hao Wang}, \bibinfo{person}{Zheng-Jun Zha},
  \bibinfo{person}{Liang Li}, \bibinfo{person}{Dong Liu}, {and}
  \bibinfo{person}{Jiebo Luo}.} \bibinfo{year}{2021}\natexlab{}.
\newblock \showarticletitle{Structured Multi-Level Interaction Network for
  Video Moment Localization via Language Query}. In
  \bibinfo{booktitle}{\emph{Proceedings of the IEEE/CVF Conference on Computer
  Vision and Pattern Recognition}}. \bibinfo{pages}{7026--7035}.
\newblock


\bibitem[\protect\citeauthoryear{Wang, Ma, and Jiang}{Wang
  et~al\mbox{.}}{2020a}]%
        {wang2020temporally}
\bibfield{author}{\bibinfo{person}{Jingwen Wang}, \bibinfo{person}{Lin Ma},
  {and} \bibinfo{person}{Wenhao Jiang}.} \bibinfo{year}{2020}\natexlab{a}.
\newblock \showarticletitle{Temporally grounding language queries in videos by
  contextual boundary-aware prediction}. In
  \bibinfo{booktitle}{\emph{Proceedings of the AAAI Conference on Artificial
  Intelligence}}, Vol.~\bibinfo{volume}{34}. \bibinfo{pages}{12168--12175}.
\newblock


\bibitem[\protect\citeauthoryear{Wang, Qiao, and Tang}{Wang
  et~al\mbox{.}}{2014}]%
        {wang2014action}
\bibfield{author}{\bibinfo{person}{Limin Wang}, \bibinfo{person}{Yu Qiao},
  {and} \bibinfo{person}{Xiaoou Tang}.} \bibinfo{year}{2014}\natexlab{}.
\newblock \showarticletitle{Action recognition and detection by combining
  motion and appearance features}.
\newblock \bibinfo{journal}{\emph{THUMOS14 Action Recognition Challenge}}
  \bibinfo{volume}{1}, \bibinfo{number}{2} (\bibinfo{year}{2014}),
  \bibinfo{pages}{2}.
\newblock


\bibitem[\protect\citeauthoryear{Wang, Huang, and Wang}{Wang
  et~al\mbox{.}}{2019}]%
        {wang2019language}
\bibfield{author}{\bibinfo{person}{Weining Wang}, \bibinfo{person}{Yan Huang},
  {and} \bibinfo{person}{Liang Wang}.} \bibinfo{year}{2019}\natexlab{}.
\newblock \showarticletitle{Language-Driven Temporal Activity Localization: {A}
  Semantic Matching Reinforcement Learning Model}. In
  \bibinfo{booktitle}{\emph{{IEEE} Conference on Computer Vision and Pattern
  Recognition, {CVPR} 2019, Long Beach, CA, USA, June 16-20, 2019}}.
  \bibinfo{publisher}{Computer Vision Foundation / {IEEE}},
  \bibinfo{pages}{334--343}.
\newblock
\urldef\tempurl%
\url{https://doi.org/10.1109/CVPR.2019.00042}
\showDOI{\tempurl}


\bibitem[\protect\citeauthoryear{Wu and Han}{Wu and Han}{2018}]%
        {wu2018multi}
\bibfield{author}{\bibinfo{person}{Aming Wu} {and} \bibinfo{person}{Yahong
  Han}.} \bibinfo{year}{2018}\natexlab{}.
\newblock \showarticletitle{Multi-modal Circulant Fusion for Video-to-Language
  and Backward}. In \bibinfo{booktitle}{\emph{Proceedings of the Twenty-Seventh
  International Joint Conference on Artificial Intelligence, {IJCAI} 2018, July
  13-19, 2018, Stockholm, Sweden}},
  \bibfield{editor}{\bibinfo{person}{J{\'{e}}r{\^{o}}me Lang}} (Ed.).
  \bibinfo{publisher}{ijcai.org}, \bibinfo{pages}{1029--1035}.
\newblock
\urldef\tempurl%
\url{https://doi.org/10.24963/ijcai.2018/143}
\showDOI{\tempurl}


\bibitem[\protect\citeauthoryear{Wu, Li, Han, and Lin}{Wu
  et~al\mbox{.}}{2020a}]%
        {wu2020reinforcement}
\bibfield{author}{\bibinfo{person}{Jie Wu}, \bibinfo{person}{Guanbin Li},
  \bibinfo{person}{Xiaoguang Han}, {and} \bibinfo{person}{Liang Lin}.}
  \bibinfo{year}{2020}\natexlab{a}.
\newblock \showarticletitle{Reinforcement Learning for Weakly Supervised
  Temporal Grounding of Natural Language in Untrimmed Videos}. In
  \bibinfo{booktitle}{\emph{{MM} '20: The 28th {ACM} International Conference
  on Multimedia, Virtual Event / Seattle, WA, USA, October 12-16, 2020}}.
  \bibinfo{pages}{1283--1291}.
\newblock
\urldef\tempurl%
\url{https://doi.org/10.1145/3394171.3413862}
\showDOI{\tempurl}


\bibitem[\protect\citeauthoryear{Wu, Li, Liu, and Lin}{Wu
  et~al\mbox{.}}{2020b}]%
        {wu2020tree}
\bibfield{author}{\bibinfo{person}{Jie Wu}, \bibinfo{person}{Guanbin Li},
  \bibinfo{person}{Si Liu}, {and} \bibinfo{person}{Liang Lin}.}
  \bibinfo{year}{2020}\natexlab{b}.
\newblock \showarticletitle{Tree-structured policy based progressive
  reinforcement learning for temporally language grounding in video}. In
  \bibinfo{booktitle}{\emph{Proceedings of the AAAI Conference on Artificial
  Intelligence}}, Vol.~\bibinfo{volume}{34}. \bibinfo{pages}{12386--12393}.
\newblock


\bibitem[\protect\citeauthoryear{Xiao, Chen, Zhang, Ji, Shao, Ye, and
  Xiao}{Xiao et~al\mbox{.}}{2021}]%
        {xiao2021boundary}
\bibfield{author}{\bibinfo{person}{Shaoning Xiao}, \bibinfo{person}{Long Chen},
  \bibinfo{person}{Songyang Zhang}, \bibinfo{person}{Wei Ji},
  \bibinfo{person}{Jian Shao}, \bibinfo{person}{Lu Ye}, {and}
  \bibinfo{person}{Jun Xiao}.} \bibinfo{year}{2021}\natexlab{}.
\newblock \showarticletitle{Boundary Proposal Network for Two-Stage Natural
  Language Video Localization}. In \bibinfo{booktitle}{\emph{Proceedings of the
  AAAI Conference on Artificial Intelligence}}, Vol.~\bibinfo{volume}{35}.
  \bibinfo{pages}{2986--2994}.
\newblock


\bibitem[\protect\citeauthoryear{Xu, He, Plummer, Sigal, Sclaroff, and
  Saenko}{Xu et~al\mbox{.}}{2019a}]%
        {xu2019multilevel}
\bibfield{author}{\bibinfo{person}{Huijuan Xu}, \bibinfo{person}{Kun He},
  \bibinfo{person}{Bryan~A Plummer}, \bibinfo{person}{Leonid Sigal},
  \bibinfo{person}{Stan Sclaroff}, {and} \bibinfo{person}{Kate Saenko}.}
  \bibinfo{year}{2019}\natexlab{a}.
\newblock \showarticletitle{Multilevel language and vision integration for
  text-to-clip retrieval}. In \bibinfo{booktitle}{\emph{Proceedings of the AAAI
  Conference on Artificial Intelligence}}, Vol.~\bibinfo{volume}{33}.
  \bibinfo{pages}{9062--9069}.
\newblock


\bibitem[\protect\citeauthoryear{Xu, Yang, and Mao}{Xu et~al\mbox{.}}{2019b}]%
        {xu2019semantic}
\bibfield{author}{\bibinfo{person}{Yuecong Xu}, \bibinfo{person}{Jianfei Yang},
  {and} \bibinfo{person}{Kezhi Mao}.} \bibinfo{year}{2019}\natexlab{b}.
\newblock \showarticletitle{Semantic-filtered Soft-Split-Aware video captioning
  with audio-augmented feature}.
\newblock \bibinfo{journal}{\emph{Neurocomputing}}  \bibinfo{volume}{357}
  (\bibinfo{year}{2019}), \bibinfo{pages}{24--35}.
\newblock


\bibitem[\protect\citeauthoryear{Yang, Feng, Ji, Wang, and Chua}{Yang
  et~al\mbox{.}}{2021}]%
        {yang2021deconfounded}
\bibfield{author}{\bibinfo{person}{Xun Yang}, \bibinfo{person}{Fuli Feng},
  \bibinfo{person}{Wei Ji}, \bibinfo{person}{Meng Wang}, {and}
  \bibinfo{person}{Tat-Seng Chua}.} \bibinfo{year}{2021}\natexlab{}.
\newblock \showarticletitle{Deconfounded Video Moment Retrieval with Causal
  Intervention}. In \bibinfo{booktitle}{\emph{Proceedings of the 44th
  International ACM SIGIR Conference on Research and Development in Information
  Retrieval}} (Virtual Event, Canada) \emph{(\bibinfo{series}{SIGIR '21})}.
  \bibinfo{publisher}{Association for Computing Machinery},
  \bibinfo{pages}{1–10}.
\newblock
\urldef\tempurl%
\url{https://doi.org/10.1145/3404835.3462823}
\showDOI{\tempurl}


\bibitem[\protect\citeauthoryear{Yao, Mei, and Rui}{Yao et~al\mbox{.}}{2016}]%
        {yao2016highlight}
\bibfield{author}{\bibinfo{person}{Ting Yao}, \bibinfo{person}{Tao Mei}, {and}
  \bibinfo{person}{Yong Rui}.} \bibinfo{year}{2016}\natexlab{}.
\newblock \showarticletitle{Highlight Detection with Pairwise Deep Ranking for
  First-Person Video Summarization}. In \bibinfo{booktitle}{\emph{2016 {IEEE}
  Conference on Computer Vision and Pattern Recognition, {CVPR} 2016, Las
  Vegas, NV, USA, June 27-30, 2016}}. \bibinfo{publisher}{{IEEE} Computer
  Society}, \bibinfo{pages}{982--990}.
\newblock
\urldef\tempurl%
\url{https://doi.org/10.1109/CVPR.2016.112}
\showDOI{\tempurl}


\bibitem[\protect\citeauthoryear{Yeung, Russakovsky, Mori, and Fei{-}Fei}{Yeung
  et~al\mbox{.}}{2016}]%
        {yeung2016end}
\bibfield{author}{\bibinfo{person}{Serena Yeung}, \bibinfo{person}{Olga
  Russakovsky}, \bibinfo{person}{Greg Mori}, {and} \bibinfo{person}{Li
  Fei{-}Fei}.} \bibinfo{year}{2016}\natexlab{}.
\newblock \showarticletitle{End-to-End Learning of Action Detection from Frame
  Glimpses in Videos}. In \bibinfo{booktitle}{\emph{2016 {IEEE} Conference on
  Computer Vision and Pattern Recognition, {CVPR} 2016, Las Vegas, NV, USA,
  June 27-30, 2016}}. \bibinfo{publisher}{{IEEE} Computer Society},
  \bibinfo{pages}{2678--2687}.
\newblock
\urldef\tempurl%
\url{https://doi.org/10.1109/CVPR.2016.293}
\showDOI{\tempurl}


\bibitem[\protect\citeauthoryear{Yuan, Lan, Chen, Liu, and Zhu}{Yuan
  et~al\mbox{.}}{2021}]%
        {yuan2021closer}
\bibfield{author}{\bibinfo{person}{Yitian Yuan}, \bibinfo{person}{Xiaohan Lan},
  \bibinfo{person}{Long Chen}, \bibinfo{person}{Wei Liu}, {and}
  \bibinfo{person}{Wenwu Zhu}.} \bibinfo{year}{2021}\natexlab{}.
\newblock \showarticletitle{A Closer Look at Temporal Sentence Grounding in
  Videos: Datasets and Metrics}.
\newblock \bibinfo{journal}{\emph{arXiv preprint arXiv:2101.09028}}
  (\bibinfo{year}{2021}).
\newblock
\urldef\tempurl%
\url{https://arxiv.org/abs/2101.09028}
\showURL{%
\tempurl}


\bibitem[\protect\citeauthoryear{Yuan, Ma, Wang, Liu, and Zhu}{Yuan
  et~al\mbox{.}}{2019a}]%
        {yuan2019semantic}
\bibfield{author}{\bibinfo{person}{Yitian Yuan}, \bibinfo{person}{Lin Ma},
  \bibinfo{person}{Jingwen Wang}, \bibinfo{person}{Wei Liu}, {and}
  \bibinfo{person}{Wenwu Zhu}.} \bibinfo{year}{2019}\natexlab{a}.
\newblock \showarticletitle{Semantic Conditioned Dynamic Modulation for
  Temporal Sentence Grounding in Videos}. In \bibinfo{booktitle}{\emph{Advances
  in Neural Information Processing Systems 32: Annual Conference on Neural
  Information Processing Systems 2019, NeurIPS 2019, December 8-14, 2019,
  Vancouver, BC, Canada}}, \bibfield{editor}{\bibinfo{person}{Hanna~M.
  Wallach}, \bibinfo{person}{Hugo Larochelle}, \bibinfo{person}{Alina
  Beygelzimer}, \bibinfo{person}{Florence d'Alch{\'{e}}{-}Buc},
  \bibinfo{person}{Emily~B. Fox}, {and} \bibinfo{person}{Roman Garnett}}
  (Eds.). \bibinfo{pages}{534--544}.
\newblock
\urldef\tempurl%
\url{https://proceedings.neurips.cc/paper/2019/hash/6883966fd8f918a4aa29be29d2c386fb-Abstract.html}
\showURL{%
\tempurl}


\bibitem[\protect\citeauthoryear{Yuan, Mei, Cui, and Zhu}{Yuan
  et~al\mbox{.}}{2017}]%
        {yuan2017video}
\bibfield{author}{\bibinfo{person}{Yitian Yuan}, \bibinfo{person}{Tao Mei},
  \bibinfo{person}{Peng Cui}, {and} \bibinfo{person}{Wenwu Zhu}.}
  \bibinfo{year}{2017}\natexlab{}.
\newblock \showarticletitle{Video summarization by learning deep side semantic
  embedding}.
\newblock \bibinfo{journal}{\emph{IEEE Transactions on Circuits and Systems for
  Video Technology}} \bibinfo{volume}{29}, \bibinfo{number}{1}
  (\bibinfo{year}{2017}), \bibinfo{pages}{226--237}.
\newblock


\bibitem[\protect\citeauthoryear{Yuan, Mei, and Zhu}{Yuan
  et~al\mbox{.}}{2019b}]%
        {yuan2019find}
\bibfield{author}{\bibinfo{person}{Yitian Yuan}, \bibinfo{person}{Tao Mei},
  {and} \bibinfo{person}{Wenwu Zhu}.} \bibinfo{year}{2019}\natexlab{b}.
\newblock \showarticletitle{To find where you talk: Temporal sentence
  localization in video with attention based location regression}. In
  \bibinfo{booktitle}{\emph{Proceedings of the AAAI Conference on Artificial
  Intelligence}}, Vol.~\bibinfo{volume}{33}. \bibinfo{pages}{9159--9166}.
\newblock


\bibitem[\protect\citeauthoryear{Zeng, Xu, Huang, Chen, Tan, and Gan}{Zeng
  et~al\mbox{.}}{2020}]%
        {zeng2020dense}
\bibfield{author}{\bibinfo{person}{Runhao Zeng}, \bibinfo{person}{Haoming Xu},
  \bibinfo{person}{Wenbing Huang}, \bibinfo{person}{Peihao Chen},
  \bibinfo{person}{Mingkui Tan}, {and} \bibinfo{person}{Chuang Gan}.}
  \bibinfo{year}{2020}\natexlab{}.
\newblock \showarticletitle{Dense Regression Network for Video Grounding}. In
  \bibinfo{booktitle}{\emph{2020 {IEEE/CVF} Conference on Computer Vision and
  Pattern Recognition, {CVPR} 2020, Seattle, WA, USA, June 13-19, 2020}}.
  \bibinfo{publisher}{{IEEE}}, \bibinfo{pages}{10284--10293}.
\newblock
\urldef\tempurl%
\url{https://doi.org/10.1109/CVPR42600.2020.01030}
\showDOI{\tempurl}


\bibitem[\protect\citeauthoryear{Zhang, Hu, Lee, Zhao, Chammas, Jain, Ie, and
  Sha}{Zhang et~al\mbox{.}}{2020a}]%
        {zhang2020hierarchical}
\bibfield{author}{\bibinfo{person}{Bowen Zhang}, \bibinfo{person}{Hexiang Hu},
  \bibinfo{person}{Joonseok Lee}, \bibinfo{person}{Ming Zhao},
  \bibinfo{person}{Sheide Chammas}, \bibinfo{person}{Vihan Jain},
  \bibinfo{person}{Eugene Ie}, {and} \bibinfo{person}{Fei Sha}.}
  \bibinfo{year}{2020}\natexlab{a}.
\newblock \showarticletitle{A Hierarchical Multi-Modal Encoder for Moment
  Localization in Video Corpus}.
\newblock \bibinfo{journal}{\emph{arXiv preprint arXiv:2011.09046}}
  (\bibinfo{year}{2020}).
\newblock
\urldef\tempurl%
\url{https://arxiv.org/abs/2011.09046}
\showURL{%
\tempurl}


\bibitem[\protect\citeauthoryear{Zhang, Dai, Wang, Wang, and Davis}{Zhang
  et~al\mbox{.}}{2019a}]%
        {zhang2019man}
\bibfield{author}{\bibinfo{person}{Da Zhang}, \bibinfo{person}{Xiyang Dai},
  \bibinfo{person}{Xin Wang}, \bibinfo{person}{Yuan{-}Fang Wang}, {and}
  \bibinfo{person}{Larry~S. Davis}.} \bibinfo{year}{2019}\natexlab{a}.
\newblock \showarticletitle{{MAN:} Moment Alignment Network for Natural
  Language Moment Retrieval via Iterative Graph Adjustment}. In
  \bibinfo{booktitle}{\emph{{IEEE} Conference on Computer Vision and Pattern
  Recognition, {CVPR} 2019, Long Beach, CA, USA, June 16-20, 2019}}.
  \bibinfo{publisher}{Computer Vision Foundation / {IEEE}},
  \bibinfo{pages}{1247--1257}.
\newblock
\urldef\tempurl%
\url{https://doi.org/10.1109/CVPR.2019.00134}
\showDOI{\tempurl}


\bibitem[\protect\citeauthoryear{Zhang, Sun, Jing, Nan, Zhen, Zhou, and
  Goh}{Zhang et~al\mbox{.}}{2021a}]%
        {zhang2021video}
\bibfield{author}{\bibinfo{person}{Hao Zhang}, \bibinfo{person}{Aixin Sun},
  \bibinfo{person}{Wei Jing}, \bibinfo{person}{Guoshun Nan},
  \bibinfo{person}{Liangli Zhen}, \bibinfo{person}{Joey~Tianyi Zhou}, {and}
  \bibinfo{person}{Rick Siow~Mong Goh}.} \bibinfo{year}{2021}\natexlab{a}.
\newblock \showarticletitle{Video Corpus Moment Retrieval with Contrastive
  Learning}. In \bibinfo{booktitle}{\emph{Proceedings of the 44th International
  ACM SIGIR Conference on Research and Development in Information Retrieval}}
  (Virtual Event, Canada) \emph{(\bibinfo{series}{SIGIR '21})}.
  \bibinfo{publisher}{Association for Computing Machinery}.
\newblock


\bibitem[\protect\citeauthoryear{Zhang, Sun, Jing, and Zhou}{Zhang
  et~al\mbox{.}}{2020d}]%
        {zhang2020span}
\bibfield{author}{\bibinfo{person}{Hao Zhang}, \bibinfo{person}{Aixin Sun},
  \bibinfo{person}{Wei Jing}, {and} \bibinfo{person}{Joey~Tianyi Zhou}.}
  \bibinfo{year}{2020}\natexlab{d}.
\newblock \showarticletitle{Span-based Localizing Network for Natural Language
  Video Localization}. In \bibinfo{booktitle}{\emph{Proceedings of the 58th
  Annual Meeting of the Association for Computational Linguistics}}.
  \bibinfo{publisher}{Association for Computational Linguistics},
  \bibinfo{address}{Online}, \bibinfo{pages}{6543--6554}.
\newblock
\urldef\tempurl%
\url{https://doi.org/10.18653/v1/2020.acl-main.585}
\showDOI{\tempurl}


\bibitem[\protect\citeauthoryear{Zhang, Grauman, and Sha}{Zhang
  et~al\mbox{.}}{2018}]%
        {zhang2018retrospective}
\bibfield{author}{\bibinfo{person}{Ke Zhang}, \bibinfo{person}{Kristen
  Grauman}, {and} \bibinfo{person}{Fei Sha}.} \bibinfo{year}{2018}\natexlab{}.
\newblock \showarticletitle{Retrospective encoders for video summarization}. In
  \bibinfo{booktitle}{\emph{Proceedings of the European Conference on Computer
  Vision (ECCV)}}. \bibinfo{pages}{383--399}.
\newblock


\bibitem[\protect\citeauthoryear{Zhang, Yang, Chen, Ji, Xu, Li, and Shen}{Zhang
  et~al\mbox{.}}{2021b}]%
        {zhang2021multi}
\bibfield{author}{\bibinfo{person}{Mingxing Zhang}, \bibinfo{person}{Yang
  Yang}, \bibinfo{person}{Xinghan Chen}, \bibinfo{person}{Yanli Ji},
  \bibinfo{person}{Xing Xu}, \bibinfo{person}{Jingjing Li}, {and}
  \bibinfo{person}{Heng~Tao Shen}.} \bibinfo{year}{2021}\natexlab{b}.
\newblock \showarticletitle{Multi-Stage Aggregated Transformer Network for
  Temporal Language Localization in Videos}. In
  \bibinfo{booktitle}{\emph{Proceedings of the IEEE/CVF Conference on Computer
  Vision and Pattern Recognition}}. \bibinfo{pages}{12669--12678}.
\newblock


\bibitem[\protect\citeauthoryear{Zhang, Peng, Fu, and Luo}{Zhang
  et~al\mbox{.}}{2020c}]%
        {zhang2020learning}
\bibfield{author}{\bibinfo{person}{Songyang Zhang}, \bibinfo{person}{Houwen
  Peng}, \bibinfo{person}{Jianlong Fu}, {and} \bibinfo{person}{Jiebo Luo}.}
  \bibinfo{year}{2020}\natexlab{c}.
\newblock \showarticletitle{Learning 2d temporal adjacent networks for moment
  localization with natural language}. In \bibinfo{booktitle}{\emph{Proceedings
  of the AAAI Conference on Artificial Intelligence}},
  Vol.~\bibinfo{volume}{34}. \bibinfo{pages}{12870--12877}.
\newblock


\bibitem[\protect\citeauthoryear{Zhang, Lin, Zhao, and Xiao}{Zhang
  et~al\mbox{.}}{2019b}]%
        {zhang2019cross}
\bibfield{author}{\bibinfo{person}{Zhu Zhang}, \bibinfo{person}{Zhijie Lin},
  \bibinfo{person}{Zhou Zhao}, {and} \bibinfo{person}{Zhenxin Xiao}.}
  \bibinfo{year}{2019}\natexlab{b}.
\newblock \showarticletitle{Cross-Modal Interaction Networks for Query-Based
  Moment Retrieval in Videos}. In \bibinfo{booktitle}{\emph{Proceedings of the
  42nd International {ACM} {SIGIR} Conference on Research and Development in
  Information Retrieval, {SIGIR} 2019, Paris, France, July 21-25, 2019}},
  \bibfield{editor}{\bibinfo{person}{Benjamin Piwowarski}, \bibinfo{person}{Max
  Chevalier}, \bibinfo{person}{{\'{E}}ric Gaussier}, \bibinfo{person}{Yoelle
  Maarek}, \bibinfo{person}{Jian{-}Yun Nie}, {and} \bibinfo{person}{Falk
  Scholer}} (Eds.). \bibinfo{publisher}{{ACM}}, \bibinfo{pages}{655--664}.
\newblock
\urldef\tempurl%
\url{https://doi.org/10.1145/3331184.3331235}
\showDOI{\tempurl}


\bibitem[\protect\citeauthoryear{Zhang, Lin, Zhao, Zhu, and He}{Zhang
  et~al\mbox{.}}{2020b}]%
        {zhang2020regularized}
\bibfield{author}{\bibinfo{person}{Zhu Zhang}, \bibinfo{person}{Zhijie Lin},
  \bibinfo{person}{Zhou Zhao}, \bibinfo{person}{Jieming Zhu}, {and}
  \bibinfo{person}{Xiuqiang He}.} \bibinfo{year}{2020}\natexlab{b}.
\newblock \showarticletitle{Regularized Two-Branch Proposal Networks for
  Weakly-Supervised Moment Retrieval in Videos}. In
  \bibinfo{booktitle}{\emph{{MM} '20: The 28th {ACM} International Conference
  on Multimedia, Virtual Event / Seattle, WA, USA, October 12-16, 2020}}.
  \bibinfo{pages}{4098--4106}.
\newblock
\urldef\tempurl%
\url{https://doi.org/10.1145/3394171.3413967}
\showDOI{\tempurl}


\bibitem[\protect\citeauthoryear{Zhang, Zhao, Lin, He, et~al\mbox{.}}{Zhang
  et~al\mbox{.}}{2020e}]%
        {zhang2020counterfactual}
\bibfield{author}{\bibinfo{person}{Zhu Zhang}, \bibinfo{person}{Zhou Zhao},
  \bibinfo{person}{Zhijie Lin}, \bibinfo{person}{Xiuqiang He}, {et~al\mbox{.}}}
  \bibinfo{year}{2020}\natexlab{e}.
\newblock \showarticletitle{Counterfactual Contrastive Learning for
  Weakly-Supervised Vision-Language Grounding}.
\newblock \bibinfo{journal}{\emph{Advances in Neural Information Processing
  Systems}}  \bibinfo{volume}{33} (\bibinfo{year}{2020}),
  \bibinfo{pages}{18123--18134}.
\newblock


\end{thebibliography}

\end{document}